\documentclass[acmlarge,anonymous=False]{acmart}
\usepackage{ragged2e} 
\AtBeginDocument{%
  }

\begin{document}

\title{A Survey on Federated Learning in Human Sensing}

\author{Mohan Li}
\affiliation{%
 \institution{Università della Svizzera italiana}
 \city{Lugano}
 \country{Switzerland}}

 \author{Martin Gjoreski}
\affiliation{%
 \institution{Università della Svizzera italiana}
 \city{Lugano}
 \country{Switzerland}}

 \author{Pietro Barbiero}
\affiliation{%
 \institution{Università della Svizzera italiana}
 \city{Lugano}
 \country{Switzerland}}

  \author{Gašper Slapničar}
\affiliation{%
 \institution{Jožef Stefan Institute}
 \city{Ljubljana}
 \country{Slovenia}}

  \author{Mitja Luštrek}
\affiliation{%
 \institution{Jožef Stefan Institute}
 \city{Ljubljana}
 \country{Slovenia}}

   \author{Nicholas D. Lane}
\affiliation{%
 \institution{University of Cambridge}
 \city{Cambridge}
 \country{United Kingdom}}
 
 \author{Marc Langheinrich}
\affiliation{%
 \institution{Università della Svizzera italiana}
 \city{Lugano}
 \country{Switzerland}}

\renewcommand{\shortauthors}{Li et al.}
\begin{abstract}

Human Sensing---a field that leverages technology to monitor human activities, psycho-physiological states, and interactions with the environment—enhances our understanding of human behavior and drives the development of advanced services that improve overall quality of life. However, its reliance on detailed and often privacy-sensitive data as the basis for its machine learning (ML) models raises significant legal and ethical concerns. The recently proposed ML approach of \emph{Federated Learning (FL)} promises to alleviate many of these concerns, as it is able to create accurate ML models without sending raw user data to a central server. While FL has demonstrated its usefulness across a variety of areas, such as text prediction and cyber security, its benefits in Human Sensing are under-explored, given the particular challenges in this domain. This survey conducts a comprehensive analysis of the current state-of-the-art studies on FL in Human Sensing, and proposes a taxonomy and an eight-dimensional assessment for FL approaches. Through the eight-dimensional assessment, we then evaluate whether the surveyed studies consider a specific FL-in-Human-Sensing challenge or not. Finally, based on the overall analysis, we discuss open challenges and highlight five research aspects related to FL in Human Sensing that
require urgent research attention. Our work provides a comprehensive corpus of FL studies and aims to assist FL practitioners in developing and evaluating solutions that effectively address the real-world complexities of Human Sensing.
\end{abstract}

\ccsdesc[500]{General and reference~Surveys and overviews}
\ccsdesc[500]{Computing methodologies~Distributed artificial intelligence}
\ccsdesc[500]{Human-centered computing~Ubiquitous and mobile computing}

\keywords{Federated Learning, Human Sensing}


\maketitle

\section{Introduction}
\label{sec:intro}

The development of sensor technology and the proliferation of consumer-grade wearable devices \cite{heikenfeld2018wearable,10.1145/3596599}, e.g., smartwatches, smart glasses, and smart rings, have significantly advanced the field of Human Sensing \cite{10.1145/3491208,teixeira2010survey}. This progress facilitates access to personal physiological and behavioral data outside traditional healthcare settings and allows for a variety of applications. Utilizing Deep Learning (DL) and data from diverse sensing modalities, notable applications include, e.g., Human Activity Recognition (HAR) \cite{9353723,8794643}, physiological sensing, human affective and mental states recognition (e.g., emotions, stress, and cognitive load), mobility prediction \cite{barbosa2018human,10.1145/3485125}, user identification, and interface development \cite{10.1145/3234149}.
The growing research efforts to utilize personal activity data highlight a shift towards more adaptive and robust Human Sensing systems for unique individual needs.

However, even though billions of wearable devices collect data from humans daily, most of these systems do not explicitly take their users' privacy into account. 
While many legal frameworks in place today (e.g., the EU General Data Protection Regulation (GDPR) \cite{regulation2016regulation} or the California Consumer Privacy Act (CCPA) \cite{pardau2018california}) require explicit user consent before transferring the collected personal data to centralized ML servers, few applications go beyond the legal requirement and consider data minimization principles \cite{pfitzmann2010terminology}, i.e., avoid such transfer in the first place. This is because service providers require such privacy-sensitive data to enable personalized services, such as health monitoring or location-based recommendations. Furthermore, in typical data-driven systems, the quality of the services is proportional to the size of the training data, indicating a delicate trade-off between privacy and utility. Thus, it is essential for future machine learning (ML) and Artificial Intelligence (AI)-enhanced societies to design privacy frameworks that enable users to safely share their data for the purpose of personalized and accurate services. 

Avoiding the transfer of personal data goes beyond satisfying user privacy concerns. Even in scenarios where a sufficient number of users are willing to share their data, the cost of transmission and storage is proportional to the size of the data. This cost can be substantial for use cases based on video, audio, and other multi-channel and high-sampling data modalities (e.g., EEG -- Electroencephalography devices).

\begin{figure}[htb]
  \centering
  \includegraphics[width=12cm]{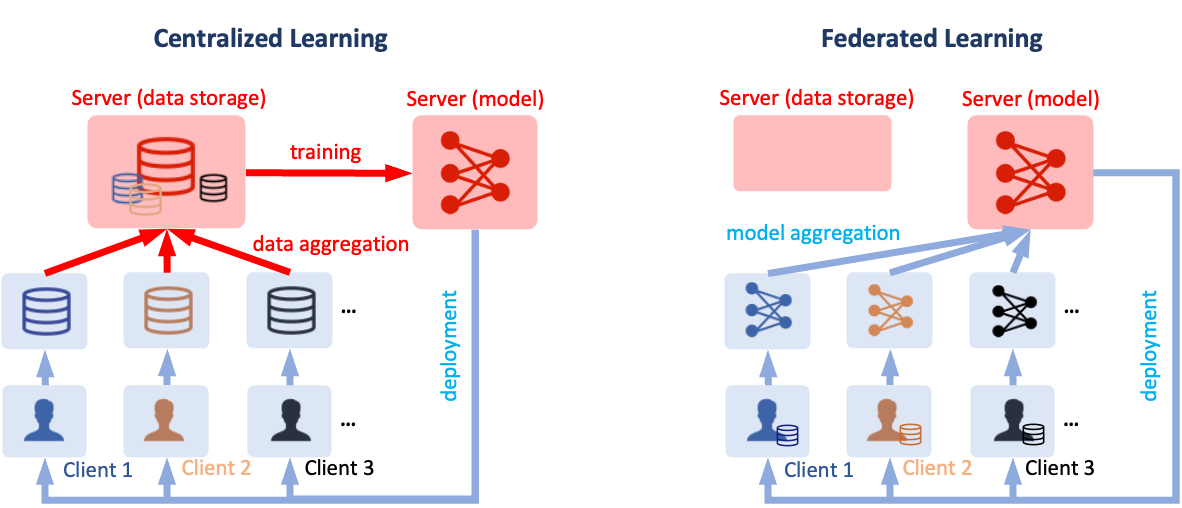}
  \caption{A comparison between Centralized Learning and Federated Learning}
  \Description{A Comparison between Centralized Learning and Federated Learning}
  \label{fig:fig_comp}
  \vspace{-1.2em}
\end{figure}

To address the privacy issue inherent in traditional ML applications, McMahan et al. \cite{pmlr-v54-mcmahan17a} presented a novel decentralized ML approach named Federated Learning (FL), with its initial implementation Federated stochastic gradient descent (FedSGD) and FederatedAveraging (FedAvg). We illustrate FL's basic idea in Fig. \ref{fig:fig_comp}. FL assumes a set of distributed participating devices (\textit{clients}) coordinated by a central \textit{server}. The clients first train local models with local data. The local models are then communicated to the server, which aggregates them into a global model. The global model is then sent back to the clients for another iteration of local training. This process is performed iteratively until a model convergence is achieved –– a process orchestrated by the server. In some scenarios, the server might train an initial model using publicly available data and then communicate this pre-trained model to the clients as a starting point. Compared to centralized learning, FL enables clients to transmit model parameters only, instead of raw user data, promising clear theoretical advantages in both privacy and communication cost compared to traditional centralized ML approaches. 


However, real-world tasks do not always fit into the theoretical FL setup. FL still faces many challenges, like federated model selection, backdoor attacks, and communication inefficiency, which we will explore further in later sections. Regardless of a large amount of follow-up work on FL to enhance its performance in different aspects, most studies evaluate FL-based frameworks on well-established datasets like MNIST \cite{726791}. While such experiments enable controlled and systematic advancement in the field, challenges specific to Human Sensing remain underexplored, including system heterogeneity, limited labeled data, and domain shifts in the sensor data or labels due to subjectivity. In this survey, we explicitly review the advances in FL for Human Sensing tasks and address the following research question: \textit{To what extent can FL address real-world human-sensing tasks?} To answer this question, we conducted a systematic literature search and evaluated each study using an eight-dimensional assessment (elaborated in Section \ref{sec:bg}).


This review article provides the following key contributions:

\begin{itemize} 
    \item It introduces FL in conjunction with Human Sensing and creates an application-oriented taxonomy to investigate real-world FL deployments in human-sensing scenarios. 
    \item It provides a comprehensive corpus of FL studies in Human Sensing. We identify six main application fields and analyze papers based on our proposed eight-dimensional assessment related to: (1) privacy and security, (2) communication cost, (3) system heterogeneity, (4) statistical heterogeneity, (5) unlabeled data usage, (6) simplified setup, (7) server-optimized FL, and (8) client-optimized FL.
    \item It discusses the open challenges and highlights five research aspects related to FL in Human Sensing that require urgent research attention.
\end{itemize}


The rest of the article is structured as follows. Section \ref{sec:bg} provides more background information on FL with a focus on the key challenges that motivate the need for this survey. Section \ref{sec:method} presents the systematic methodology for this survey. Sections \ref{sec:asp} to \ref{sec:hci} present insights specific to each of the six main application domains that we identified in our systematic literature search. The application domains include: \emph{Audio and Speech Processing} in Section \ref{sec:asp}; \emph{Well-being} in Section \ref{sec:Well-being}, \emph{User Identification} in Section \ref{sec:ii}, \emph{Human Mobility and Localization} in Section \ref{sec:hml}, \emph{Activity Recognition} in Section \ref{sec:ar}, and \emph{Interface Development} in Section \ref{sec:hci}. Lastly, Section \ref{sec:future} discusses the challenges and future directions, while Section \ref{sec:con} concludes the paper.


\section{Challenges in Federated Learning, and Assessment}
\label{sec:bg}

Many real-world deployments have successfully demonstrated the benefit of a Federated Learning approach yet many challenges remain. Fig.~\ref{fig:fig_challenge} summarizes eight principal dimensions that we identified in this context. We will use these eight dimensions to evaluate\footnote{Our eight dimensions evaluate studies only from an FL perspective. Lack of consideration in any of the eight dimensions does not diminish the significance of a study, as it may contribute substantially to other aspects.} the application of FL in the field of Human Sensing, allowing us to discover which challenges are adequately addressed in general or require more research attention.  



\begin{enumerate}
\item \textbf{Privacy and Security}:
Although FL systems do not share the underlying data used for training, communicating model weight updates to the server can still pose a risk to sensitive personal information \cite{9084352}. These weight updates mean that FL systems are susceptible to various client or server attacks during training and deployment \cite{9634881}.
For example, malicious attackers can distort the aggregated model by performing poisoning attacks such as ``label flipping'' \cite{fung2018mitigating} or uploading false model parameters \cite{pmlr-v97-bhagoji19a}. 
In a Membership Inference Attack (MIA) \cite{mothukuri2021survey}, a malicious client tries to identify whether a certain data point (e.g., user) has participated in a training iteration. In a Property Inference Attack (PIA) \cite{10.1145/3243734.3243834} instead, a malicious attacker can derive statistical properties about other clients' training data, e.g., the demographic distribution of their users.
In this survey, we consider a system to address \emph{Privacy and Security} if it employs defense strategies such as Differential Privacy (DP) \cite{mcmahan2017learning} or Secure Aggregation \cite{10.1145/3133956.3133982}, or if the authors conducted a vulnerability analysis (e.g., through attack models).

\item
\textbf{Communication Cost}:
FL involves frequent communication between numerous clients and a central server to aggregate local model updates. High communication costs can lead to increased latency, reduced bandwidth efficiency, and higher energy consumption, particularly on resource-constrained devices such as smartphones and Internet-of-Things (IoT) devices.
In our survey, we consider a system to address \emph{Communication Cost} if it explicitly considers edge devices (e.g., by developing a lightweight framework) or otherwise attempts to lower communication costs without substantially sacrificing model performance.


\item
\textbf{System Heterogeneity}:
System heterogeneity refers to the disparity in storage, computational resources, communication capabilities, and other hardware aspects among clients \cite{MLSYS2019_7b770da6}.
Current research often assumes an idealized scenario of homogeneous client systems, equipped with consistent resources and capabilities, and without connectivity issues. Clients with ``non-standard'' devices are typically excluded from the training process as ``outliers'' \cite{hard2018federated}. However, in practical scenarios, system heterogeneity is prevalent, and consistent connectivity cannot be guaranteed. Furthermore, excluding certain types of devices may lead to unfair and biased models \cite{kairouz2021advances}.
 In our survey, we consider a system to address \emph{System Heterogeneity} if its setup involves training or deployment on heterogeneous devices. 

\item \textbf{Statistical Heterogeneity}:
Statistical heterogeneity arises in FL from non-independent and non-identically distributed (non-IID) \cite{zhao2018federated} data between clients. The prevalence of non-IID scenarios in FL has led to extensive research \cite{zhu2021federated}, acknowledging the common occurrence of data skewness in attributes, labels, or temporal correlations.
Solutions for this heterogeneity encompass a wide array of strategies \cite{zhu2021federated}, such as
meta-learning \cite{NEURIPS2020_24389bfe} and clustering \cite{ghosh2020efficient}.
Nevertheless, statistical heterogeneity is still an open challenge that requires research attention.
We consider a system to address \emph{Statistical Heterogeneity} if it applies strategies to ensure robustness against non-IID-related attacks. The need for this of course highly depends on the underlying data, i.e., if it is actually non-IID. While we do not conduct a detailed examination of datasets if they are sourced from publicly available platforms, we expect systems using datasets that are inherently heterogeneous to explicitly analyze the impact of such heterogeneity on FL performance.

\begin{figure}[bt]
  \centering
  \includegraphics[width=12cm]{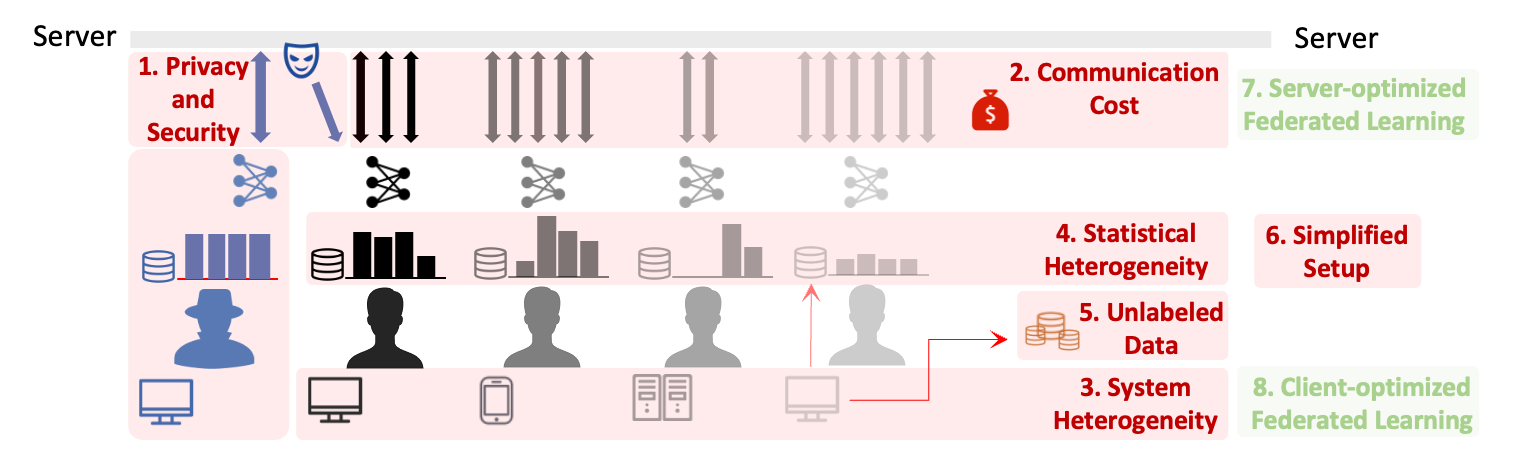}
  \caption{Our proposed eight-dimensional assessment of a typical FL framework}
  \Description{}
  \label{fig:fig_challenge}
\end{figure}


\item \textbf{Unlabeled Data Usage}:
In many FL scenarios, the server has a moderate amount of labeled data, while clients have little to no labeled data but a substantial quantity of unlabeled data \cite{jin2023federated}. Recent studies have shown a growing interest in effectively utilizing such unlabeled data \cite{jin2023federated}, but the number of these studies is very limited. To promote more data-efficient and realistic FL deployments, we consider a system to address \emph{Unlabeled Data} if it employs techniques such as pseudolabeling, unsupervised learning, semi-supervised learning, and other types of representation learning that do not require labeled data.

\item \textbf{Simplified Setup}: 
We consider a system to use a \emph{Simplified Setup} if it employs unnatural task definitions or synthetic data partitioning. 
For example, in an FL face recognition system, a mobile phone client uses a dataset comprising face images of multiple individuals, which is contrary to the typical scenario where the data comes from a single individual. 
Another example might be a system that divides an activity recognition dataset from 12 individuals between only four artificial clients during training, which does not reflect the real-world scenario.

\item \textbf{Server-optimized Federated Learning}:
From a server's viewpoint, a key advantage of FL is the ability to expand the dataset by training a model with billions of data points from client devices while avoiding privacy concerns. We classify an FL deployment as \emph{Server-optimized} when the server maintains a well-performing model that represents the majority of clients. In this context, the server employs methods to enhance model convergence speed and overall performance. However, such server-based optimization may come at the cost of decreased client fairness (e.g., ``outlier'' clients may not be well represented by the joint model) or a degraded user experience (e.g., excluding clients that lack the necessary hardware or connection capabilities).

\item \textbf{Client-optimized Federated Learning}:
Clients participate in FL primarily because the shared model improves their local performance.
A ``selfish'' client might focus solely on personal gains, disregarding the overall performance of the system, leading to dissatisfaction among other clients, especially in non-IID settings.
Many Human Sensing tasks are ``selfish'' in nature, e.g., individual users (clients) are concerned with the accuracy of their own smartwatches in measuring heart rate, and this accuracy should not be compromised for the sake of improving average performance across diverse users, i.e., clients do not need to know and do not care about each other.
Implementing personalization techniques such as clustering or local retraining is a valid approach to achieve \emph{client-optimized} FL. These methods tailor the learning process to individual client characteristics, ensuring that each client's unique requirements are met.
It should be noted that \emph{server-optimized} FL is not the opposite of \emph{client-optimized} FL. Non-server-optimized FL typically involves minimal server participation, where the server acts more like a relay station to facilitate information exchange among clients.
An FL framework can excel in both areas or may represent an initial attempt that fits neither category.

\end{enumerate}

\section{REVIEW METHODOLOGY}
\label{sec:method}
\begin{figure}[htb]
  \centering
  \includegraphics[width=9cm]{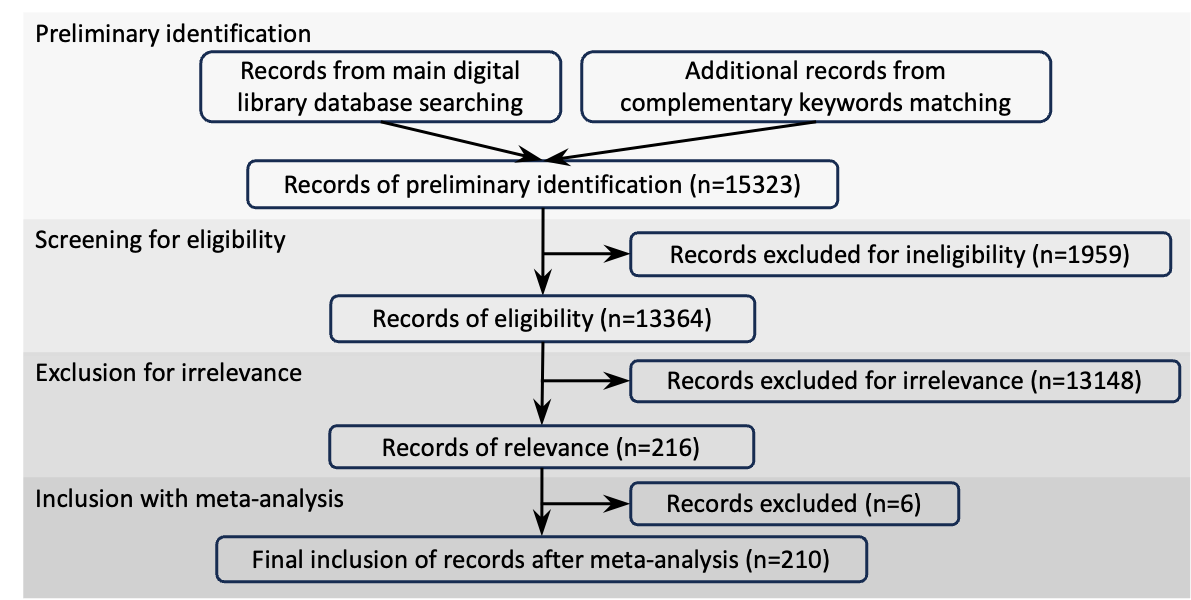}
  \caption{An Overall Workflow of PRISMA in This Survey}
  \label{fig:fig_prisma}
\end{figure}

To thoroughly understand current FL research in human sensing, we employed the PRISMA (Preferred Reporting Items for Systematic Reviews and Meta-Analyses) methodology \cite{KITCHENHAM20097,MOHER2010336} for constructing our systematic literature corpus. We executed a four-step filtering process: preliminary identification, screening for eligibility, exclusion for irrelevance, and final inclusion with meta-analysis. The following subsections detail each step and clarify our approach. A comprehensive PRISMA flow diagram is included as Fig. \ref{fig:fig_prisma} in our report.

\subsection{Preliminary Identification}
We identify three main digital libraries for the preliminary paper retrieval: IEEE Xplore (IEEE-X), SpringerLink, and ACM Digital Library (ACM-DL). Our initial search strategy involved using the term "federated learning" without specific constraints for human sensing. This approach was based on three considerations:
First, we did not use "human sensing" as the main keyword because it is an interdisciplinary and extensive field. For this reason, relevant works might not explicitly mention this term in their title, abstract, or body. Also, keyword combination searches could be incomplete, given the diverse ways authors articulate their research.
Second, FL has a manageable volume of literature across these libraries as it represents a relatively recent field.
Third, it automatically excludes other decentralized training strategies such as blockchain-based ML \cite{8622598}.
The search patterns for each library were as follows:

\begin{itemize}
\item \textbf{IEEE-X}: Advanced Search with Search Term: "federated learning" in "All Metadata". Publication Year with Specify Year Range from "2016" (the year that "Federated Learning" is defined for the first time) to "2023".
\item \textbf{SpringerLink}: Search term: "federated learning" with Filters: Custom dates: Start year from "2016" and End year to "2023"; Languages as "English".
\item \textbf{ACM-DL}: Advanced Search with: Search items from "The ACM Guide to Computing Literature", Search Within: "Anywhere" by "federated learning"; Filters: "Publisher" with "Match None" in "IEEE Press" or "Springer-Verlag"; Publication Date with Custom range as from "Jan, 2016" to "Nov, 2023".
\end{itemize}

Besides digital libraries, we selected the following 13 proceedings of conferences based on their average ranking and relevance to human sensing:
\begin{itemize}
\item \textbf{AAAI}: National Conference of the American Association for Artificial Intelligence
\item \textbf{AAMAS}: International Conference on Autonomous Agents and Multiagent Systems
\item \textbf{ACL}: Association of Computational Linguistics
\item \textbf{CAV}: Computer Aided Verification
\item \textbf{ICAPS}: International Conference on Automated Planning and Scheduling
\item \textbf{ICIS}: International Conference on Information Systems
\item \textbf{ICML}: International Conference on Machine Learning
\item \textbf{IJCAI}: International Joint Conference on Artificial Intelligence
\item \textbf{IJCAR}: International Joint Conference on Automated Reasoning
\item \textbf{KR}: International Conference on Principles of Knowledge Representation \& Reasoning
\item \textbf{NeurIPS}: Conference on Neural Information Processing Systems
\item \textbf{ICLR}: International Conference on Learning Representations
\item \textbf{UAI}: Conference for Uncertainty in Artificial Intelligence

\end{itemize}

We additionally extended our search strategy using the ACM Computing Classification System (CCS) concepts \cite{code1998acm} for more targeted retrieval.
CCS concepts under "Human-centered computing", such as "Interaction paradigms" and "Ubiquitous and mobile devices", offer an expansive scope relevant to human sensing.
We use them for keyword matching in Google Scholar, in combination with "federated learning" to collect additional relevant papers from sources like arXiv or other platforms.
This comprehensive approach resulted in an initial corpus of 15,323 papers, with the final cut-off date for all venues being November 5, 2023.

\subsection{Screening for Eligibility}
We applied similar criteria as \cite{10.1145/3582272} for paper eligibility in the digital libraries: we only keep the journal, conference papers, and chapters of books, while excluding posters, work in progress, or other unrelated content types. The additional filters for each library are listed below:
\begin{itemize}
\item IEEE-X: Filters Applied: "Conferences", "Journals", "Early Access Articles".
\item SpringerLink: Filters: Content type: "Chapter", "Article", "Conference Paper", "Book", "Conference Proceedings".
\item ACM-DL: Publications: Content type: "Research Article", "Review Article", "Short paper".
\end{itemize}
1,959 papers are excluded due to ineligibility (including unavailability), and 13,364 papers are left for relevance screening.

\subsection{Exclusion for Irrelevance}
To narrow down our collection to papers at the intersection of FL and human sensing, we implemented two key exclusion principles:
\begin{itemize}
\item \textbf{Exclusion of non-application papers}: We excluded papers primarily focused on theoretical advancements in FL, such as those proposing enhancements in privacy, security, communication cost, data heterogeneity, and other technical areas. This also extends to reviews, comparative studies, and any research predominantly theoretical in nature, with only simulations on well-defined datasets.
\item \textbf{Exclusion of application papers in other fields}: We excluded papers that, while addressing real-world problems, focused on areas other than human sensing. This includes research in fields such as IoT (Internet of Things), recommender systems, blockchains, or any other application domains not directly related to human sensing.
For IoMT (Internet of Medical Things), we exclude papers on inpatient care with medical interventions, or medical and clinical research with patient data. The remaining studies in this review are focused on general daily health monitoring and other related healthcare topics.
\end{itemize}
There are 216 papers left after exclusion for irrelevance.

\subsection{Inclusion with Meta-analysis}
To finalize our taxonomy of human sensing in FL, we refined our selection based on the type of raw data used in the studies.
We further excluded 4 papers that primarily focused on the fundamental design of FL frameworks without task-oriented analysis, making them irrelevant to our review's scope.
In total, our selection process culminated in a final corpus of 211 papers.

\subsection{Overview of Taxonomy}

\begin{figure}[tb]
  \centering
  \includegraphics[width=\linewidth]{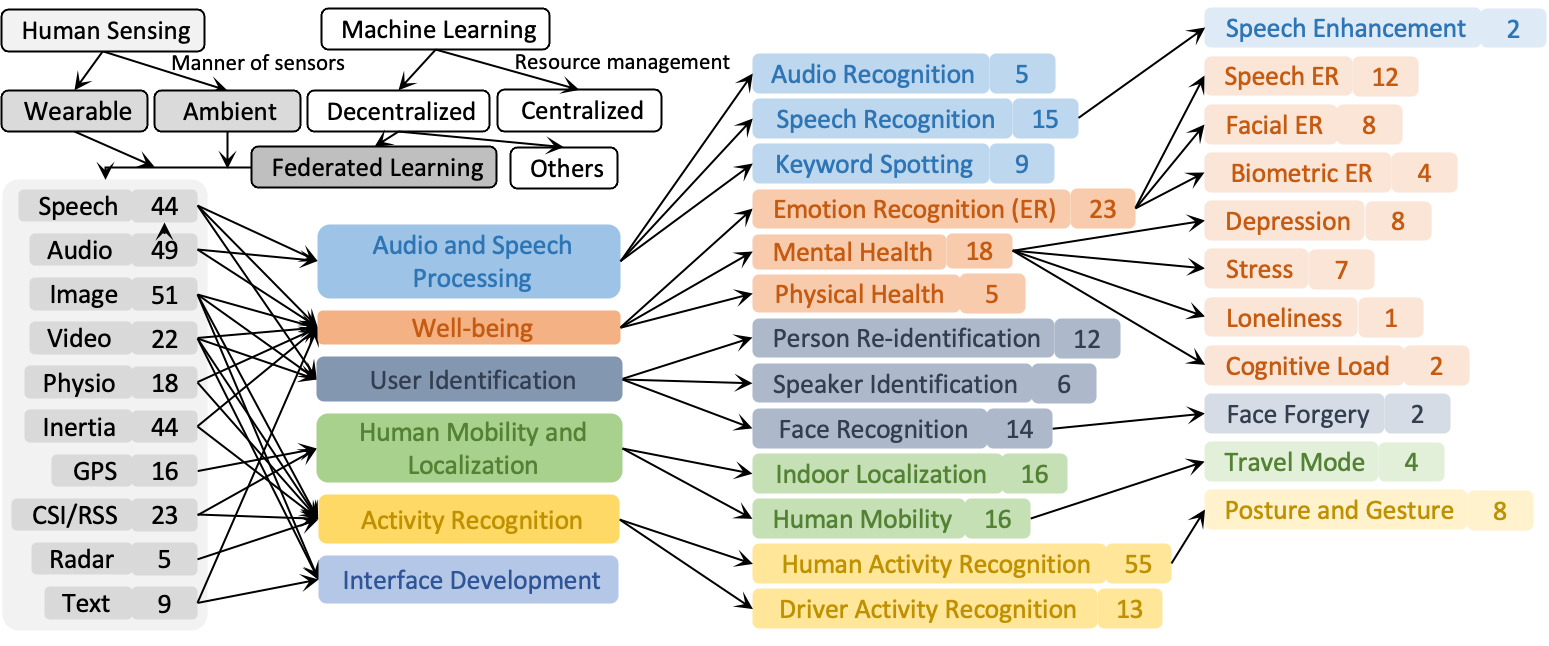}
  \caption{An overview of the proposed taxonomy in this survey. \textbf{Physio}: Physiological Signal. \textbf{CSI/RSS}: Channel State Information/Received Signal Strength (Indicator). The numbers represent the count of studies within the corresponding domain.}
  \Description{Taxonomy and A Statistic Overview}
  \label{fig:fig_taxonomy}
\end{figure}

Fig. \ref{fig:fig_taxonomy} shows an overview of the taxonomy we created based on the identified studies.
FL represents a decentralized approach against ML, grounded in principles of distributed resource management.
Regarding Human Sensing, we encompass a broad spectrum of modalities, including ambient sensing techniques, such as video recording, as well as more obtrusive methods, such as wearable devices. 
Each study is sorted by downstream tasks and raw data types. In total, we identified nine types of raw data: \emph{Radar}, \emph{Channel State Information/Received Signal Strength}, \emph{GPS (Global Positioning System)}, \emph{Physiological Signals}, \emph{Image}, \emph{Video}, \emph{Inertia}, \emph{Audio}, and \emph{Text}. Furthermore, we grouped the identified studies into six application domains: \emph{Audio and Speech Processing} in Section \ref{sec:asp}; \emph{Well-being} in Section \ref{sec:Well-being}, \emph{User Identification} in Section \ref{sec:ii}, \emph{Human Mobility and Localization} in Section \ref{sec:hml}, \emph{Activity Recognition} in Section \ref{sec:ar}, and \emph{Interface Development} in Section \ref{sec:hci}. Finally, we evaluate each study using our proposed eight-dimensional assessment strategy.

Existing survey studies on FL predominantly investigate fields distinct from the focus of our survey. For example, the IoT domain has been extensively explored in conjunction with FL due to its distributed nature \cite{9415623,9460016}, including applications in resource-constrained environments \cite{9475501} and vehicular IoT \cite{9086790}. The digital healthcare sector, particularly with its sensitive and decentralized medical data, similarly benefits significantly from FL, often referred to as the IoMT \cite{xu2021federated,rieke2020future}. Furthermore, the increasing number of FL studies in Recommender Systems is establishing a new area known as Federated Recommender Systems \cite{alamgir2022federated,10262054}.

Existing FL survey studies focus on fields different from what we cover in our survey.
For example, the IoT domain has been extensively investigated in conjunction with FL due to its distributed nature \cite{9415623,9460016}, as well as in resource-constrained environments \cite{9475501} and vehicular IoT \cite{9086790}.
The digital healthcare area, particularly with sensitive and decentralized medical data, benefits greatly from FL, referred to as the IoMT \cite{xu2021federated,rieke2020future}. Additionally, the growing number of FL studies in Recommender Systems is forging a new field known as Federated Recommender \cite{alamgir2022federated,10262054}.
In this review, we only focus on the Human Sensing domain, and investigate studies using FL as the main approach.

\begin{table*}
  \caption{Relevant FL Surveys and Reviews}
  \label{tb:collectionReview}
  \begin{tabular}{ll}
    \toprule    Topics&References\\
    \midrule
    General & \cite{9084352,kairouz2021advances,9220780,10.1145/3298981,10542323} \\
    Privacy and Security & \cite{9634881,mothukuri2021survey,10420449}   \\
    (Data) Heterogeneity &  \cite{zhu2021federated,ma2022state,9835537}  \\
    Application in IoT & \cite{9415623,9460016,9773116,9475501,9086790} \\
    Application in IoMT & \cite{xu2021federated,rieke2020future,10.1145/3412357,10.1145/3501296,rani2023federated} \\
    Application in Recommender Systems & \cite{alamgir2022federated,10262054} \\
  \bottomrule
\end{tabular}
\end{table*}

\section{Audio and Speech Processing}
\label{sec:asp}

In this section, we explore how the audio and speech signal processing community employs FL to safeguard sensitive individual information and examine the impact of heterogeneity on performance. We summarize all studies of this section in Table. \ref{tb:asp}. The following subsections include audio recognition, speech recognition, and keyword spotting, and we investigate how FL was implemented in terms of our eight-dimensions assessment.

\subsection{Audio Recognition}
An audio recognition system identifies and distinguishes audio signals from environmental sounds and human speech, including multi-class audio classification and binary-class event detection.
Key challenges include underexplored edge device data, unseen classes, and privacy concerns with centralized training.
This subsection focuses on FL insights from studies using ambient sound datasets.

\citet{gudur2021zero} introduced a zero-shot learning method to address unseen classes in FL for audio classification. Tested on the US8K dataset \cite{10.1145/2647868.2655045}, they generated Anonymized Data Impressions (crafted samples from a teacher model) with class similarity matrices to protect local raw data and identify new classes in ambient sound datasets.
\citet{9746356,10.1145/3520128} introduced the FedSTAR framework, leveraging unlabeled audio data on edge devices. In FedSTAR, a teacher model, trained on a small labeled dataset, generates pseudo-labels for a large unlabeled dataset. These pseudo-labels, combined with original labels, train a student model for global aggregation.
Tests with only 3\% labeled data showed competitive accuracy, highlighting the value of unlabeled data usage.
\citet{10.1007/978-3-031-45392-2_25} addressed FL audio-based violence detection using binary classification.
They converted audio into mel-spectrograms for image classification rather than using raw audio signals.
FedAudio, introduced in \cite{10096500}, is the first FL benchmark for audio tasks. It includes four audio datasets for three tasks: keyword spotting, speech emotion recognition, and sound event classification.
This work aims to advance privacy-friendly research in acoustic and audio processing.

\subsection{Speech Recognition}
Automatic Speech Recognition (ASR) aims to convert human speech into text, with applications like voice commands and transcription. However, speech contains sensitive attributes such as language, gender, age, height, and weight. 

We first present three studies on the vulnerability of FL-ASR.
\citet{9746541} 
found significant speaker information in the first layer of personalized acoustic models, highlighting the need for robust defenses in FL-ASR research.
\citet{10096426} addressed vulnerabilities highlighted by \citet{9746541} through a neural network layer-wise analysis.
They applied FL with a pre-trained wav2vec 2.0 \cite{NEURIPS2020_92d1e1eb}, and this approach effectively utilizes unlabeled speech data and shows promising performance, especially with unseen speakers.
\citet{10.1145/3447687} developed TFE, an ASR ecosystem combining FL with transfer and evolutionary learning, incorporating DP to ensure privacy.

Next, we introduce three papers focused on the computational cost of FL-ASR.
\citet{9413397} examined non-IID datasets' impact in ASR tasks, finding that random client data sampling can reduce this effect but increases computation cost.
\citet{9414305} expanded ASR tasks to diverse datasets, including broadcast news, hospitality speech, and speech with accents.
They minimized communication costs with minimal performance loss and introduced a client-adaptive FL approach to handle non-IID data by estimating client-specific transformations.
\citet{9746226} advocate for Federated Dropout \cite{caldas2018expanding} to address client-side computational and memory constraints, enabling on-device training.
This method downsizes client models by selectively sampling from the server model.

\begin{table*}
  \caption{FL Studies of Audio and Speech Processing in Terms of Our Eight-dimension Assessment}
  \label{tb:asp}
  \begin{tabular}{cc cccc cccc c}
    \toprule    App.&Data&Privacy&CommCost&SysHetero&StatHetero&U.D.U.&Simplified&Ser.O.&Cli.O.&Ref\\
    \midrule
    A-Rec & Audio & &  & & \checkmark &    & \checkmark &  & & \cite{gudur2021zero}  \\
    A-Rec & Audio & &  & & \checkmark & \checkmark  &  &  & & \cite{10.1145/3520128}  \\ 
    A-Rec & Audio & &  & &  & \checkmark  &  &  & & \cite{9746356}  \\    
    A-Rec & Audio & &  & & &   &\checkmark  &  & & \cite{10.1007/978-3-031-45392-2_25}  \\     
    A-Rec & Audio & N/A & N/A & N/A& N/A&N/A   &N/A  & N/A &N/A & \cite{10096500}  \\  
    
    ASR & Speech & \checkmark & N/A& N/A&\checkmark & N/A  &N/A & N/A& N/A& \cite{9746541}  \\ 
    ASR & Speech & \checkmark & & & & \checkmark  & & & & \cite{10096426}  \\
    ASR & Speech & \checkmark & & & &   & \checkmark& \checkmark & & \cite{10.1145/3447687}  \\
    
    ASR & Speech & N/A & \checkmark & N/A&\checkmark & N/A  & N/A & N/A & N/A& \cite{9413397}  \\ 
    ASR & Speech &  & \checkmark & &\checkmark &   &  &  & & \cite{9414305}  \\ 
    ASR & Speech &  &\checkmark & & &   & &\checkmark & & \cite{9746226}  \\

    ASR & Speech & &  & &\checkmark &   &  & \checkmark & & \cite{dimitriadis2020federated}  \\     
    ASR & Speech &  & & &\checkmark &   & & \checkmark & & \cite{9747161}  \\ 
    ASR & Speech &  & & &\checkmark &  &\checkmark & &\checkmark & \cite{10175520}  \\
    
    ASR & Speech &  &  & & &   & \checkmark &  & & \cite{10.1007/978-3-030-59419-0_54}  \\ 
    ASR & Speech &  & & & &   & \checkmark& & & \cite{10.1145/3571732}  \\ 
    ASR & Speech &  & & & &   & \checkmark & N/A &N/A & \cite{9413453}  \\ 
    ASR & Speech & N/A & N/A&N/A &N/A &  N/A &N/A & N/A &N/A & \cite{9657499}  \\ 
    
    SE. & Speech & & & &\checkmark &\checkmark  & & \checkmark & & \cite{9632783}  \\
    SE. & Speech &  & & &\checkmark &  & & & & \cite{LIN2023110396}  \\
    
    KWS & Speech &  & \checkmark& & &  & & & & \cite{8683546}  \\
    KWS & Speech &  & \checkmark& &\checkmark & \checkmark & &\checkmark & & \cite{hard2020training}  \\
    KWS & Speech &  & & & \checkmark&\checkmark & & & & \cite{diao2023semi}  \\
    KWS & Speech &  & & &\checkmark & & &\checkmark & & \cite{li2022avoid}  \\
    KWS & Speech &  & & & & \checkmark & &\checkmark & & \cite{hard2022production}  \\
  \bottomrule
\end{tabular}
\justifying
Abbreviations and concept extension in the table: \textbf{App.}: Applications. \textbf{Data}: Raw Data Type. \textbf{Privacy}: Privacy (and Security). \textbf{CommCost}: Communication Cost. \textbf{SysHetero}: System Heterogeneity. \textbf{StatHetero}: Statistical Heterogeneity. \textbf{U.D.U.}: Unlabeled Data Usage. \textbf{Simplified}: Simplified Setup. \textbf{Ser.O.}: Server-optimized Federated Learning. \textbf{Cli.O.}: Client-optimized Federated Learning. \textbf{Ref}: Reference. 
\textbf{A-Rec}: Audio Recognition. 
\textbf{ASR}: (Automatic) Speech Recognition. 
\textbf{SE.}: Speech Enhancement. 
\textbf{KWS}: Keyword Spotting. 
\textbf{\checkmark} notes considerations (with an exception that a \textbf{\checkmark} in Simplified is undesired).
\end{table*}

Next, we present three studies focusing on statistical heterogeneity.
\citet{dimitriadis2020federated} addressed statistical heterogeneity using weighted model averaging, where weights depend on training loss. Their experiments with LibriSpeech \cite{7178964} showed significant training acceleration with a word error rate of 6\%.
\citet{9747161} tackled realistic FL configurations in ASR with varied datasets.
They performed a cross-device ASR study with 4,000 clients and a heterogeneous dataset, using a novel word error rate-based aggregation method for FL-ASR.
Their study represents the first study for realistic FL scenarios on attention-based sequence-to-sequence end-to-end ASR models.
Another case study on model personalization through clustering in FL-ASR was presented in \citet{10175520}, targeting non-IID data in speech for local model customization.

\citet{10.1007/978-3-030-59419-0_54} integrated a basic FL framework with the Kaldi “chain” model and backoff n-gram language model \cite{povey2011kaldi}.
\citet{9657499} applied FedAvg to ASR with Pytorch-Kaldi and ESPnet \cite{watanabe2018espnet}, showing slight improvements but highlighting FedAvg's limitations with complex time series data.
\citet{9413453} proposed a privacy-enhancing feature extraction method using Quantum Convolutional Neural Networks (QCNN), showing stable performance on the Google Speech Command dataset \cite{warden2018speech}. \citet{10.1145/3571732} addressed Code-Switching in Indian languages with FL, effectively managing the integration of words from different languages in informal multilingual dialogues.

In the field of speech enhancement, \citet{9632783} introduced FEDENHANCE, an unsupervised FL method for speech enhancement, handling non-IID data across clients.
They created the new LibriFSD50K dataset and showed that pre-training accelerates convergence and improves performance.
Inspired by \citet{9632783}, \citet{LIN2023110396} proposed a self-adaptive noise distribution network, SASE, to address slow convergence, suboptimal outcomes, and heterogeneous data.
Their approach showed high robustness but the large model size could challenge real-time enhancement on constrained edge devices.

\subsection{Keyword Spotting}
Keyword Spotting (KWS) aims to detect specific keywords in an audio stream, crucial for voice command systems, and is a specialized ASR subset requiring high accuracy with low computational demands \cite{9665775}. 

\citet{8683546} pioneered FL-KWS tasks, using a crowdsourced dataset from 1.8k contributors, reflecting a realistic non-IID and unbalanced scenario.
They analyzed communication costs for feasible training and deployment, noting an 8 MB cost per user, suitable for smart home voice assistants.
\citet{hard2020training} expanded to a production-grade dataset with 1.5M utterances from 8k participants. They introduced speech data augmentation for non-IID training, used teacher-student modeling to address label scarcity, and reduced computational costs by controlling model size.
\citet{hard2022production} then implemented this approach on real user devices, incorporating contextual user-feedback signals to correct semi-supervised learning labels and a hybrid federated-centralized training method.
Their results showed that FL can reduce mistriggers compared to centralized models.

\citet{li2022avoid} addressed data heterogeneity in FL-KWS from the server's perspective, using an adversarial learning strategy to prevent local updates from overfitting and enhance training for users with high-quality data. \citet{diao2023semi} introduced a semi-supervised approach for FL-KWS that uses small amount of labeled server data to enhance performance.
Other studies in audio recognition also used KWS to evaluate various FL frameworks  \cite{gudur2021zero,9746356,10.1145/3520128,10096500}, achieving similar results as discussed before.
Below we present a summary of typical future directions of FL in audio and speech processing.

\section{Well-being}
\label{sec:Well-being}

This section explores the use of FL in outpatient healthcare and general well-being, emphasizing wearable devices and sensor-based technologies for daily health monitoring without clinical intervention.
Ubiquitous computing with ML shows potential in both mental \cite{10.1145/3398069} and physical healthcare \cite{9153891}, and FL can address privacy concerns. We discuss contributions to emotion recognition, depression, stress, and cognitive load, as well as advancements in physical health monitoring, summarizing in Tables \ref{tb:Well-being} and \ref{tb:Well-beingC}.

\begin{table*}
  \caption{FL Studies of Well-being in Terms of Our Eight-dimension Assessment}
  \label{tb:Well-being}
  \begin{tabular}{cc cccc cccc c}
    \toprule    App.&Data&Privacy&CommCost&SysHetero&StatHetero&U.D.U.&Simplified&Ser.O.&Cli.O.&Ref\\
    \midrule
     SER& Speech &\checkmark &N/A &N/A & N/A & N/A   & N/A & N/A &N/A & \cite{feng2021attribute}  \\
     SER& Speech &\checkmark &N/A &N/A & N/A &  N/A  &N/A  & N/A &N/A & \cite{feng2022user}  \\
     SER& Speech &\checkmark &  & &  &    &  &  & & \cite{10196918}  \\
     SER& Speech &\checkmark & & &  &    &  & \checkmark & & \cite{10306049}  \\
     SER& Speech &\checkmark & & &  &    &  & \checkmark & & \cite{chang2022robust}  \\
     SER& Speech &\checkmark & & &  &    &  & \checkmark & & \cite{10095737}  \\
     SER& Speech &\checkmark & \checkmark & &  &    &  & \checkmark & & \cite{mohammadi2023secure}  \\
     
     SER& Speech & &  & & \checkmark & \checkmark   &  & \checkmark & & \cite{9767445}  \\
     SER& Speech & &  & & \checkmark & \checkmark    &  & \checkmark & & \cite{feng2022semi}  \\
     SER& Speech & & & & &    &  & \checkmark & & \cite{singh2022privacy}  \\
     
     S,FER& Multi & & & & &    &  & \checkmark & & \cite{9253631}  \\
     FER& Video &\checkmark &  & &\checkmark  &    &  &  &\checkmark & \cite{10.1145/3503161.3548278}  \\
     FER& Multi & &\checkmark  & &  &\checkmark    &  & \checkmark &\checkmark & \cite{10.1145/3536221.3556614}  \\
     FER& Multi & &  & & \checkmark &    &  & \checkmark & & \cite{9894363}  \\
     FER& Image & & & & & \checkmark   &  & \checkmark & & \cite{Shome_2021_ICCV}  \\
     FER& Image & & & &  &    & \checkmark &  & & \cite{9942417}  \\
     FER& Image & & & &  &    & \checkmark & & & \cite{9765165}  \\
     FER& Image & &  & &  &    & \checkmark  & & & \cite{10.1145/3592686.3592715}  \\
     
     BER& Physio & & \checkmark & &  &    &  &  & & \cite{nandi2022federated}  \\
     BER& Physio & &  & &  &    &  &  & & \cite{gahlan2023federated}  \\
     BER& Physio & &  & &  &    & \checkmark &  & & \cite{10041308}  \\
     BER& Physio & &  & &  &    &\checkmark &  & & \cite{10112028}  \\

     Dep.& Text & & \checkmark & &  &    &  &  & & \cite{9724416}  \\
     Dep.& Text & &  & &\checkmark  &    &  & \checkmark  & & \cite{xu2021fedmood}  \\
     Dep.& Text & &  & &\checkmark  &    &  & \checkmark & & \cite{9540999}  \\
     Dep.& Text & &  & &  & \checkmark   & \checkmark &  & & \cite{9767700}  \\
     Dep.& Text & &  & &  &    & \checkmark &  & & \cite{9885430}  \\
     Dep.& Multi & &  & &  &    & \checkmark &  & & \cite{9978600}  \\
     Dep.& Speech & &  & &  &    & \checkmark &  & & \cite{9746827}  \\
     Dep.& Speech & \checkmark &  & & \checkmark &    &  &  & & \cite{9871861}  \\

     St.& Physio & & \checkmark & & \checkmark &    &  & \checkmark & \checkmark & \cite{10197159}  \\
     St.& Physio & &  & &  &    &  & \checkmark & & \cite{10.1145/3428152}  \\
     St.& Physio & &  & &  &    &  &  & \checkmark & \cite{pham2023personalized}  \\
     St.& Physio & &  & &  &    &  &  & \checkmark & \cite{rafi2022personalization}  \\
     St.& Physio & &  & &  &    &  &  & & \cite{vyas2023federated}  \\
     St.& Physio & &  & &  &    & \checkmark &  & & \cite{almadhor2023wrist}  \\
     St.& Physio &N/A & N/A &N/A &N/A  &N/A    &N/A  &N/A  &N/A & \cite{fauzi2022comparative}  \\

  \bottomrule
\end{tabular}
\justifying
Abbreviations and concept extension in the table: \textbf{App.}: Applications. \textbf{Data}: Raw Data Type. \textbf{Privacy}: Privacy (and Security). \textbf{CommCost}: Communication Cost. \textbf{SysHetero}: System Heterogeneity. \textbf{StatHetero}: Statistical Heterogeneity. \textbf{U.D.U.}: Unlabeled Data Usage. \textbf{Simplified}: Simplified Setup. \textbf{Ser.O.}: Server-optimized Federated Learning. \textbf{Cli.O.}: Client-optimized Federated Learning. \textbf{Ref}: Reference. 
\textbf{SER}: Speech Emotion Recognition.
\textbf{FER}: Facial Expression Recognition.
\textbf{BER}: Biometric Emotion Recognition.
\textbf{Physio}: Physiological Signal.
\textbf{Dep.}: Depression.
\textbf{St.}: Stress.
\textbf{\checkmark} notes considerations (with an exception that a \textbf{\checkmark} in Simplified is undesired).
\end{table*}

\begin{table*}
  \caption{FL Studies of Well-being in Terms of Our Eight-dimension Assessment (continued)}
  \label{tb:Well-beingC}
  \begin{tabular}{cc cccc cccc c}
    \toprule    App.&Data&Privacy&CommCost&SysHetero&StatHetero&U.D.U.&Simplified&Ser.O.&Cli.O.&Ref\\
    \midrule
     Lone& Multi & &  & &  &    &  &  & & \cite{9861090}  \\
     
     Cog.& Physio & &  & &  &    &  &  & \checkmark & \cite{10.1145/3626705.3631796}  \\
     Cog.& Physio & &  & &  &    &  &  & & \cite{10.1145/3626705.3627783}  \\
     P.H.& Physio & &  & &  &    &  &  & & \cite{fang2021bayesian}  \\
     P.H.& Physio & &  & &  &    &  &  & & \cite{9630505}  \\
     P.H.& Physio & &  & &  &    &  &  & & \cite{10.1145/3512731.3534207}  \\
     P.H.& Inertia & &  & &  &    &  &  & & \cite{9669395}  \\
     P.H.& Inertia &  &  & &  &    &  &  & & \cite{9984667}  \\
  \bottomrule
\end{tabular}
\justifying
Abbreviations and concept extension in the table: \textbf{App.}: Applications. \textbf{Data}: Raw Data Type. \textbf{Privacy}: Privacy (and Security). \textbf{CommCost}: Communication Cost. \textbf{SysHetero}: System Heterogeneity. \textbf{StatHetero}: Statistical Heterogeneity. \textbf{U.D.U.}: Unlabeled Data Usage. \textbf{Simplified}: Simplified Setup. \textbf{Ser.O.}: Server-optimized Federated Learning. \textbf{Cli.O.}: Client-optimized Federated Learning. \textbf{Ref}: Reference. 
\textbf{Lone}: Loneliness.
\textbf{Cog.}: Cognitive Load.
\textbf{P.H.}: Physical Health.
\textbf{Physio}: Physiological Signal.
\textbf{\checkmark} notes considerations (with an exception that a \textbf{\checkmark} in Simplified is undesired).
\vspace{-10pt}
\end{table*}

\subsection{Emotion Recognition}
This survey focuses on three types of ER: Speech Emotion Recognition (SER), Facial Expression Recognition (FER), and Biometric (Physiological-based) Emotion Recognition (BER).

SER's goal is to detect emotional states from speech for further affective tasks. Many studies simplify emotional states into basic categories \cite{ekman2013emotion}: Anger, Disgust, Fear, Joy, Sadness, and Surprise, as benchmarks for research, with modeling approaches varying from statistics \cite{el2011survey} to DL \cite{lieskovska2021review}.

Studies of FL-SER focused on enhanced security and privacy.
Feng et al. significantly contributed to FL-SER with several studies \cite{feng2021attribute, feng2022user, feng2022semi}.
They highlighted the vulnerability of FL privacy through a gender inference attack \cite{feng2021attribute}. They found a serious risk of privacy leakage in standard FL where we have access to model parameters and updates. Their analysis revealed that the initial layers of a neural network are most susceptible to information leakage. In response, they evaluated User-Level Differential Privacy (UDP) \cite{feng2022user}, finding it only partially effective against such attacks under conditions of minimal leakage. Additionally, they introduced Semi-FedSER \cite{feng2022semi}, a semi-supervised FL framework for SER that uses pseudo-labeling to expand the training dataset.
Mohammadi et al. also advanced privacy for FL-SER through key publications \cite{10196918,mohammadi2023secure,10306049}. They introduced Optimized Paillier Homomorphic Encryption (OPHE) \cite{10196918} to improve client confidentiality in FL, suitable for use on resource-limited edge devices through pruning. Their subsequent work developed SEFL, a Secure and Efficient FL framework for SER \cite{mohammadi2023secure}, addressing challenges posed by DP mechanisms, which degrade model accuracy due to noise addition in speech data. They later proposed using Local Differential Privacy (LDP) combined with a novel client selection strategy \cite{10306049}.
\citet{chang2022robust} investigated adversarial attacks in SER and developed a novel FL framework incorporating adversarial training and randomization to enhance model robustness. This approach outperforms standard FL by resisting both single-step attacks \cite{kurakin2016adversarial} and more complex iterative attacks \cite{madry2017towards}.
\citet{10095737} revisited attribute inference attack and introduced a novel privacy-enhanced FL model employing a self-attention mechanism, focusing on utterances with significant emotional content while sidestepping irrelevant sensitive features.
\citet{9767445} utilized attention mechanisms and self-training to enhance representation with large amounts of unlabeled data, showing competitive results with models that use non-IID and minimally labeled datasets.
A comparison of SER between FL and centralized training \cite{singh2022privacy} revealed a 7\% performance drop in FL. The FedAudio framework \cite{10096500} also benchmarked SER with similar findings.

Considering facial expression, \citet{9253631} combined facial images and speech signals using ensemble classifiers and CNN to predict emotional fluctuations. EmoFed \cite{10.1145/3503161.3548278} addressed facial data heterogeneity and privacy concerns with culturally similar clustering and encrypted personalization. \citet{10.1145/3536221.3556614} developed a lightweight FL-FER model with MobileNetV2, maintaining performance with reduced parameters. FedAffect \cite{Shome_2021_ICCV} used few-shot learning on unlabeled data and outperformed traditional centralized methods on the FERG dataset \cite{aneja2017modeling}.
\citet{9942417} proposed a Split FL framework for FER, allowing clients to upload only specific segments of the model, enhancing privacy and efficiency. \citet{9765165} proposed FedNet and suggested suitable neural network structures to reduce overfitting and stabilize the training process. \citet{10.1145/3592686.3592715} emphasized the importance of FL-FER in online teaching, analyzing the facial expressions of students and teachers to improve teaching while protecting privacy. \citet{9894363} introduced AG2M, a multi-domain graph autoencoder Gaussian mixture model for environmental changes.

Considering physiological signals, \citet{nandi2022federated} introduced Fed-ReMECS, a multimodal FL method employing feature fusion from various sources such as electrodermal activity (EDA) and respiration (RESP).
\citet{10041308,10112028} explored EEG-based emotion recognition with the DREAMER \cite{7887697} datasets, using Russell's two-dimensional model \cite{lang1995emotion} of affect. \citet{gahlan2023federated} proposed an F-MERS multimodal framework, integrating EEG, EDA, Electrocardiogram (ECG), and RESP, enhancing performance through data fusion techniques.

\subsection{Mental Health}
In this subsection, we first introduce the FL studies using text and speech signals to detect depression, and then studies using physiological signals to detect stress, with one study addressing loneliness.
Finally, we include two studies on the cognitive load that employed FL mechanisms.

\citet{xu2021fedmood,9540999} introduced FedMood, an FL framework using DeepMood \cite{10.1145/3097983.3098086} to predict mood disorders from mobile typing dynamics and accelerometer data, based on the Hamilton Depression Rating Scale (HDRS) \cite{hamilton1986hamilton} and the Young Man Mania Scale (YMRS) \cite{young1978rating} questionnaires. \citet{9767700} used hyper-graph and attention-based models for emotion detection from word embeddings, enabling the extensive use of unlabeled text data for generalization without annotation. \citet{9724416} proposed a Text-CNN Asynchronous Federated optimization method, CAFed, that reduces communication overhead. \citet{9885430} proposed an alert system for depression. 
\citet{9978600} designed a mobile app for depression, with a 6-week study showing reduced power consumption and internet usage.
Speech analysis for depression assessment was also explored by \citet{9746827} using FedAvg and FedMA, although it showed reduced accuracy in binary classification.
\citet{9871861} integrated FL defenses into a speech-based diagnosis system, maintaining performance with approaches like Norm Bordering, Differential Privacy, and Secure Aggregation.

Considering stress recognition, \citet{10.1145/3428152} introduced an FL framework based on wearable devices and photoplethysmography (PPG) data. \citet{fauzi2022comparative} evaluated smartwatch-based stress detection on the WESAD dataset \cite{10.1145/3242969.3242985}, revealing a performance gap between FL and centralized approaches and highlighting the need for optimization for FL. \citet{almadhor2023wrist} enhanced performance on the WESAD dataset by implementing the Synthetic Minority Oversampling Technique \cite{chawla2002smote}. \citet{10197159} applied heuristic clustering and transfer learning for personalization, reducing computational overhead. \citet{pham2023personalized} used personalized FL for driving stress detection on the AffectiveROAD dataset \cite{10.1145/3167132.3167395}. \citet{rafi2022personalization} employed a Support Vector Machine (SVM) model for personalization in the stress estimation task. \citet{vyas2023federated} proposed FedSafe to predict driver stress and behavior with up to 98\% accuracy, reducing communication overhead by 25 times compared to traditional ML systems.

Addressing loneliness, \citet{9861090} developed an FL framework using the StudentLife dataset \cite{10.1145/2632048.2632054}, which includes data from various sensors. Despite performance differences with centralized methods, this study is a unique approach to managing loneliness through FL.
Additionally, we highlight two papers addressing cognitive workload \cite{10.1145/3626705.3627783,10.1145/3626705.3631796}. 
\citet{10.1145/3626705.3627783} quantitatively evaluate cognitive workload, with eye-tracking and a collection of physiological data. \citet{10.1145/3626705.3631796} further applied an unsupervised client personalization strategy to accommodate unseen clients. FL showed similar results compared to a centralized setting in both studies.

\subsection{Physical Health}

Although there are numerous studies on FL in physical health, many were outside the scope of our survey because they focused primarily on clinical settings. Readers interested in these excluded studies may refer to reviews specializing in the Internet of Medical Things (IOMT) in Section \ref{sec:bg}. Instead, we focus on studies on physical health in everyday contexts, including heart rate prediction for exercise, walking surface identification, gestational weight gain prediction, fall detection, and a framework for sleep quality prediction.

\citet{fang2021bayesian} developed two Bayesian FL frameworks to enhance training efficiency and safety.
These frameworks use heart rate data to provide real-time coaching guidance. 
The models achieved comparable accuracy to centralized methods.
\citet{9669395} created an FL approach for identifying walking surfaces using data from 6 inertial measurement units (IMUs) at different body locations, resulting in up to 85\% accuracy in the final FL model.
\citet{9630505} introduced an FL approach for predicting gestational weight gain to mitigate health risks during pregnancy. The study involved 80 women, incorporating data like age and height, achieving state-of-the-art results with a simple FL model. \citet{9984667} introduced the Fed-ELM framework, an FL system for fall detection that uses an extreme learning machine \cite{HUANG2006489} to achieve high accuracy in experimental settings, aiming to reduce the time to medical assistance and minimize injury risk.
\citet{10.1145/3512731.3534207} employed FL with multiple CNNs (FedMCRNN) and multimodal data from wearables for sleep quality prediction. The study collected 16 different physiological signals from 16 participants, identifying critical external factors influencing sleep quality. 

\section{User Identification}
\label{sec:ii}

User Identification is closely linked to Human Sensing and the integration of HCI systems. However, developing ML-based identification systems poses a paradox: while enhancing security against unauthorized access, it raises privacy concerns from data collection. This section explores FL applications in User Identification, focusing on person re-identification, speaker identification, and face recognition, including both authentication and proactive recognition. We summarized all studies in Table \ref{tb:ii}.

\subsection{Person Re-identification}

Person re-identification (ReID) is a task in computer vision aimed at identifying and tracking an individual across different times and locations from a large image gallery \cite{zheng2016person}.
It involves three main modules: person detection, tracking, and person retrieval.
Key concerns in ReID research are data availability, data quality, and data privacy due to sensitive information in video recordings.

We first present studies on reducing computational costs in FL-ReID \cite{10142016,9590400,GIRIJA2023100793,9949249,10.1145/3474085.3475182}.
\citet{10142016} integrated FL with lifelong learning \cite{parisi2019continual} in ReID, addressing domain drift and emphasizing spatial-temporal correlations. Their framework improves incremental knowledge acquisition and communication efficiency. \citet{9590400} proposed ReID approach for unmanned aerial vehicle (UAV) delivery services, using a tailored dataset and knowledge distillation to reduce model size and computational costs, enhancing real-time inference. \citet{10.1145/3474085.3475182} introduced FedUReID, leveraging vast unlabeled data on edge devices. Their feature encoding and clustering method, along with a joint cloud-edge optimization strategy, addressed statistical heterogeneity and improved performance with reduced computation costs by 29\%. \citet{9949249} presented a federated unsupervised ReID approach using camera-aware clustering for pseudo-label generation and contrastive learning. It demonstrated superior performance while reducing communication costs by 50\%. \citet{GIRIJA2023100793} developed FedTransferLoss, a federated aggregation algorithm with transfer learning. They highlighted the benefits of decentralizing tasks to edge devices to reduce cloud server load and communication costs.

Next, we introduce studies focused on heterogeneity.
\citet{10248272} were among the first to address system heterogeneity in FL-ReID, targeting edge devices with limited resources like drones and phones.
They used trustworthy edge servers to identify and manage heterogeneous devices, employing a local Deep Convolutional Generative Adversarial Network for global data augmentation to mitigate non-IID issues.
\citet{10.1145/3394171.3413814} introduced FeDReID, establishing a benchmark for non-IID data and real-world applications. They proposed solutions like knowledge distillation and dynamic weight adjustment, outlining scenarios for federated-by-camera and federated-by-dataset. Further work \cite{10.1145/3531013} extended this with clustering enhancements. \citet{WENG2023103831} advanced FL-ReID using unsupervised learning, addressing the challenge of a universal stopping criterion in FedUReID \cite{10.1145/3474085.3475182}.
Their Federated Unsupervised Cluster-Contrastive Learning (FedUCC) method showed notable improvements, refining learning from coarse to fine granularity. \citet{10.1145/3581783.3612350} also tackled label shifting and device heterogeneity. They argued that \citet{10.1145/3394171.3413814,10.1145/3531013} provoke conflicts between global aggregation and local training, which degrades performance in unseen domains.
They introduced FFReID, enhancing local training with a proximal and feature regularization term, and used feature-aware weights for effective global aggregation. \citet{yang2022federated} addressed domain generalization challenges in ReID, proposing the Domain and Feature Hallucinating (DFH) strategy for generating new features, significantly enhancing over methods like FedPav \cite{10.1145/3394171.3413814} and FedReID \cite{Wu_Gong_2021}, another framework designed to adapt to unseen domains.

\subsection{Speaker Identification}
Speaker identification, including recognition, verification, and authentication, uses voice characteristics to identify individuals \cite{TIRUMALA2017250}.
Voice attributes can reveal sensitive information like gender, age, and language. The presented studies involve FL methods to improve model robustness and privacy protection, addressing domain shifts and various model and data attacks.

\citet{granqvist2020improving} explored FL in speaker identification by having local clients train an auxiliary classification model to predict vocal characteristics as side information.
This information enhances the main speaker verification system by using side information in multi-task learning, avoiding labor-intensive labeling and privacy concerns. \citet{chen2023learning} addressed domain shifts by customizing local models to manage diverse data characteristics, such as language variations and background noise. They also utilize continual learning to improve model robustness. \citet{10.1007/978-3-030-95391-1_29} proposed an FL framework that hides sensitive attributes in raw speech data, addressing data scarcity and enhancing privacy protection. This method overcomes limitations in prior works like \cite{granqvist2020improving}, which safeguard side information but not the speech signal itself.
The framework in \cite{9746541} also addresses speaker identification, yielding results comparable to the ASR task. \citet{9592761} proposed using generative adversarial networks (GANs) to generate impostor speech data on edge devices, differing from the approach in \cite{granqvist2020improving}. This method aims to enhance model resilience against impostor attacks by training with impostor data.  \citet{10094675} highlighted vulnerabilities in FL systems, and introduced Personalized Federated Aggregation and the Global Spectral Cluster method to limit attacker influence. 



\subsection{Face Recognition}
Face recognition (FR) involves identifying individuals from images or videos, a prominent area of research in computer vision that has evolved significantly with DL, known as Deep face recognition \cite{WANG2021215}.
The process involves detecting and aligning faces into coordinates, followed by image augmentation or normalization.
FR is crucial in HCI for tasks like authentication. \citet{WANG2021215} highlighted that FR requires processing vast datasets and maintaining extensive parameters, making it data-intensive.
Real-time processing is also essential for user experience but is challenged by two-dimensional data.

\begin{table*}
  \caption{FL Studies of User Identification in Terms of Our Eight-dimension Assessment}
  \label{tb:ii}
  \begin{tabular}{cc cccc cccc c}
    \toprule    App.&Data&Privacy&CommCost&SysHetero&StatHetero&U.D.U.&Simplified&Ser.O.&Cli.O.&Ref\\
    \midrule
    ReID. & Image &  & \checkmark & & \checkmark &\checkmark &  & \checkmark & \checkmark& \cite{10.1145/3474085.3475182}  \\
    ReID. & Image &  & \checkmark & & \checkmark &\checkmark &  & \checkmark & \checkmark& \cite{9949249}  \\
    ReID. & Image &  & \checkmark & & \checkmark & &  & \checkmark &\checkmark& \cite{10142016}  \\
    ReID. & Image &  & \checkmark & & \checkmark & &  & \checkmark & & \cite{9590400}  \\
    ReID. & Image &  & \checkmark & &  & &  & &\checkmark & \cite{GIRIJA2023100793}  \\
    
    ReID. & Image &  &  & \checkmark   & \checkmark &\checkmark &  & \checkmark & & \cite{10248272}  \\
    ReID. & Image &  &  & & \checkmark & \checkmark &  & \checkmark & \checkmark& \cite{10.1145/3531013}  \\
    ReID. & Image &  &  & & \checkmark & \checkmark &  &\checkmark &\checkmark& \cite{WENG2023103831}  \\
    ReID. & Image &  &  & & \checkmark & &  & \checkmark &\checkmark & \cite{10.1145/3581783.3612350}  \\
    ReID. & Image &  &  & & \checkmark & \checkmark   &  & \checkmark & & \cite{10.1145/3394171.3413814}  \\
    ReID. & Image &  &  & & \checkmark & &  & \checkmark & & \cite{yang2022federated}  \\
    
    ReID. & Image &  &  & &  & &  & \checkmark & & \cite{Wu_Gong_2021}  \\

    SID. & Speech & \checkmark &  & & \checkmark & &  &\checkmark && \cite{granqvist2020improving}  \\
    SID. & Speech & \checkmark &  & &  & &  & && \cite{10.1007/978-3-030-95391-1_29}  \\

    SID. & Speech & \checkmark &N/A  &N/A & N/A & N/A& N/A &N/A &N/A& \cite{10094675}  \\
    SID. & Speech & &  & & \checkmark & &  &\checkmark &\checkmark& \cite{chen2023learning}  \\
    SID. & Speech &  &  & &  &\checkmark &  &\checkmark && \cite{9592761}  \\

    FR & Multi & \checkmark & N/A & N/A& N/A  &N/A & N/A &N/A &N/A& \cite{9780603}  \\  
    FR & Image & \checkmark & & & \checkmark & &  \checkmark& && \cite{10185525}  \\
    FR & Image & \checkmark & & &  & &  & && \cite{9886171}  \\
    FR & Image & \checkmark & \checkmark & &   & &  \checkmark &\checkmark && \cite{meng2022improving}  \\  
    
    FR & Image &  &\checkmark & & &\checkmark & \checkmark &\checkmark && \cite{zhuang2021towards}  \\
    FR & Image &  &\checkmark & & &\checkmark & \checkmark &\checkmark && \cite{9859587}  \\
    FR & Image &  &\checkmark& &  & &  &\checkmark && \cite{9482149}  \\
    
    FR & Image &  &  & & \checkmark & & \checkmark &\checkmark && \cite{bai2021federated}  \\
    FR & Image &  & & & \checkmark & &  &\checkmark &\checkmark& \cite{liu2022fedfr}  \\
    FR & Image &  & & &\checkmark  & &  &\checkmark && \cite{niu2022federated}   \\
    FR & Image &  & & & \checkmark & &  & && \cite{9484386}  \\
    
    FR & Video &  & & &  & &  \checkmark& && \cite{10.1145/3501814}  \\
    FR & Multi &  & & &  & &  \checkmark& && \cite{10177776}  \\
    FR & Image &  & & &  & &  \checkmark& && \cite{9484338}  \\
    ID. & Physio &  & & &  & &  \checkmark& && \cite{10301542}  \\
  \bottomrule
\end{tabular}
\justifying
Abbreviations and concept extension in the table: \textbf{App.}: Applications. \textbf{Data}: Raw Data Type. \textbf{Privacy}: Privacy (and Security). \textbf{CommCost}: Communication Cost. \textbf{SysHetero}: System Heterogeneity. \textbf{StatHetero}: Statistical Heterogeneity. \textbf{U.D.U.}: Unlabeled Data Usage. \textbf{Simplified}: Simplified Setup. \textbf{Ser.O.}: Server-optimized Federated Learning. \textbf{Cli.O.}: Client-optimized Federated Learning. \textbf{Ref}: Reference. 
\textbf{ReID.}: Person Re-identification.  
\textbf{SID.}: Speaker Identification. 
\textbf{FR}: Face Recognition. 
\textbf{ID.}: Identification. 
\textbf{Physio}: Physiological Signal. 
\textbf{\checkmark} notes considerations (with an exception that a \textbf{\checkmark} in Simplified is undesired).
\end{table*}

First, we introduce several studies on privacy attacks.
\citet{9780603} focused on face presentation attack detection (fPAD), known as anti-spoofing, a technique that identifies non-human presentations like videos and photos. They introduced FedPAD and FedGPAD, with the latter showing better generalization. \citet{10185525} investigated adversarial attacks in FR systems in smart homes and proposed a method to address FL’s vulnerability to malicious adversarial examples. 
\citet{9886171} introduced an FL-FR framework using blockchain, replacing centralized servers with a decentralized training approach. This method reduces data processing and storage demands on a central server while enhancing security against malicious nodes.
\citet{meng2022improving} argued that class embeddings are crucial for FR, but broadcasting class centers among clients improves performance at the cost of privacy. Their Differentially Private Local Clustering (DPLC) approach generates privacy-agnostic clusters where individual membership is unknown, improving privacy guarantees. 

Concerning communication cost, \citet{zhuang2021towards,9859587} proposed unsupervised domain adaptation framework that has three stages: pre-training with source domain data, pseudo-labeling for target domain data, and a final FL training. The author analyzed the communication cost and advocated for additional local iterations. \citet{9482149} introduced an evolutionary-based FL method to handle inconsistent client participation, balancing personalization and generality while mitigating communication delays.

Next, we introduce four studies on statistical heterogeneity. \citet{bai2021federated} pioneered FL-FR with the FedFace framework, including Partially Federated Momentum (PFM) and Federated Validation (FV) algorithms.
PFM addresses client drift by incorporating global gradient statistics into local training, while FV enhances validation data privacy. Another FedFace framework was introduced by \citet{9484386}, which improves class embeddings by integrating additional face data from edge nodes, using a mean feature initialization scheme and spread-out regularization. \citet{liu2022fedfr} proposed a framework targeting joint server training with local personalization. They incorporated shared class embedding transmission to mitigate local overfitting and used a hard negative sampling strategy to reduce computational overhead.
\citet{niu2022federated} addressed the FL non-IID challenge with a theoretical focus. They observed a tendency for orthogonal class embeddings due to uncomplicated local tasks and overfitting with limited data.
To counter this, they introduced FedGC, combining local and cross-client optimization through a softmax regularization, achieving high effectiveness.

Two studies \cite{10.1145/3501814,10177776} explore face forgery detection in FL-FR.
Face forgery involves altering facial features in images or videos, posing significant security concerns.
Both works present privacy-friendly FL methods for detecting face forgeries, showing promising results.
However, their primary innovation lies in vision techniques, with the FL aspect being less developed by limiting the number of clients.
\citet{9484338} focus on user active authentication using facial images, addressing the challenges of traditional systems that rely on one-class classification due to the absence of negative class data.
They propose the Federated Active Authentication (FAA) framework, which uses distributed data across edge devices to improve training effectiveness.
In contrast, only one paper \cite{10301542} explores the use of vital signs for FL personal identification.
This study employs ECG and radar signals for identification, noting that while ECG necessitates physical contact with devices, radar sensors offer a contact-free alternative. 

\section{Human Mobility and Localization}
\label{sec:hml}
Ubiquitous computing and the IoT industry have advanced research in human mobility and localization, generating vast amounts of sensitive geospatial data. Despite privacy and legal concerns, this area has attracted significant research interest due to its potential benefits.
This section is divided into indoor localization and human mobility in outdoor environments.
Both track human location and movement but differ in context, data usage, and applications.
Indoor localization pinpoints a user’s location within buildings \cite{8692423}, while human mobility involves broader movement patterns typically tracked via GPS \cite{10.1145/3485125}. Current systems often prioritize performance over privacy, highlighting the need for trust between users and service providers to ensure data security.
Table \ref{tb:hml} presents all relevant studies in this field. The subsequent subsections offer a detailed examination of each study.




\begin{table*}
  \caption{FL Studies of Human Mobility and Localization in Terms of Our Eight-dimension Assessment}
  \label{tb:hml}
  \begin{tabular}{cc cccc cccc c}
    \toprule    App.&Data&Privacy&CommCost&SysHetero&StatHetero&U.D.U.&Simplified&Ser.O.&Cli.O.&Ref\\
    \midrule
     IL.& C/R & \checkmark&  & & \checkmark &    &  & \checkmark &\checkmark & \cite{9658722}  \\
     IL.& C/R & & \checkmark & & \checkmark &  \checkmark  &  & \checkmark & & \cite{9917443} \\  
     IL.& C/R & & \checkmark & & \checkmark &    &  & \checkmark & & \cite{9764747}  \\
     IL.& C/R & & \checkmark & &  &    &  & \checkmark & & \cite{etiabi2022federated} \\
     IL.& C/R & & \checkmark & \checkmark&  \checkmark&    &  & \checkmark & & \cite{10.1145/3607919} \\
     IL.& C/R & &  &\checkmark &\checkmark  &    &  & \checkmark & \checkmark& \cite{10214616} \\
     IL.& C/R & &  & & \checkmark &  \checkmark  &  & \checkmark &\checkmark & \cite{9749277}  \\
     IL.& C/R & &  & & \checkmark &  \checkmark  & &  & & \cite{9103044}  \\
     IL.& C/R & &  & & \checkmark &    &  &  & & \cite{9148111}  \\
     IL.& C/R & &  & & \checkmark &    &  &  & & \cite{9761235}  \\
     IL.& C/R & &  & & \checkmark &    &  & \checkmark &\checkmark & \cite{9734052} \\
     IL.& C/R & &  & & \checkmark &    &  &  &\checkmark & \cite{9593115}  \\
     IL.& C/R &  &  & &  &    &  &  & \checkmark& \cite{10005038} \\
     IL.& C/R & &  & &  &    &  &  & & \cite{9960135} \\
     IL.& C/R & &  & &  &    &  &  & & \cite{10118848} \\
     IL.& C/R & &  & &  &    &  \checkmark&  & & \cite{9066152}  \\
     
     HM& GPS & \checkmark &  & &  &    &  & \checkmark & \checkmark& \cite{10.1145/3381006} \\
     HM& GPS & \checkmark &  &  &  &    &  &  & & \cite{gjoreski2022toward} \\ 
     HM& GPS &  &  & \checkmark & \checkmark &    &  &  & \checkmark& \cite{10.1145/3369830} \\
     HM& GPS &  &  & \checkmark & \checkmark &    &  &  & & \cite{10274418} \\
     HM& GPS &  &  &  & \checkmark &    &  & \checkmark & & \cite{10074979} \\
     HM& GPS &  &  & &  &    &  &  & \checkmark& \cite{10.1145/3397536.3422270} \\     
     HM& GPS &  &  &  &  &    &  &  & & \cite{9671447} \\
     HM& GPS &  &  &  &  &    &  &  & & \cite{errounda2022mobility} \\    
     HM& GPS &  &  &  &  &    &\checkmark  &  & & \cite{ezequiel2022federated} \\    
     HM& GPS &  &  &  &  &    & \checkmark &  && \cite{10.1145/3487351.3490964} \\
     HM& GPS &  &  &  &  &    & \checkmark &     & & \cite{10257271} \\
     HM& GPS & N/A & N/A & N/A & N/A &  N/A  & N/A & N/A & N/A& \cite{9250516} \\
     
     MODE& GPS &  &  &  & \checkmark & \checkmark   &  & \checkmark & & \cite{9632695} \\
     MODE& GPS &  &  &  & \checkmark & \checkmark   &  & \checkmark & & \cite{9514368} \\
     MODE& GPS &  &  &  & \checkmark &    &  &  & & \cite{9359187} \\
     MODE& GPS &  &  &  &  &    &  &  & & \cite{9922022} \\
  \bottomrule
\end{tabular}
\justifying
Abbreviations and concept extension in the table: \textbf{App.}: Applications. \textbf{Data}: Raw Data Type. \textbf{Privacy}: Privacy (and Security). \textbf{CommCost}: Communication Cost. \textbf{SysHetero}: System Heterogeneity. \textbf{StatHetero}: Statistical Heterogeneity. \textbf{U.D.U.}: Unlabeled Data Usage. \textbf{Simplified}: Simplified Setup. \textbf{Ser.O.}: Server-optimized Federated Learning. \textbf{Cli.O.}: Client-optimized Federated Learning. \textbf{Ref}: Reference. 
\textbf{IL.}: Indoor Localization.
\textbf{C/R}: Channel State Information/Received Signal Strength (Indicator).
\textbf{HM}: Human Mobility.
\textbf{MODE}: Travel Mode.
\textbf{\checkmark} notes considerations (with an exception that a \textbf{\checkmark} in Simplified is undesired).
\end{table*}


\subsection{Indoor Localization}

We first introduce \cite{9658722} as an example study using DP to protect user privacy. To address the dynamic nature of data due to random user participation, they introduced a framework that facilitates training without complete data preparation and incorporates personalization to balance global and local models. The framework achieved competitive accuracy with DP for further privacy protection.
Subsequently, the team extended their work with semi-supervised learning to exploit unlabeled data.

Regarding communication costs, \citet{10118848} addressed complex multi-building, multi-floor mapping using FL approach that simultaneously predicts building, floor, and 2D coordinates, which, while integral to indoor localization, incurs high communication costs. A subsequent study \citet{9917443} addressed communication cost using pre-training on the FL server. They also incorporated unlabeled data augmentation, addressing network instability and data heterogeneity.
\citet{9764747} proposed client selection based on reliability, measured as model uncertainty, to mitigate global degradation due to heterogeneous client updates. They addressed the computational challenges of Bayesian models for uncertainty quantification by employing Monte Carlo dropout.
\citet{etiabi2022federated} shared model outputs with knowledge and lower dimensions instead of sharing FL models, and extended the application of state-of-the-art FL distillation techniques, previously tested only on classification tasks, to regression tasks in localization.

Regarding device and statistical heterogeneity, \citet{10214616} used reinforcement learning for dynamic environments and user heterogeneity in indoor positioning. It supports various device types and employs a few-shot learning paradigm for accommodating new users. \citet{10.1145/3607919} addressed device heterogeneity with an embedded ML framework. They evaluated it using a dataset from six smartphone manufacturers, considering noise, scalability, and data skewness. \citet{9103044} argued collecting labeled RSS fingerprint data is costly and location-specific, and proposed a centralized indoor localization method using pseudo-label (CRNP) to utilize unlabeled data via mobile crowdsourcing, improving localization accuracy while maintaining privacy. \citet{9148111} developed a crowdsourcing method for RSS fingerprint-based indoor localization and simulated three scenarios of statistical heterogeneity, showing FL reduces mean absolute error by up to 1.8 meters in a 390 m x 270 m area. \citet{9761235} introduced a convex hull of sampling positions as a computable characteristic to represent local data to confront data heterogeneity.
\citet{9734052} proposed a multi-level Federated Graph Learning framework for location prediction and self-attention for feature extraction. \citet{9593115} enhanced FedAMP \cite{huang2021personalized} by integrating a Bayesian fusion strategy for global aggregation, significantly improving both server and client performance through tailored model fusion and similarity-based clustering.


Other FL studies for indoor localization include \citet{10005038} proposing FedPos, which uses CSI to enhance position estimation, \citet{9960135} introducing a zone-based approach to simplify distance error calculation and enhance privacy, and \citet{9066152} presenting FLoc, which updates fingerprints while preserving privacy.

\subsection{Human Mobility}
This subsection focuses on GPS-based FL human mobility studies, covering FL-based trajectory and location prediction and traveling mode recognition.
For related applications like Place of Interest Recommendation or traffic prediction, we suggest reading other surveys like \citet{belal2023survey}.

We first present studies with enhanced privacy compared to centralized approaches. \citet{10.1145/3381006} proposed the Privacy-Preserving Mobility Prediction (PMF) framework for next-place forecasting. They addressed FL challenges, such as security risks and local performance degradation, by implementing a group optimization strategy, novel aggregation, and client selection methods.  \citet{gjoreski2022toward} employed FL and online learning to estimate cohort percentages (e.g. percentage of gender) for locations, focusing on cohort privacy. Clients select their cohort for participation, enhancing overall system privacy. \citet{10.1145/3397536.3422270} developed the Spatial-Temporal Self-Attention Network (STSAN) and Adaptive Model Fusion FL (AMF) for location prediction, achieving personalized models and outperforming baselines on Foursquare, Tweets, and Yelp datasets. \citet{errounda2022mobility} proposed collaborative training from multiple organizations, suggesting asynchronous FL for future work. \citet{10.1145/3487351.3490964} proposed LocationTrails for learning location embeddings, demonstrating FL's potential but lacking comprehensive implementation. \citet{10257271} also worked on human mobility with a limited number of clients. \citet{9250516} provided an overview of localization services, introducing the FedLoc framework for improved accuracy, calibration, efficiency, and privacy in indoor settings, and identified future challenges for federated localization.

Regarding system heterogeneity, \citet{10.1145/3369830} developed a few-shot learning approach based on a personalized attention mechanism, enabling personalized predictions with minimal data per user. Experiments on distinct devices demonstrated adaptability to system heterogeneity and they suggested pre-training on diverse data to lower costs.
\citet{10274418} explored collaobrative map matching task with heterogeneous devices \cite{10.1145/3550486}, using cellular moving trajectories for continuous and low-cost coverage. \citet{10074979} addressed data heterogeneity and scarcity using clustering and representation learning that effectively handle diverse clients with small datasets.

Regarding communication cost, \citet{9671447} introduced Federated Meta-Location Learning (FMLL) to improve spatial accuracy for real-time usage. FMLL consists of meta-location generation, a predictive model, and an FL framework, achieving low-latency predictions on smartphones. \citet{ezequiel2022federated} investigated the performances of direct integration of FL with centralized models, and also found that simpler architectures might perform better for next-place prediction.

Regarding studies on travel mode identification, \citet{9359187} used an attention mechanism-based CNN, addressing non-IID data by sharing public data to balance datasets. \citet{9632695} introduced a teacher-student scheme to generate pseudo-labels, with consistency updating and mean-teacher-averaging mechanisms, showing effectiveness with limited data. \citet{9514368} also worked on semi-supervised FL, similar to \citet{9632695}, but innovated with a grouping-based aggregation scheme and data-flipping augmentation to enhance performance under non-IID conditions. \citet{9922022} presented a method that uses a simple architecture with extra prediction layers at the server.

\section{Human Activity Recognition}
\label{sec:ar}
Activity Recognition involves classifying human activities in various settings, such as daily life routines \cite{10.1145/3123024.3129271} and sports activities \cite{host2022overview}, often utilizing wearable sensors \cite{qiu2022multi} or video-based methods \cite{10.1145/3447744}.

We highlight advancements in privacy-friendly FL for HAR, also including posture and gesture recognition and driver status monitoring. Tables \ref{tb:ar} and \ref{tb:arc} summarize the studies.

\begin{table*}
  \caption{FL Studies of Activity Recognition in Terms of Our Eight-dimension Assessment}
  \label{tb:ar}
  \begin{tabular}{cc cccc cccc c}
    \toprule    App.&Data&Privacy&CommCost&SysHetero&StatHetero&U.D.U.&Simplified&Ser.O.&Cli.O.&Ref\\
    \midrule
     HAR& Inertia &\checkmark& N/A& N/A & N/A &N/A &N/A  &N/A &N/A & \cite{9767215}  \\
     HAR& Inertia &\checkmark & & &  &    &  &  & & \cite{XIAO2021107338}  \\
     HAR& Inertia &\checkmark & & & & &  &\checkmark & & \cite{10.1145/3594739.3610675}  \\
     HAR& Inertia &\checkmark & & &\checkmark & &  & & \checkmark& \cite{liu2022federated}  \\
     HAR& Inertia &\checkmark & & & \checkmark & \checkmark &  & \checkmark & \checkmark& \cite{9656620}  \\
     HAR& Multi   &\checkmark &\checkmark & &\checkmark & &  &\checkmark &\checkmark & \cite{10.1145/3580795}  \\

     HAR& Multi   & & \checkmark & & & &  &  &\checkmark & \cite{10223407}  \\
     HAR& Multi   & & \checkmark& &  &    &  & \checkmark &\checkmark & \cite{10.1145/3485730.3485946}  \\
     HAR& Inertia & & \checkmark & &  & \checkmark   &  & \checkmark & & \cite{zhao2020semi}  \\
     HAR& Inertia & & \checkmark & & \checkmark &    & \checkmark & \checkmark & & \cite{8672262}  \\
     HAR& Multi   & & \checkmark& & \checkmark &    &  & \checkmark & & \cite{10.1145/3458864.3467681}  \\
     HAR& Multi   & & \checkmark& & \checkmark &    &  & \checkmark & & \cite{10.1145/3554980}  \\
     HAR& Inertia & & \checkmark& &\checkmark & &  & &\checkmark & \cite{10122911}  \\

     HAR& Inertia & & \checkmark&\checkmark &\checkmark & &  & N/A& N/A& \cite{10162197}  \\
     HAR& Inertia & & \checkmark&\checkmark &  && \checkmark & \checkmark & & \cite{imteaj2021exploiting}  \\
     HAR& Inertia & & &\checkmark & \checkmark & & \checkmark && \checkmark& \cite{gudur2021resource}  \\
     HAR& Inertia & & &\checkmark & \checkmark &    &  & & & \cite{9693094}  \\
     HAR& Multi   & & & \checkmark & \checkmark & &  &\checkmark & & \cite{yang2022cross}  \\

     HAR& Multi & & & &\checkmark & &  &  & & \cite{iacob2023privacy}  \\
     HAR& Video & & & &\checkmark & &  &  & & \cite{psaltis2022deep}  \\
     HAR& Inertia & & & & \checkmark &    &  &  & \checkmark& \cite{10.1145/3442381.3450006}  \\
     HAR& Inertia & & & & \checkmark &    &  & &\checkmark & \cite{ek2022evaluation}  \\
     HAR& Inertia & & & & \checkmark &    &  & & \checkmark& \cite{SHAIK2022109929}  \\
     HAR& Inertia & & & & \checkmark & &  & &\checkmark & \cite{9762352}  \\
     HAR& Inertia & & & &\checkmark & &  & & \checkmark& \cite{shen2022federated}  \\
     HAR& Inertia & & & &\checkmark & &  & &\checkmark & \cite{al2023group}  \\
     HAR& Multi & & & &\checkmark & &  & &\checkmark & \cite{shen2023federated}  \\
     
     HAR& Inertia & & & &\checkmark &\checkmark &  & &\checkmark & \cite{10.1145/3594739.3610739}  \\
     HAR& Image & & & &\checkmark &\checkmark &  &  & \checkmark& \cite{10266762}  \\
     HAR& Inertia & & & & &\checkmark &  & &\checkmark & \cite{presotto2022federated}  \\
     HAR& Inertia & & & &  & \checkmark   &  &  & & \cite{9680185}  \\
     HAR& Inertia &N/A & N/A& N/A& N/A & \checkmark   & N/A &N/A &N/A & \cite{9767369}  \\
     HAR& Inertia & & & &  &  \checkmark  & \checkmark &  & \checkmark& \cite{bettini2021personalized}  \\
     HAR& Inertia & & & &  &  \checkmark  & \checkmark &  & \checkmark& \cite{presotto2022semi}  \\
     
     HAR& Multi & & & & & &  & &\checkmark & \cite{Wang_2022_CVPR}  \\
     HAR& Inertia & & & &  & &  &\checkmark & & \cite{9853484}  \\
     HAR& Inertia & & & &  & &  & & & \cite{9821084}  \\
     HAR& Inertia & & & & & &  & & & \cite{gonul2022human}  \\

  \bottomrule
\end{tabular}
\justifying
Abbreviations and concept extension in the table: \textbf{App.}: Applications. \textbf{Data}: Raw Data Type. \textbf{Privacy}: Privacy (and Security). \textbf{CommCost}: Communication Cost. \textbf{SysHetero}: System Heterogeneity. \textbf{StatHetero}: Statistical Heterogeneity. \textbf{U.D.U.}: Unlabeled Data Usage. \textbf{Simplified}: Simplified Setup. \textbf{Ser.O.}: Server-optimized Federated Learning. \textbf{Cli.O.}: Client-optimized Federated Learning. \textbf{Ref}: Reference. 
\textbf{HAR}: Human Activity Recognition.
\textbf{\checkmark} notes considerations (with an exception that a \textbf{\checkmark} in Simplified is undesired).
\end{table*}

\begin{table*} 
  \caption{FL Studies of Activity Recognition in Terms of Our Eight-dimension Assessment (continue)}
  \label{tb:arc}
  \begin{tabular}{cc cccc cccc c}
    \toprule    App.&Data&Privacy&CommCost&SysHetero&StatHetero&U.D.U.&Simplified&Ser.O.&Cli.O.&Ref\\
    \midrule
     HAR& Inertia & & & & & &  & & & \cite{sachin2022federated}  \\
     HAR& Inertia & & & & & &  & & & \cite{sanchez2023federated}  \\
     HAR& Inertia & & & & & &  & & & \cite{10.1145/3594739.3611323}  \\
     HAR& Inertia & & & & & &  &  & & \cite{10223939}  \\
     HAR& Video & & & & & &  &  & & \cite{9892150}  \\
     HAR& Image & & & & & &  &  & & \cite{10196311}  \\
     HAR& C/R & & & & & &  &  &\checkmark & \cite{khan4395564federated}  \\
     HAR& Inertia & & & &  &    &  &  & & \cite{9488681}  \\
     HAR& Inertia &N/A &N/A &N/A & N/A & N/A   & N/A & N/A & N/A& \cite{10.1145/3410530.3414321}  \\
     
     P.\&G.& C/R & &\checkmark & & & &  &  & & \cite{10.1145/3556562.3558575}  \\
     P.\&G.& C/R & &\checkmark & & & &  &  & & \cite{10005134}  \\
     P.\&G.& C/R & &\checkmark & &\checkmark & &  &  &\checkmark & \cite{10283763}  \\
     P.\&G.& C/R & & & &\checkmark & &  &  & & \cite{zhang2022wifi}  \\
     P.\&G.& C/R & & & & & &  &  & & \cite{9546898}  \\
     
     P.\&G.& Video & & & &\checkmark & &  &\checkmark  & & \cite{10.1145/3581783.3612134}  \\
     P.\&G.& Image & & & & & &  &  & \checkmark& \cite{10101124}  \\
     P.\&G.& Multi & & & & & &  &  & & \cite{9956474}  \\
     
     DAR& Video &\checkmark & & & & &  &  & & \cite{10224953}  \\
     DAR& Multi &\checkmark &N/A &N/A &N/A & N/A&N/A  & N/A &N/A & \cite{novikova2022analysis}  \\

     DAR& Image & &\checkmark& \checkmark&\checkmark & &  & \checkmark &\checkmark & \cite{guo2023icmfed}  \\
     DAR& Image & &\checkmark &\checkmark &\checkmark & &  & \checkmark &\checkmark & \cite{10131959}  \\
     DAR& Multi & &\checkmark &\checkmark &\checkmark & &  &  &\checkmark & \cite{Yuan_2023_CVPR}  \\
     DAR& Image & &\checkmark & &\checkmark & &&&\checkmark & \cite{10041112}  \\
     DAR& Video & &\checkmark & & & &  & \checkmark & & \cite{zhao2023fedsup}  \\
     DAR& Multi & & & &\checkmark & &  &  & \checkmark & \cite{10.1145/3631421} \\

     DAR& Image & & & & & &  &  &\checkmark & \cite{wang2020fed}  \\
     DAR& Image & & & & & &\checkmark  &  &\checkmark & \cite{10186757}  \\
     DAR& Multi &N/A & N/A& N/A& N/A & N/A&  \checkmark& N/A &N/A & \cite{Doshi_2022_CVPR}  \\
     DAR& Image &N/A & N/A& N/A& N/A & N/A&  \checkmark& N/A &N/A & \cite{10.1145/3507971.3507985}  \\
     DAR& Multi & & & & & &  &  & & \cite{10146344}  \\
     DAR& Video & & & & & &  &  & & \cite{9564936}  \\

  \bottomrule
\end{tabular}
\justifying
Abbreviations and concept extension in the table: \textbf{App.}: Applications. \textbf{Data}: Raw Data Type. \textbf{Privacy}: Privacy (and Security). \textbf{CommCost}: Communication Cost. \textbf{SysHetero}: System Heterogeneity. \textbf{StatHetero}: Statistical Heterogeneity. \textbf{U.D.U.}: Unlabeled Data Usage. \textbf{Simplified}: Simplified Setup. \textbf{Ser.O.}: Server-optimized Federated Learning. \textbf{Cli.O.}: Client-optimized Federated Learning. \textbf{Ref}: Reference. 
\textbf{HAR}: Human Activity Recognition.
\textbf{C/R}: Channel State Information/Received Signal Strength (Indicator).
\textbf{P.\&G.}: Posture and Gesture.
\textbf{DAR}: Driver Activity Recognition.
\textbf{\checkmark} notes considerations (with an exception that a \textbf{\checkmark} in Simplified is undesired).
\end{table*}

\subsection{Human Activity Recognition}

We first present FL-HAR studies with privacy and security considerations.
\citet{9767215} were the first to investigate security attacks in FL-HAR, focusing on Membership Inference Attacks (MIA), revealing significant vulnerabilities in personal information exposure.
\citet{10.1145/3594739.3610675} addressed diverse privacy concerns by allowing clients to apply local DP techniques, ensuring fairness with a client update selection mechanism to filter out malicious participants.
\citet{10.1145/3580795} introduced a personalized FL framework using hierarchical clustering to enhance accuracy, fairness, and robustness, identifying malicious nodes through abnormal local updates and incorporating dynamic clustering for scalability. \citet{XIAO2021107338} introduced a method for improved representation learning and homomorphic encryption for security.
\citet{liu2022federated} introduced a federated personalized Random Forest model, using DP in training and a two-step personalization process for improved privacy and accuracy.

Next, we introduce studies considering communication costs.
\citet{8672262} outlined a standard FL-HAR workflow as an early work, including modules like data acquisition from sensors and optional data sampling. They suggest less complex models for reducing communication costs while achieving acceptable accuracy.
Subsequent studies \citet{10.1145/3410530.3414321,9488681} further explored FL's benefits over centralized or local methods.
\citet{10.1145/3485730.3485946} introduced FedDL, an FL-HAR framework with dynamic layer-sharing, leveraging client similarities to enhance communication efficiency and customization. FedDL showed efficacy across diverse datasets, including IMUs, UWB, and depth camera images. It was built on their earlier ClusterFL work \cite{10.1145/3458864.3467681}, and the team subsequently released an updated ClusterFL \cite{10.1145/3554980}.
\citet{zhao2020semi} proposed a method in which clients learn an encoder-decoder using unlabeled data, aggregated on the server for prediction, achieving higher accuracy and reduced model size.
\citet{10122911} addressed non-IID data issues by learning and sharing activity prototypes, which also reduced communication costs by exchanging prototypes instead of model parameters.
\citet{Wang_2022_CVPR,10223407} addressed pedestrian movement estimation using images and videos. \citet{Wang_2022_CVPR} developed a client-optimized FL framework with personalization, and \citet{10223407} used reinforcement learning and a transformer to deliver enhanced performance and computational efficiency.

Studies of FL-HAR system heterogeneity are focused on multi-device and limited resource situations.
\citet{yang2022cross} analyzed the multimodality issue and proposed a method to create a modality-agnostic feature space that achieves HAR performance.
\citet{9693094} proposed a two-dimensional FL framework, vertically integrating data between devices and vendors and horizontally aggregating models. 
\citet{gudur2021resource} employed knowledge distillation to address system heterogeneity. In contrast, \citet{imteaj2021exploiting} excluded struggling clients for a robust server model. 
\citet{10162197} provided a system-level analysis of FL-HAR against centralized models, focusing on sensor heterogeneity and also data integrity.

Statistical heterogeneity is also a significant challenge in FL-HAR. \citet{ek2022evaluation} addressed this challenge using a dynamic growth algorithm to detect diverging neurons, achieving high F-scores but with an added computational burden.  \citet{10.1145/3442381.3450006} showed that a multi-task approach with a common embedding encoder enhances model generalization and reduces overfitting. \citet{9762352} proposed a federated clustering approach that computes similarity with subject-specific parameters and incorporates transfer learning for personalization.
\citet{SHAIK2022109929} introduced FedStack, which supports architectural heterogeneity by stacking predictions from the client model to train a central model. \citet{shen2023federated} introduced DivAR, employing federated meta-learning for various sensor types and incorporating social factors for clustering. \citet{shen2022federated} proposed a federated multi-task attention model for HAR with personalized attention modules.
\citet{al2023group} proposed a GP-FL method to personalize clients in groups, optimized for wearable devices with limited computational resources. \citet{psaltis2022deep} suggested incorporating depth and 3D flow information in HAR with federated aggregation. They worked with highly heterogeneous datasets and introduced a large-scale 3D action recognition dataset suitable for FL.

Channel State Information (CSI) has also been used as a non-invasive method for FL-HAR frameworks \cite{iacob2023privacy,khan4395564federated}.
\citet{iacob2023privacy} applied CSI from a Network Card Interface and Passive Wi-Fi Radar, using mutual learning of group-level models to reduce accuracy degradation to just 7\% compared to the centralized approach. \citet{khan4395564federated} used CSI with FedDist \cite{9439129} for efficient model aggregation, achieving better performance than FedAvg and comparable results to centralized learning. Local learning further improves accuracy for each client.

Unlabeled data usage is crucial in FL-HAR due to the abundance of unlabeled sensor data. \citet{9680185} introduced a method based on semi-supervised learning that also allows clients to choose architectures based on their resource constraints.
\citet{10.1145/3594739.3610739} presented an approach with unsupervised learning, using a representation encoder and pseudo-labeled data for model personalization based on confidence and uncertainty. \citet{bettini2021personalized} proposed a more client-optimized framework with personalization, and \citet{9656620} developed a framework 
that accommodates the real-world scenario of limited labeled data, proposing an unsupervised gradient aggregation method to address online learning challenges.
In addition, \citet{presotto2022federated} introduced SS-FedCLAR based on prior findings in \cite{9762352,presotto2022semi}. SS-FedCLAR is an innovative HAR framework that leverages clustering and semi-supervised learning. \citet{9767369} showed that self-supervised FL outperforms centralized learning with autoencoders. \citet{10266762} designed FedAWR for air writing recognition, using interactive active learning with unlabeled data to collect user annotations. 

\citet{10.1145/3594739.3611323} explored smart glasses IMU data for HAR, and FL models achieving slightly lower F1 scores than centralized models, but showed different sensor placements might influence the HAR results \cite{10162197}.
\citet{10223939} used Modified Topology Preserving Domain Adaptation (modTPDA) to enhance performance with scarce labeled data by aligning the topological structures in datasets.
\citet{9853484} introduced P-FedAvg within the Hyperparameter FL (HFL) framework, modulating model weights across participants.
\citet{sachin2022federated} demonstrated FL models on the Mhealth dataset achieve similar results to centralized models, suggesting on-device training in real mobile devices.
\citet{sanchez2023federated} also studied the Mhealth dataset, showing significant score improvements in the first five communication rounds.
\citet{9821084} validated FL-HAR by separating clients into Sensing Devices and Federated Aggregators to fit a distributed scenario.
\citet{gonul2022human} tested FL-HAR with the UT Smoking dataset, finding FL equivalent to centralized approaches.
\citet{9892150} used FL to detect physical violence in videos, and 
\citet{10196311} classified E-learning on-screen activities through screenshots, achieving high accuracy with different models.

Unlike HAR, posture and gesture recognition specifically support smart home systems and interactive environments. 
CSI-based gesture recognition typically uses body-coordinate velocity profiles (BVPs). In this domain, \citet{9546898,zhang2022wifi} introduced an averaging algorithm instead of FedAvg to improve robustness.
\citet{10.1145/3556562.3558575} used transfer learning to address cross-domain challenges, detailed in their subsequent research \cite{10005134}.
\citet{10283763} presented a personalized and communication-efficient framework for WiFi-based gesture recognition systems to address challenges arising from non-IID data.

In addition, visual elements, such as videos and images, also enhance system reliability.
\citet{9956474} introduced FedGait, an FL benchmark for gait recognition, and demonstrated its effectiveness across diverse institutions and devices in non-IID settings.
\citet{10101124} developed methods for correcting sitting and yoga postures, as well as detecting hand gestures.
\citet{10.1145/3581783.3612134} presented FedCSR, the first framework for federated cued speech recognition, employing a mutual knowledge distillation approach to learn both linguistic and visual information.

\subsection{Driver Activity Recognition}

Driver Activity Recognition (DAR) uses in-vehicle sensor data to enhance navigation and safety, monitoring conditions like drowsiness, fatigue, and distractions \cite{Doshi_2022_CVPR}.
DAR employs videos, images, physiological signals, and transportation sensors, unlike HAR, which typically uses wearable or mobile devices.
This subsection discusses recent FL-DAR research, focusing on Human Sensing with human drivers, excluding studies of autonomous driving and smart vehicles.

\citet{9564936} detected driver drowsiness --characterized by eye closure, nodding, and yawning-- through facial expressions captured by cameras, showing competitive performance of FL compared to centralized models, even with highly non-IID client data. Furthermore, several studies have jointly addressed communication costs and data heterogeneity. \citet{zhao2023fedsup} a communication-efficient FL method for detecting driver fatigue. The method uses a client-edge-cloud architecture with a Bayesian CNN for high-uncertainty image selection and an Uncertainty Weighted Asynchronous Aggregation (UWAA) algorithm. \citet{10224953} enhanced this work by integrating DP to balance security and accuracy. \citet{10131959} used transfer learning and orderly dropout to minimize communication cost and overfitting. \citet{Yuan_2023_CVPR} developed a peer-to-peer FL framework emphasizing continual learning without a server, reducing hardware and communication overhead. \citet{10041112} proposed bidirectional knowledge distillation for distraction detection, improving communication efficiency and convergence speed. \citet{guo2023icmfed} introduced an incremental and cost-efficient federated meta-learning mechanism for detecting distractions. It optimized client-server interactions to reduce communication costs.

\citet{10.1145/3631421} applied FL to driver maneuver identification, proposing an anomaly-aware framework (AF-DMIL) to classify driving behaviors like left and right turns based on smartphone sensor data, addressing challenges with decentralized settings and anomaly detection.
\citet{wang2020fed} developed Fed-SCNNl to recognize distractions from driving such as phone use and handoffs from the wheel. Fed-SCNN uses a weighted mix model with an aggregated network and a locally preserved shallow CNN.
\citet{Doshi_2022_CVPR} assessed federated DAR performance on AI City Challenge and StateFarm datasets, showing that FL results can be close to centralized ones. \citet{10.1145/3507971.3507985} tested four different CNN models for FL, achieving competitive results compared to centralized training. \citet{10186757} developed a lane-change prediction model based on DAR, using drivers' head positions and rotations from video data, applying clustering for model personalization but facing latency challenges.
We recommend further reading of \cite{novikova2022analysis} on an analysis review of privacy enhancements in FL-DAR and \cite{10146344} on backbone model choice.

\section{Interface Development}
\label{sec:hci}


Designing user-friendly interfaces requires analyzing user preferences through various sensing modalities, enhancing user experience. However, continuous monitoring of user experiences for interface development often faces privacy concerns. In this section, we briefly discuss the potential of FL to construct a collaborative and effective interface, including keyboards and virtual reality devices. Table. \ref{tb:ide} summarizes the studies presented in this section.

\begin{table*}
  \caption{FL Studies of Interface Development in Terms of Our Eight-dimension Assessment}
  \label{tb:ide}
  \begin{tabular}{cc cccc cccc c}
    \toprule    App.&Data&Privacy&CommCost&SysHetero&StatHetero&U.D.U.&Simplified&Ser.O.&Cli.O.&Ref\\
    \midrule
     KB.&Text  & & & & &N/A &  & \checkmark & & \cite{hard2018federated}  \\
     KB.&Text  & & & & & &  & \checkmark & & \cite{yang2018applied}  \\
     KB.&Text  & & & & & &  & \checkmark & & \cite{ramaswamy2019federated}  \\
     KB.&Text  & & & & & &  & \checkmark & & \cite{chen2019federated}  \\
     VR& Video  & & & & & &  & \checkmark & & \cite{9013419}  \\
     VR& Video & & & & & &  & \checkmark & & \cite{8851408}  \\
     VR& Image & & & & & &  &  & & \cite{10012859}  \\
  \bottomrule
\end{tabular}
\justifying
Abbreviations and concept extension in the table: \textbf{App.}: Applications. \textbf{Data}: Raw Data Type. \textbf{Privacy}: Privacy (and Security). \textbf{CommCost}: Communication Cost. \textbf{SysHetero}: System Heterogeneity. \textbf{StatHetero}: Statistical Heterogeneity. \textbf{U.D.U.}: Unlabeled Data Usage. \textbf{Simplified}: Simplified Setup. \textbf{Ser.O.}: Server-optimized Federated Learning. \textbf{Cli.O.}: Client-optimized Federated Learning. \textbf{Ref}: Reference. 
\textbf{KB.}: Keyboard.
\textbf{VR}: Virtual Reality.
\textbf{\checkmark} notes considerations (with an exception that a \textbf{\checkmark} in Simplified is undesired).
\end{table*}

\citet{hard2018federated} introduced an FL system for typing prediction using data from consenting Gboard users.
An LSTM model was refined locally on devices, and the trained model surpassed server-trained CIFG and traditional N-gram models in evaluations.
FL was also used for search query suggestions \cite{yang2018applied}, Emoji enhancements \cite{ramaswamy2019federated}, and handling out-of-vocabulary words \cite{chen2019federated} in the later work, significantly enriching user interaction.
\citet{9013419,8851408} used FL to enhance VR experiences by minimizing the break-in presence (BIP) using deep echo state networks, which predicted user movements and orientations better than centralized models. \citet{10012859} explored multi-view synthesizing for FL-VR with neural radiance field (NeRF) models to collect data without privacy issues.
While these studies showed excellent performance, most did not explore advanced FL frameworks. 

\section{Challenges and Future Directions}
\label{sec:future}
\begin{figure}[bt]
  \centering
  \includegraphics[width=\linewidth]{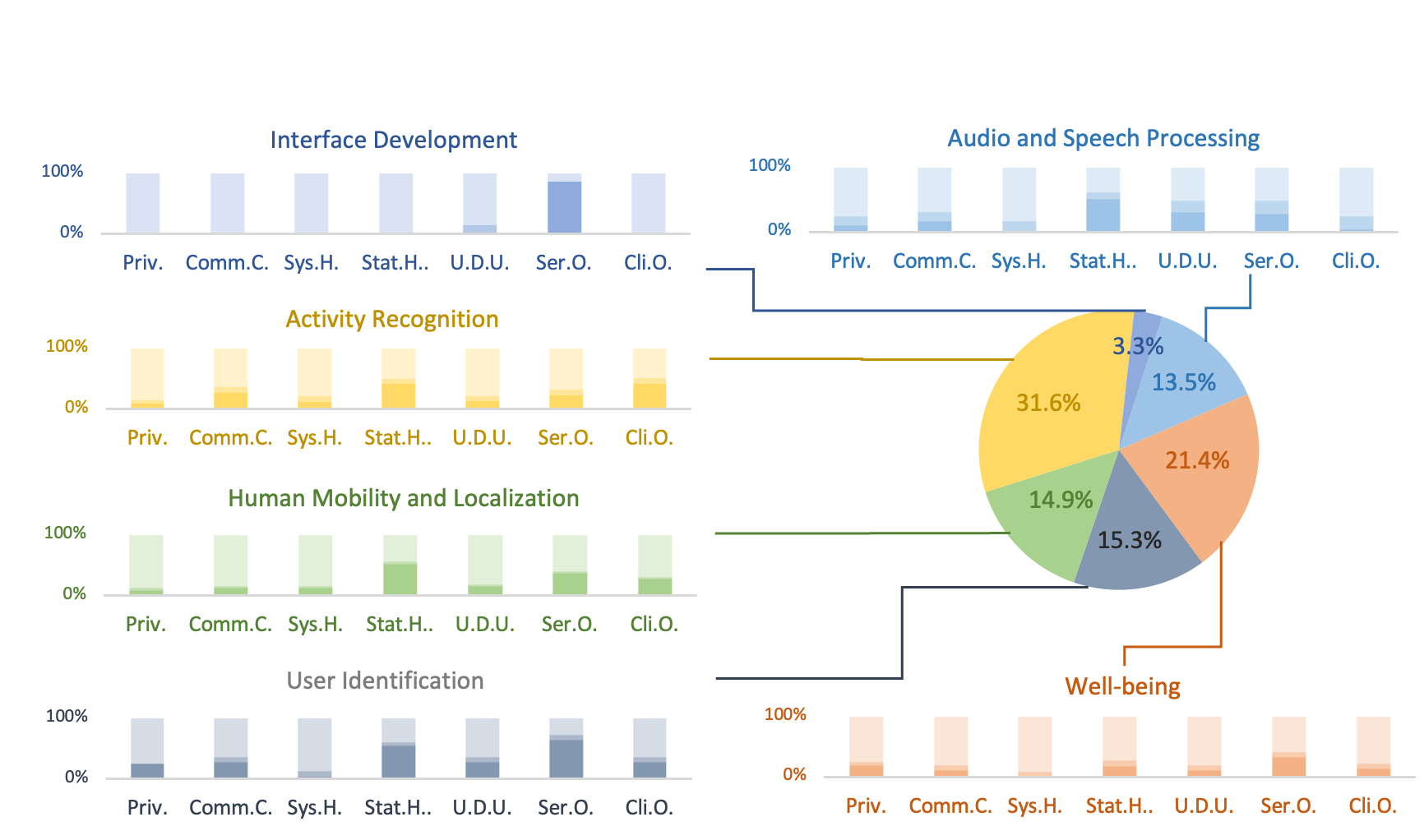}
  \caption{A summary pie chart. For each section, a histogram illustrates the percentage of studies that covered a specific FL characteristic. Dark color represents Consideration, moderate color represents Not Applicable, and light color represents No Consideration. \textbf{Priv.}: Privacy (and Security). \textbf{Comm.C.}: Communication Cost. \textbf{Sys.H.}: System Heterogeneity. \textbf{Stat.H.}: Statistical Heterogeneity. \textbf{U.D.U.}: Unlabeled Data Usage. \textbf{Ser.O.}: Server-optimized Federated Learning. \textbf{Cli.O.}: Client-optimized Federated Learning.}
  \label{fig:fig_summary}
\end{figure}

Fig. \ref{fig:fig_summary} summarizes the corpus reviewed in this survey. The pie chart shows that \emph{Activity Recognition} (31.6\%) and \emph{Well-being }(21.4\%) are the domains with the highest number of studies, followed by \emph{User Identification} (15.3\%), \emph{Human Mobility and Localization} (14.9\%), and \emph{Audio and Speech Processing} (13.5\%). \emph{Interface development} (3.3\%) is the domain with the smallest number of studies.

The histograms in each domain show different levels of consideration for different FL characteristics. A darker color represents \emph{Consideration}, a moderate color represents \emph{Not Applicable}, and a lighter color represents \emph{No Consideration}. This representation provides insights into current priorities and research directions in the application of FL in different domains. Note that the assessment related to \emph{Simplified} FL setup was left out of the figure because it is an undesired feature that shows that the basic FL setup needs improvements.

An analysis of the histograms for each domain shows:
\begin{itemize}
    \item \textbf{Audio and Speech Processing:} The community showed great effort to solve statistical heterogeneity, leverage unlabeled audio data, and optimize the server-side model. The other FL characteristics require more research attention.
    \item \textbf{Well-being:} This area shows minimal overall consideration of the eight FL characteristics we analyzed, with a somewhat even distribution across studies, but none being particularly emphasized, except the Server-optimized FL which is relatively well covered.
    \item \textbf{User Identification:} More than half of the studies considered statistical heterogeneity and server-side optimization. One future direction is reducing the model size and bandwidth requirements for real-time response and resource-limited biometric identification systems.
    \item \textbf{Human Mobility and Localization:} A limited number of studies explored privacy and communication cost, while both are important in this field since traveling history and location contain abundant individual information, and the clients are typically using mobile device with limited resources.
    \item \textbf{Activity Recognition:} Statistical heterogeneity and client-side optimization received more attention compared to the other FL characteristics. Leveraging unlabeled data to improve performance is a promising future direction in this field.
    \item \textbf{Interface Development:} Server-optimized FL is given significant consideration compared to others, indicating a focus on optimizing server-side processes in interface development. The other FL challenges require more research attention.
\end{itemize}

In general across all domains, \emph{Statistical Heterogeneity} and \emph{Server-optimized FL} are the two FL characteristics that are analyzed more frequently in FL studies compared to the others. This is probably because it is easier to explore and simulate such experimental setups compared to privacy attacks, for example, leaving other areas less explored. Building on the less researched challenges in FL for Human Sensing, in the following subsections, we highlight five research aspects related to FL in Human Sensing that require urgent research attention.


\subsection{Privacy and Security under Attacks}
FL enhances the privacy and security of the ML system by replacing direct data sharing with model sharing, and we observed that most studies in this review consider that a basic implementation of FL is already robust enough for privacy or security. However, FL remains vulnerable to attacks: leakage from gradients is possible \cite{zhu2019deep}.

For client privacy, inference attacks, including the property inference attack \cite{10.1145/3243734.3243834}, the membership inference attack \cite{10.1145/3523273}, and the model inversion attack \cite{10.1145/3359789.3359824}, may pose a great threat: The distributions of datasets, inputs, and labels could be inferred from gradients, as well as information not related to the task.
For server security, data poisoning attack \cite{9618642} or model poisoning attack \cite{bagdasaryan2020backdoor} can greatly undermine the integrity of the aggregated model.

Real-world implementations should not tolerate any data leakage. For example, \citet{feng2021attribute} demonstrated the potential for high-accuracy decoding of sensitive attributes like gender from speech-related tasks.
Our review reveals that only a few studies have prioritized enhanced privacy measures, underscoring the significant challenges in developing fully trustworthy FL systems for real-world applications with approaches such as DP \cite{mcmahan2017learning}.

\subsection{Unlimited Participation for All}
As FL experimental simulations involving heterogeneous systems are barely practical compared to heterogeneous datasets, system heterogeneity is the least addressed attribute in the corpus of reviewed studies. Our analysis showed that truly heterogeneous FL systems include datasets collected across diverse devices or include models deployed on heterogeneous edge devices. Both scenarios require significant effort. However, most studies are based on publicly available data collected under controlled experimental conditions that favor homogeneity over heterogeneity. Consequently, heterogeneous datasets are often excluded. Transitioning from experimental FL simulations to real-world device deployment would be a significant step forward for FL.

For example, \citet{hard2018federated} presented real-world criteria for client eligibility: sufficient memory, stable network connection, charging status, idle state, and updated software installation.
Although these criteria may be specific to a specific implementation rather than inherent in FL, they illustrate the feasibility of imposing higher participation standards. However, such high standards significantly limit the applicability of the FL system at a large scale, which is a typical final goal in Human Sensing (e.g., worldwide deployment of an FL system). Not to mention that these high standards may introduce fairness and bias problems if a system is deployed worldwide. Therefore, addressing the issue of stragglers, whose participation is lagged due to lower hardware or connection quality, is crucial.

Future work should focus on developing self-adaptive and personalized strategies to enable servers to facilitate training and deployment across clients with diverse conditions and hardware configurations, which encourages unlimited participation by accommodating edge devices in dynamic environments.

\subsection{Exploiting Unlabeled Data in The Wild}
The common assumption that clients possess fully labeled data most often does not hold in practice \cite{jin2023federated}. Clients often lack the incentive or the expertise required to label extensive local datasets. The practice of distributing well-labeled data among simulated clients, prevalent in more than half of the reviewed studies, does not represent real-world scenarios.
Typically, servers might possess a limited pool of publicly available labeled data, while the majority of client-held data are unlabeled or only partially labeled. Moreover, domain shift \cite{zhang2021adaptive} is a constant challenge: the data collected by the clients often diverge from the server data, showcasing different underlying distributions. Addressing this challenge is especially important in FL for Human Sensing because of the large individual differences among us (clients in FL terms).

Studies on learning without full labels include semi-supervised \cite{zhu2005semi}, transfer \cite{5288526}, and self-supervised learning \cite{9462394}. \citet{jin2023federated} summarized that current research focuses primarily on computer vision and natural language processing experiments \cite{peng2019federated}, and also lacks natural data partitions \cite{jeong2020federated}.
In Human Sensing within FL, only a few studies have explored the benefits of pseudo labeling \cite{9103044}, teacher-student models \cite{hard2020training}, and representation learning \cite{Shome_2021_ICCV}, indicating room for broader application and research.

However, centralized methods for learning without full labels face novel challenges when integrated into FL frameworks. Firstly, labeling is an inherent privacy-sensitive procedure, and sharing knowledge in the labels among clients risks individuals' private information. Secondly, traditional methods that use both labeled and unlabeled data on the server cannot be directly transferred to FL, where the data are isolated. Lastly, the potential utility of unlabeled data remains ambiguous. Researchers need to determine the optimal balance between the extent of data required and the potential for performance enhancement.

\subsection{Recognizing Your Primary Target: Server and/or Clients}
The transition from learning with only local client data to FL is often motivated by the potential for enhanced model performance. However, when a server prioritizes validation performance on its dataset, it may either disregard less-represented (``outlier'') clients as noise/threats, or simply minimize their contributions through the averaging process resulting in a model that reflects only the majority of clients.
Consequently, the reduced performance for these ``outlier'' clients can lower their participation incentive, leading to a more homogeneous training dataset. This scenario might not immediately worsen model performance but overlooks the broader implications for model generalization.

The majority client selection strategy in FL aims to protect the server from clients presenting low-quality data, favoring average clients over unique ones. However, this approach encounters difficulties in client-centric tasks like Human Sensing. Clients are ``selfish'' since they withdraw if no advantages are perceived, especially under the threat of data leakage. Thus, a significant challenge in implementing FL effectively involves ensuring the satisfaction of all participants. Moreover, the server must foster an accommodating environment for participation, allowing new clients to join seamlessly with a readily available model and ensuring an effortless take-over. It should withstand fluctuation in participation, recognize contributions, and accurately identify malicious attacks versus stragglers. Recognizing the requirements of both the server and clients is essential for developing an FL framework suited to real-world applications.

\subsection{No Longer Impossible with Federated Learning}
The growth of the FL community highlights the emerging field of collaborative and privacy-friendly ML, advocating for broader interdisciplinary studies with reduced concerns about personal data leakage and low participation costs.
Besides Human Sensing, in healthcare, where hospitals and medical institutes are prohibited from sharing confidential patient data, FL enables collaboration through model sharing. This approach has shown success in areas such as tumor segmentation \cite{li2019privacy} and management of Electronic Health Records \cite{brisimi2018federated}.

Future directions for FL in practice could focus on two main areas:
Firstly, conducting experiments that reflect real-world complexities. Our analysis showed that many studies have simplified FL settings, neglecting computational costs and data diversity. For example, a smartphone-based FL system for face recognition, tested under ideal conditions, might encounter issues in real applications such as failure due to insufficient storage, inability to process low-resolution images, or missed detection of unique facial features. This underscores the need to incorporate task-specific robustness criteria to achieve high-level performance. Secondly, given the theoretical utility of FL, these types of decentralized and more privacy-friendly approaches could be expanded to other research areas, given that privacy concerns are ubiquitous in data-driven research \cite{langheinrich2001privacy}.
An example is monitoring the cognitive load of workers \cite{10.1145/3626705.3627783}. However, such application causes privacy issues with sensing modalities based on eye tracking, cameras, and keystroke monitors.
In this case, implementing FL can prevent privacy intrusions, and promote job satisfaction, creativity, and productivity.

\section{Conclusion}
\label{sec:con}
This study conducted a systematic review of FL applications since its introduction in Human Sensing.
We established a taxonomy based on raw data types and task categorization to discern trends across six research domains: \emph{Audio and Speech Processing}, \emph{Well-being}, \emph{User Identification}, \emph{Human Mobility and Localization}, \emph{Activity Recognition and Driver Monitoring}, and \emph{Interface Development}.
We assessed each publication against eight dimensions: \emph{Privacy and Security}, \emph{Communication Cost}, \emph{System Heterogeneity}, \emph{Statistical Heterogeneity}, \emph{Unlabeled Data Usage}, \emph{Simplified Setup}, \emph{Server-optimized Federated Learning}, and \emph{Client-optimized Federated Learning}.

We highlighted both limitations and strengths of the reviewed studies, to reveal the gaps between the theoretical benefits of FL and its real-world applicability. Despite analyzing a vast body of related studies, only a minority achieve the reliability necessary for deployment in practical settings. 

\begin{acks}
This study was funded by the projects TRUST-ME (205121L 214991) and XAI-PAC (PZ00P2 216405).
\end{acks}

\bibliographystyle{ACM-Reference-Format}
\bibliography{sample-base}


\begin{thebibliography}{315}


\ifx \showCODEN    \undefined \def \showCODEN     #1{\unskip}     \fi
\ifx \showDOI      \undefined \def \showDOI       #1{#1}\fi
\ifx \showISBNx    \undefined \def \showISBNx     #1{\unskip}     \fi
\ifx \showISBNxiii \undefined \def \showISBNxiii  #1{\unskip}     \fi
\ifx \showISSN     \undefined \def \showISSN      #1{\unskip}     \fi
\ifx \showLCCN     \undefined \def \showLCCN      #1{\unskip}     \fi
\ifx \shownote     \undefined \def \shownote      #1{#1}          \fi
\ifx \showarticletitle \undefined \def \showarticletitle #1{#1}   \fi
\ifx \showURL      \undefined \def \showURL       {\relax}        \fi
\providecommand\bibfield[2]{#2}
\providecommand\bibinfo[2]{#2}
\providecommand\natexlab[1]{#1}
\providecommand\showeprint[2][]{arXiv:#2}

\bibitem[Abdulrahman et~al\mbox{.}(2021)]%
        {9220780}
\bibfield{author}{\bibinfo{person}{Sawsan Abdulrahman}, \bibinfo{person}{Hanine Tout}, \bibinfo{person}{Hakima Ould-Slimane}, \bibinfo{person}{Azzam Mourad}, \bibinfo{person}{Chamseddine Talhi}, {and} \bibinfo{person}{Mohsen Guizani}.} \bibinfo{year}{2021}\natexlab{}.
\newblock \showarticletitle{A Survey on Federated Learning: The Journey From Centralized to Distributed On-Site Learning and Beyond}.
\newblock \bibinfo{journal}{\emph{IEEE Internet of Things Journal}} \bibinfo{volume}{8}, \bibinfo{number}{7} (\bibinfo{year}{2021}), \bibinfo{pages}{5476--5497}.
\newblock
\urldef\tempurl%
\url{https://doi.org/10.1109/JIOT.2020.3030072}
\showDOI{\tempurl}


\bibitem[Aggarwal et~al\mbox{.}(2021)]%
        {9484386}
\bibfield{author}{\bibinfo{person}{Divyansh Aggarwal}, \bibinfo{person}{Jiayu Zhou}, {and} \bibinfo{person}{Anil~K. Jain}.} \bibinfo{year}{2021}\natexlab{}.
\newblock \showarticletitle{FedFace: Collaborative Learning of Face Recognition Model}. In \bibinfo{booktitle}{\emph{2021 IEEE International Joint Conference on Biometrics (IJCB)}}. \bibinfo{pages}{1--8}.
\newblock
\urldef\tempurl%
\url{https://doi.org/10.1109/IJCB52358.2021.9484386}
\showDOI{\tempurl}


\bibitem[Agrawal et~al\mbox{.}(2023)]%
        {10112028}
\bibfield{author}{\bibinfo{person}{Manan Agrawal}, \bibinfo{person}{Mohd~Ayaan Anwar}, {and} \bibinfo{person}{Rajni Jindal}.} \bibinfo{year}{2023}\natexlab{}.
\newblock \showarticletitle{FedCER - Emotion Recognition Using 2D-CNN in Decentralized Federated Learning Environment}. In \bibinfo{booktitle}{\emph{2023 6th International Conference on Information Systems and Computer Networks (ISCON)}}. \bibinfo{pages}{1--5}.
\newblock
\urldef\tempurl%
\url{https://doi.org/10.1109/ISCON57294.2023.10112028}
\showDOI{\tempurl}


\bibitem[Ahmed et~al\mbox{.}(2023)]%
        {9767700}
\bibfield{author}{\bibinfo{person}{Usman Ahmed}, \bibinfo{person}{Jerry Chun-Wei Lin}, {and} \bibinfo{person}{Gautam Srivastava}.} \bibinfo{year}{2023}\natexlab{}.
\newblock \showarticletitle{Hyper-Graph Attention Based Federated Learning Methods for Use in Mental Health Detection}.
\newblock \bibinfo{journal}{\emph{IEEE Journal of Biomedical and Health Informatics}} \bibinfo{volume}{27}, \bibinfo{number}{2} (\bibinfo{year}{2023}), \bibinfo{pages}{768--777}.
\newblock
\urldef\tempurl%
\url{https://doi.org/10.1109/JBHI.2022.3172269}
\showDOI{\tempurl}


\bibitem[Al-Saedi and Boeva(2023)]%
        {al2023group}
\bibfield{author}{\bibinfo{person}{Ahmed~A Al-Saedi} {and} \bibinfo{person}{Veselka Boeva}.} \bibinfo{year}{2023}\natexlab{}.
\newblock \showarticletitle{Group-personalized federated learning for human activity recognition through cluster eccentricity analysis}. In \bibinfo{booktitle}{\emph{International Conference on Engineering Applications of Neural Networks}}. Springer, \bibinfo{pages}{505--519}.
\newblock


\bibitem[Alamgir et~al\mbox{.}(2022)]%
        {alamgir2022federated}
\bibfield{author}{\bibinfo{person}{Zareen Alamgir}, \bibinfo{person}{Farwa~K Khan}, {and} \bibinfo{person}{Saira Karim}.} \bibinfo{year}{2022}\natexlab{}.
\newblock \showarticletitle{Federated recommenders: methods, challenges and future}.
\newblock \bibinfo{journal}{\emph{Cluster Computing}} \bibinfo{volume}{25}, \bibinfo{number}{6} (\bibinfo{year}{2022}), \bibinfo{pages}{4075--4096}.
\newblock


\bibitem[Almadhor et~al\mbox{.}(2023)]%
        {almadhor2023wrist}
\bibfield{author}{\bibinfo{person}{Ahmad Almadhor}, \bibinfo{person}{Gabriel~Avelino Sampedro}, \bibinfo{person}{Mideth Abisado}, \bibinfo{person}{Sidra Abbas}, \bibinfo{person}{Ye-Jin Kim}, \bibinfo{person}{Muhammad~Attique Khan}, \bibinfo{person}{Jamel Baili}, {and} \bibinfo{person}{Jae-Hyuk Cha}.} \bibinfo{year}{2023}\natexlab{}.
\newblock \showarticletitle{Wrist-Based Electrodermal Activity Monitoring for Stress Detection Using Federated Learning}.
\newblock \bibinfo{journal}{\emph{Sensors}} \bibinfo{volume}{23}, \bibinfo{number}{8} (\bibinfo{year}{2023}), \bibinfo{pages}{3984}.
\newblock


\bibitem[Aneja et~al\mbox{.}(2017)]%
        {aneja2017modeling}
\bibfield{author}{\bibinfo{person}{Deepali Aneja}, \bibinfo{person}{Alex Colburn}, \bibinfo{person}{Gary Faigin}, \bibinfo{person}{Linda Shapiro}, {and} \bibinfo{person}{Barbara Mones}.} \bibinfo{year}{2017}\natexlab{}.
\newblock \showarticletitle{Modeling stylized character expressions via deep learning}. In \bibinfo{booktitle}{\emph{Computer Vision--ACCV 2016: 13th Asian Conference on Computer Vision, Taipei, Taiwan, November 20-24, 2016, Revised Selected Papers, Part II 13}}. Springer, \bibinfo{pages}{136--153}.
\newblock


\bibitem[Anwar et~al\mbox{.}(2023)]%
        {10041308}
\bibfield{author}{\bibinfo{person}{Mohd~Ayaan Anwar}, \bibinfo{person}{Manan Agrawal}, \bibinfo{person}{Neha Gahlan}, \bibinfo{person}{Divyashikha Sethia}, \bibinfo{person}{Gaurav~Kumar Singh}, {and} \bibinfo{person}{Rishabh Chaurasia}.} \bibinfo{year}{2023}\natexlab{}.
\newblock \showarticletitle{FedEmo: A Privacy-Preserving Framework for Emotion Recognition using EEG Physiological Data}. In \bibinfo{booktitle}{\emph{2023 15th International Conference on COMmunication Systems \& NETworkS (COMSNETS)}}. \bibinfo{pages}{119--124}.
\newblock
\urldef\tempurl%
\url{https://doi.org/10.1109/COMSNETS56262.2023.10041308}
\showDOI{\tempurl}


\bibitem[Baevski et~al\mbox{.}(2020)]%
        {NEURIPS2020_92d1e1eb}
\bibfield{author}{\bibinfo{person}{Alexei Baevski}, \bibinfo{person}{Yuhao Zhou}, \bibinfo{person}{Abdelrahman Mohamed}, {and} \bibinfo{person}{Michael Auli}.} \bibinfo{year}{2020}\natexlab{}.
\newblock \showarticletitle{wav2vec 2.0: A Framework for Self-Supervised Learning of Speech Representations}. In \bibinfo{booktitle}{\emph{Advances in Neural Information Processing Systems}}, \bibfield{editor}{\bibinfo{person}{H.~Larochelle}, \bibinfo{person}{M.~Ranzato}, \bibinfo{person}{R.~Hadsell}, \bibinfo{person}{M.F. Balcan}, {and} \bibinfo{person}{H.~Lin}} (Eds.), Vol.~\bibinfo{volume}{33}. \bibinfo{publisher}{Curran Associates, Inc.}, \bibinfo{pages}{12449--12460}.
\newblock
\urldef\tempurl%
\url{https://proceedings.neurips.cc/paper_files/paper/2020/file/92d1e1eb1cd6f9fba3227870bb6d7f07-Paper.pdf}
\showURL{%
\tempurl}


\bibitem[Bagdasaryan et~al\mbox{.}(2020)]%
        {bagdasaryan2020backdoor}
\bibfield{author}{\bibinfo{person}{Eugene Bagdasaryan}, \bibinfo{person}{Andreas Veit}, \bibinfo{person}{Yiqing Hua}, \bibinfo{person}{Deborah Estrin}, {and} \bibinfo{person}{Vitaly Shmatikov}.} \bibinfo{year}{2020}\natexlab{}.
\newblock \showarticletitle{How to backdoor federated learning}. In \bibinfo{booktitle}{\emph{International conference on artificial intelligence and statistics}}. PMLR, \bibinfo{pages}{2938--2948}.
\newblock


\bibitem[Bai et~al\mbox{.}(2021)]%
        {bai2021federated}
\bibfield{author}{\bibinfo{person}{Fan Bai}, \bibinfo{person}{Jiaxiang Wu}, \bibinfo{person}{Pengcheng Shen}, \bibinfo{person}{Shaoxin Li}, {and} \bibinfo{person}{Shuigeng Zhou}.} \bibinfo{year}{2021}\natexlab{}.
\newblock \showarticletitle{Federated face recognition}.
\newblock \bibinfo{journal}{\emph{arXiv preprint arXiv:2105.02501}} (\bibinfo{year}{2021}).
\newblock


\bibitem[Barbosa et~al\mbox{.}(2018)]%
        {barbosa2018human}
\bibfield{author}{\bibinfo{person}{Hugo Barbosa}, \bibinfo{person}{Marc Barthelemy}, \bibinfo{person}{Gourab Ghoshal}, \bibinfo{person}{Charlotte~R James}, \bibinfo{person}{Maxime Lenormand}, \bibinfo{person}{Thomas Louail}, \bibinfo{person}{Ronaldo Menezes}, \bibinfo{person}{Jos{\'e}~J Ramasco}, \bibinfo{person}{Filippo Simini}, {and} \bibinfo{person}{Marcello Tomasini}.} \bibinfo{year}{2018}\natexlab{}.
\newblock \showarticletitle{Human mobility: Models and applications}.
\newblock \bibinfo{journal}{\emph{Physics Reports}}  \bibinfo{volume}{734} (\bibinfo{year}{2018}), \bibinfo{pages}{1--74}.
\newblock


\bibitem[Belal et~al\mbox{.}(2023)]%
        {belal2023survey}
\bibfield{author}{\bibinfo{person}{Yacine Belal}, \bibinfo{person}{Sonia~Ben Mokhtar}, \bibinfo{person}{Hamed Haddadi}, \bibinfo{person}{Jaron Wang}, {and} \bibinfo{person}{Afra Mashhadi}.} \bibinfo{year}{2023}\natexlab{}.
\newblock \showarticletitle{Survey of Federated Learning Models for Spatial-Temporal Mobility Applications}.
\newblock \bibinfo{journal}{\emph{arXiv preprint arXiv:2305.05257}} (\bibinfo{year}{2023}).
\newblock


\bibitem[Bettini et~al\mbox{.}(2021)]%
        {bettini2021personalized}
\bibfield{author}{\bibinfo{person}{Claudio Bettini}, \bibinfo{person}{Gabriele Civitarese}, {and} \bibinfo{person}{Riccardo Presotto}.} \bibinfo{year}{2021}\natexlab{}.
\newblock \showarticletitle{Personalized semi-supervised federated learning for human activity recognition}.
\newblock \bibinfo{journal}{\emph{arXiv preprint arXiv:2104.08094}} (\bibinfo{year}{2021}).
\newblock


\bibitem[Bhagoji et~al\mbox{.}(2019)]%
        {pmlr-v97-bhagoji19a}
\bibfield{author}{\bibinfo{person}{Arjun~Nitin Bhagoji}, \bibinfo{person}{Supriyo Chakraborty}, \bibinfo{person}{Prateek Mittal}, {and} \bibinfo{person}{Seraphin Calo}.} \bibinfo{year}{2019}\natexlab{}.
\newblock \showarticletitle{Analyzing Federated Learning through an Adversarial Lens}. In \bibinfo{booktitle}{\emph{Proceedings of the 36th International Conference on Machine Learning}} \emph{(\bibinfo{series}{Proceedings of Machine Learning Research}, Vol.~\bibinfo{volume}{97})}, \bibfield{editor}{\bibinfo{person}{Kamalika Chaudhuri} {and} \bibinfo{person}{Ruslan Salakhutdinov}} (Eds.). \bibinfo{publisher}{PMLR}, \bibinfo{pages}{634--643}.
\newblock
\urldef\tempurl%
\url{https://proceedings.mlr.press/v97/bhagoji19a.html}
\showURL{%
\tempurl}


\bibitem[Bn and Abdullah(2022)]%
        {9746827}
\bibfield{author}{\bibinfo{person}{Suhas Bn} {and} \bibinfo{person}{Saeed Abdullah}.} \bibinfo{year}{2022}\natexlab{}.
\newblock \showarticletitle{Privacy Sensitive Speech Analysis Using Federated Learning to Assess Depression}. In \bibinfo{booktitle}{\emph{ICASSP 2022 - 2022 IEEE International Conference on Acoustics, Speech and Signal Processing (ICASSP)}}. \bibinfo{pages}{6272--6276}.
\newblock
\urldef\tempurl%
\url{https://doi.org/10.1109/ICASSP43922.2022.9746827}
\showDOI{\tempurl}


\bibitem[Bonawitz et~al\mbox{.}(2019)]%
        {MLSYS2019_7b770da6}
\bibfield{author}{\bibinfo{person}{Keith Bonawitz}, \bibinfo{person}{Hubert Eichner}, \bibinfo{person}{Wolfgang Grieskamp}, \bibinfo{person}{Dzmitry Huba}, \bibinfo{person}{Alex Ingerman}, \bibinfo{person}{Vladimir Ivanov}, \bibinfo{person}{Chlo\'{e} Kiddon}, \bibinfo{person}{Jakub Kone\v{c}n\'{y}}, \bibinfo{person}{Stefano Mazzocchi}, \bibinfo{person}{Brendan McMahan}, \bibinfo{person}{Timon Van~Overveldt}, \bibinfo{person}{David Petrou}, \bibinfo{person}{Daniel Ramage}, {and} \bibinfo{person}{Jason Roselander}.} \bibinfo{year}{2019}\natexlab{}.
\newblock \showarticletitle{Towards Federated Learning at Scale: System Design}. In \bibinfo{booktitle}{\emph{Proceedings of Machine Learning and Systems}}, \bibfield{editor}{\bibinfo{person}{A.~Talwalkar}, \bibinfo{person}{V.~Smith}, {and} \bibinfo{person}{M.~Zaharia}} (Eds.), Vol.~\bibinfo{volume}{1}. \bibinfo{pages}{374--388}.
\newblock
\urldef\tempurl%
\url{https://proceedings.mlsys.org/paper_files/paper/2019/file/7b770da633baf74895be22a8807f1a8f-Paper.pdf}
\showURL{%
\tempurl}


\bibitem[Bonawitz et~al\mbox{.}(2017)]%
        {10.1145/3133956.3133982}
\bibfield{author}{\bibinfo{person}{Keith Bonawitz}, \bibinfo{person}{Vladimir Ivanov}, \bibinfo{person}{Ben Kreuter}, \bibinfo{person}{Antonio Marcedone}, \bibinfo{person}{H.~Brendan McMahan}, \bibinfo{person}{Sarvar Patel}, \bibinfo{person}{Daniel Ramage}, \bibinfo{person}{Aaron Segal}, {and} \bibinfo{person}{Karn Seth}.} \bibinfo{year}{2017}\natexlab{}.
\newblock \showarticletitle{Practical Secure Aggregation for Privacy-Preserving Machine Learning}. In \bibinfo{booktitle}{\emph{Proceedings of the 2017 ACM SIGSAC Conference on Computer and Communications Security}} (Dallas, Texas, USA) \emph{(\bibinfo{series}{CCS '17})}. \bibinfo{publisher}{Association for Computing Machinery}, \bibinfo{address}{New York, NY, USA}, \bibinfo{pages}{1175–1191}.
\newblock
\showISBNx{9781450349468}
\urldef\tempurl%
\url{https://doi.org/10.1145/3133956.3133982}
\showDOI{\tempurl}


\bibitem[Brisimi et~al\mbox{.}(2018)]%
        {brisimi2018federated}
\bibfield{author}{\bibinfo{person}{Theodora~S Brisimi}, \bibinfo{person}{Ruidi Chen}, \bibinfo{person}{Theofanie Mela}, \bibinfo{person}{Alex Olshevsky}, \bibinfo{person}{Ioannis~Ch Paschalidis}, {and} \bibinfo{person}{Wei Shi}.} \bibinfo{year}{2018}\natexlab{}.
\newblock \showarticletitle{Federated learning of predictive models from federated electronic health records}.
\newblock \bibinfo{journal}{\emph{International journal of medical informatics}}  \bibinfo{volume}{112} (\bibinfo{year}{2018}), \bibinfo{pages}{59--67}.
\newblock


\bibitem[Bukit et~al\mbox{.}(2023)]%
        {10223939}
\bibfield{author}{\bibinfo{person}{Tori~Andika Bukit}, \bibinfo{person}{Ericka Pamela~Bermudez Pillado}, \bibinfo{person}{Seok-Lyong Lee}, {and} \bibinfo{person}{Bernardo~Nugroho Yahya}.} \bibinfo{year}{2023}\natexlab{}.
\newblock \showarticletitle{Federated Topology Preserving Domain Adaptation for Human Activity Recognition}. In \bibinfo{booktitle}{\emph{2023 31st Signal Processing and Communications Applications Conference (SIU)}}. \bibinfo{pages}{1--4}.
\newblock
\urldef\tempurl%
\url{https://doi.org/10.1109/SIU59756.2023.10223939}
\showDOI{\tempurl}


\bibitem[Caldas et~al\mbox{.}(2018)]%
        {caldas2018expanding}
\bibfield{author}{\bibinfo{person}{Sebastian Caldas}, \bibinfo{person}{Jakub Kone{\v{c}}ny}, \bibinfo{person}{H~Brendan McMahan}, {and} \bibinfo{person}{Ameet Talwalkar}.} \bibinfo{year}{2018}\natexlab{}.
\newblock \showarticletitle{Expanding the reach of federated learning by reducing client resource requirements}.
\newblock \bibinfo{journal}{\emph{arXiv preprint arXiv:1812.07210}} (\bibinfo{year}{2018}).
\newblock


\bibitem[Can and Ersoy(2021)]%
        {10.1145/3428152}
\bibfield{author}{\bibinfo{person}{Yekta~Said Can} {and} \bibinfo{person}{Cem Ersoy}.} \bibinfo{year}{2021}\natexlab{}.
\newblock \showarticletitle{Privacy-Preserving Federated Deep Learning for Wearable IoT-Based Biomedical Monitoring}.
\newblock \bibinfo{journal}{\emph{ACM Trans. Internet Technol.}} \bibinfo{volume}{21}, \bibinfo{number}{1}, Article \bibinfo{articleno}{21} (\bibinfo{date}{jan} \bibinfo{year}{2021}), \bibinfo{numpages}{17}~pages.
\newblock
\showISSN{1533-5399}
\urldef\tempurl%
\url{https://doi.org/10.1145/3428152}
\showDOI{\tempurl}


\bibitem[Cao et~al\mbox{.}(2017)]%
        {10.1145/3097983.3098086}
\bibfield{author}{\bibinfo{person}{Bokai Cao}, \bibinfo{person}{Lei Zheng}, \bibinfo{person}{Chenwei Zhang}, \bibinfo{person}{Philip~S. Yu}, \bibinfo{person}{Andrea Piscitello}, \bibinfo{person}{John Zulueta}, \bibinfo{person}{Olu Ajilore}, \bibinfo{person}{Kelly Ryan}, {and} \bibinfo{person}{Alex~D. Leow}.} \bibinfo{year}{2017}\natexlab{}.
\newblock \showarticletitle{DeepMood: Modeling Mobile Phone Typing Dynamics for Mood Detection}. In \bibinfo{booktitle}{\emph{Proceedings of the 23rd ACM SIGKDD International Conference on Knowledge Discovery and Data Mining}} (Halifax, NS, Canada) \emph{(\bibinfo{series}{KDD '17})}. \bibinfo{publisher}{Association for Computing Machinery}, \bibinfo{address}{New York, NY, USA}, \bibinfo{pages}{747–755}.
\newblock
\showISBNx{9781450348874}
\urldef\tempurl%
\url{https://doi.org/10.1145/3097983.3098086}
\showDOI{\tempurl}


\bibitem[Chang et~al\mbox{.}(2022)]%
        {chang2022robust}
\bibfield{author}{\bibinfo{person}{Yi Chang}, \bibinfo{person}{Sofiane Laridi}, \bibinfo{person}{Zhao Ren}, \bibinfo{person}{Gregory Palmer}, \bibinfo{person}{Bj{\"o}rn~W Schuller}, {and} \bibinfo{person}{Marco Fisichella}.} \bibinfo{year}{2022}\natexlab{}.
\newblock \showarticletitle{Robust federated learning against adversarial attacks for speech emotion recognition}.
\newblock \bibinfo{journal}{\emph{arXiv preprint arXiv:2203.04696}} (\bibinfo{year}{2022}).
\newblock


\bibitem[Chawla et~al\mbox{.}(2002)]%
        {chawla2002smote}
\bibfield{author}{\bibinfo{person}{Nitesh~V Chawla}, \bibinfo{person}{Kevin~W Bowyer}, \bibinfo{person}{Lawrence~O Hall}, {and} \bibinfo{person}{W~Philip Kegelmeyer}.} \bibinfo{year}{2002}\natexlab{}.
\newblock \showarticletitle{SMOTE: synthetic minority over-sampling technique}.
\newblock \bibinfo{journal}{\emph{Journal of artificial intelligence research}}  \bibinfo{volume}{16} (\bibinfo{year}{2002}), \bibinfo{pages}{321--357}.
\newblock


\bibitem[Chen(2023)]%
        {10.1145/3592686.3592715}
\bibfield{author}{\bibinfo{person}{Caiyun Chen}.} \bibinfo{year}{2023}\natexlab{}.
\newblock \showarticletitle{Research on Online Teaching Emotion Detection Based on Federated Learning}. In \bibinfo{booktitle}{\emph{Proceedings of the 2023 3rd International Conference on Bioinformatics and Intelligent Computing}} (Sanya, China) \emph{(\bibinfo{series}{BIC '23})}. \bibinfo{publisher}{Association for Computing Machinery}, \bibinfo{address}{New York, NY, USA}, \bibinfo{pages}{159–164}.
\newblock
\showISBNx{9798400700200}
\urldef\tempurl%
\url{https://doi.org/10.1145/3592686.3592715}
\showDOI{\tempurl}


\bibitem[Chen et~al\mbox{.}(2021)]%
        {10.1145/3447744}
\bibfield{author}{\bibinfo{person}{Kaixuan Chen}, \bibinfo{person}{Dalin Zhang}, \bibinfo{person}{Lina Yao}, \bibinfo{person}{Bin Guo}, \bibinfo{person}{Zhiwen Yu}, {and} \bibinfo{person}{Yunhao Liu}.} \bibinfo{year}{2021}\natexlab{}.
\newblock \showarticletitle{Deep Learning for Sensor-Based Human Activity Recognition: Overview, Challenges, and Opportunities}.
\newblock \bibinfo{journal}{\emph{ACM Comput. Surv.}} \bibinfo{volume}{54}, \bibinfo{number}{4}, Article \bibinfo{articleno}{77} (\bibinfo{date}{may} \bibinfo{year}{2021}), \bibinfo{numpages}{40}~pages.
\newblock
\showISSN{0360-0300}
\urldef\tempurl%
\url{https://doi.org/10.1145/3447744}
\showDOI{\tempurl}


\bibitem[Chen et~al\mbox{.}(2019a)]%
        {chen2019federated}
\bibfield{author}{\bibinfo{person}{Mingqing Chen}, \bibinfo{person}{Rajiv Mathews}, \bibinfo{person}{Tom Ouyang}, {and} \bibinfo{person}{Fran{\c{c}}oise Beaufays}.} \bibinfo{year}{2019}\natexlab{a}.
\newblock \showarticletitle{Federated learning of out-of-vocabulary words}.
\newblock \bibinfo{journal}{\emph{arXiv preprint arXiv:1903.10635}} (\bibinfo{year}{2019}).
\newblock


\bibitem[Chen et~al\mbox{.}(2019b)]%
        {9013419}
\bibfield{author}{\bibinfo{person}{Mingzhe Chen}, \bibinfo{person}{Omid Semiari}, \bibinfo{person}{Walid Saad}, \bibinfo{person}{Xuanlin Liu}, {and} \bibinfo{person}{Changchuan Yin}.} \bibinfo{year}{2019}\natexlab{b}.
\newblock \showarticletitle{Federated Deep Learning for Immersive Virtual Reality over Wireless Networks}. In \bibinfo{booktitle}{\emph{2019 IEEE Global Communications Conference (GLOBECOM)}}. \bibinfo{pages}{1--6}.
\newblock
\urldef\tempurl%
\url{https://doi.org/10.1109/GLOBECOM38437.2019.9013419}
\showDOI{\tempurl}


\bibitem[Chen et~al\mbox{.}(2020)]%
        {8851408}
\bibfield{author}{\bibinfo{person}{Mingzhe Chen}, \bibinfo{person}{Omid Semiari}, \bibinfo{person}{Walid Saad}, \bibinfo{person}{Xuanlin Liu}, {and} \bibinfo{person}{Changchuan Yin}.} \bibinfo{year}{2020}\natexlab{}.
\newblock \showarticletitle{Federated Echo State Learning for Minimizing Breaks in Presence in Wireless Virtual Reality Networks}.
\newblock \bibinfo{journal}{\emph{IEEE Transactions on Wireless Communications}} \bibinfo{volume}{19}, \bibinfo{number}{1} (\bibinfo{year}{2020}), \bibinfo{pages}{177--191}.
\newblock
\urldef\tempurl%
\url{https://doi.org/10.1109/TWC.2019.2942929}
\showDOI{\tempurl}


\bibitem[Chen et~al\mbox{.}(2018)]%
        {8622598}
\bibfield{author}{\bibinfo{person}{Xuhui Chen}, \bibinfo{person}{Jinlong Ji}, \bibinfo{person}{Changqing Luo}, \bibinfo{person}{Weixian Liao}, {and} \bibinfo{person}{Pan Li}.} \bibinfo{year}{2018}\natexlab{}.
\newblock \showarticletitle{When Machine Learning Meets Blockchain: A Decentralized, Privacy-preserving and Secure Design}. In \bibinfo{booktitle}{\emph{2018 IEEE International Conference on Big Data (Big Data)}}. \bibinfo{pages}{1178--1187}.
\newblock
\urldef\tempurl%
\url{https://doi.org/10.1109/BigData.2018.8622598}
\showDOI{\tempurl}


\bibitem[Chen and Xu(2023)]%
        {chen2023learning}
\bibfield{author}{\bibinfo{person}{Zhiyong Chen} {and} \bibinfo{person}{Shugong Xu}.} \bibinfo{year}{2023}\natexlab{}.
\newblock \showarticletitle{Learning domain-heterogeneous speaker recognition systems with personalized continual federated learning}.
\newblock \bibinfo{journal}{\emph{EURASIP Journal on Audio, Speech, and Music Processing}} \bibinfo{volume}{2023}, \bibinfo{number}{1} (\bibinfo{year}{2023}), \bibinfo{pages}{33}.
\newblock


\bibitem[Cheng et~al\mbox{.}(2023)]%
        {10122911}
\bibfield{author}{\bibinfo{person}{Dongzhou Cheng}, \bibinfo{person}{Lei Zhang}, \bibinfo{person}{Can Bu}, \bibinfo{person}{Xing Wang}, \bibinfo{person}{Hao Wu}, {and} \bibinfo{person}{Aiguo Song}.} \bibinfo{year}{2023}\natexlab{}.
\newblock \showarticletitle{ProtoHAR: Prototype Guided Personalized Federated Learning for Human Activity Recognition}.
\newblock \bibinfo{journal}{\emph{IEEE Journal of Biomedical and Health Informatics}} \bibinfo{volume}{27}, \bibinfo{number}{8} (\bibinfo{year}{2023}), \bibinfo{pages}{3900--3911}.
\newblock
\urldef\tempurl%
\url{https://doi.org/10.1109/JBHI.2023.3275438}
\showDOI{\tempurl}


\bibitem[Cheng et~al\mbox{.}(2022)]%
        {9761235}
\bibfield{author}{\bibinfo{person}{Xin Cheng}, \bibinfo{person}{Chuan Ma}, \bibinfo{person}{Jun Li}, \bibinfo{person}{Haiwei Song}, \bibinfo{person}{Feng Shu}, {and} \bibinfo{person}{Jiangzhou Wang}.} \bibinfo{year}{2022}\natexlab{}.
\newblock \showarticletitle{Federated Learning-Based Localization With Heterogeneous Fingerprint Database}.
\newblock \bibinfo{journal}{\emph{IEEE Wireless Communications Letters}} \bibinfo{volume}{11}, \bibinfo{number}{7} (\bibinfo{year}{2022}), \bibinfo{pages}{1364--1368}.
\newblock
\urldef\tempurl%
\url{https://doi.org/10.1109/LWC.2022.3169215}
\showDOI{\tempurl}


\bibitem[Chhikara et~al\mbox{.}(2021)]%
        {9253631}
\bibfield{author}{\bibinfo{person}{Prateek Chhikara}, \bibinfo{person}{Prabhjot Singh}, \bibinfo{person}{Rajkumar Tekchandani}, \bibinfo{person}{Neeraj Kumar}, {and} \bibinfo{person}{Mohsen Guizani}.} \bibinfo{year}{2021}\natexlab{}.
\newblock \showarticletitle{Federated Learning Meets Human Emotions: A Decentralized Framework for Human–Computer Interaction for IoT Applications}.
\newblock \bibinfo{journal}{\emph{IEEE Internet of Things Journal}} \bibinfo{volume}{8}, \bibinfo{number}{8} (\bibinfo{year}{2021}), \bibinfo{pages}{6949--6962}.
\newblock
\urldef\tempurl%
\url{https://doi.org/10.1109/JIOT.2020.3037207}
\showDOI{\tempurl}


\bibitem[Ciftler et~al\mbox{.}(2020)]%
        {9148111}
\bibfield{author}{\bibinfo{person}{Bekir~Sait Ciftler}, \bibinfo{person}{Abdullatif Albaseer}, \bibinfo{person}{Noureddine Lasla}, {and} \bibinfo{person}{Mohamed Abdallah}.} \bibinfo{year}{2020}\natexlab{}.
\newblock \showarticletitle{Federated Learning for RSS Fingerprint-based Localization: A Privacy-Preserving Crowdsourcing Method}. In \bibinfo{booktitle}{\emph{2020 International Wireless Communications and Mobile Computing (IWCMC)}}. \bibinfo{pages}{2112--2117}.
\newblock
\urldef\tempurl%
\url{https://doi.org/10.1109/IWCMC48107.2020.9148111}
\showDOI{\tempurl}


\bibitem[Code(1998)]%
        {code1998acm}
\bibfield{author}{\bibinfo{person}{Generate Code}.} \bibinfo{year}{1998}\natexlab{}.
\newblock \showarticletitle{Acm computing classification system}.
\newblock  (\bibinfo{year}{1998}).
\newblock


\bibitem[Concone et~al\mbox{.}(2022)]%
        {9821084}
\bibfield{author}{\bibinfo{person}{Federico Concone}, \bibinfo{person}{Cedric Ferdico}, \bibinfo{person}{Giuseppe~Lo Re}, {and} \bibinfo{person}{Marco Morana}.} \bibinfo{year}{2022}\natexlab{}.
\newblock \showarticletitle{A Federated Learning Approach for Distributed Human Activity Recognition}. In \bibinfo{booktitle}{\emph{2022 IEEE International Conference on Smart Computing (SMARTCOMP)}}. \bibinfo{pages}{269--274}.
\newblock
\urldef\tempurl%
\url{https://doi.org/10.1109/SMARTCOMP55677.2022.00066}
\showDOI{\tempurl}


\bibitem[Cui et~al\mbox{.}(2021)]%
        {9414305}
\bibfield{author}{\bibinfo{person}{Xiaodong Cui}, \bibinfo{person}{Songtao Lu}, {and} \bibinfo{person}{Brian Kingsbury}.} \bibinfo{year}{2021}\natexlab{}.
\newblock \showarticletitle{Federated Acoustic Modeling for Automatic Speech Recognition}. In \bibinfo{booktitle}{\emph{ICASSP 2021 - 2021 IEEE International Conference on Acoustics, Speech and Signal Processing (ICASSP)}}. \bibinfo{pages}{6748--6752}.
\newblock
\urldef\tempurl%
\url{https://doi.org/10.1109/ICASSP39728.2021.9414305}
\showDOI{\tempurl}


\bibitem[Cui et~al\mbox{.}(2022)]%
        {9871861}
\bibfield{author}{\bibinfo{person}{Yue Cui}, \bibinfo{person}{Zhuohang Li}, \bibinfo{person}{Luyang Liu}, \bibinfo{person}{Jiaxin Zhang}, {and} \bibinfo{person}{Jian Liu}.} \bibinfo{year}{2022}\natexlab{}.
\newblock \showarticletitle{Privacy-preserving Speech-based Depression Diagnosis via Federated Learning}. In \bibinfo{booktitle}{\emph{2022 44th Annual International Conference of the IEEE Engineering in Medicine \& Biology Society (EMBC)}}. \bibinfo{pages}{1371--1374}.
\newblock
\urldef\tempurl%
\url{https://doi.org/10.1109/EMBC48229.2022.9871861}
\showDOI{\tempurl}


\bibitem[de~S.~Silva et~al\mbox{.}(2023)]%
        {10.1007/978-3-031-45392-2_25}
\bibfield{author}{\bibinfo{person}{Victor~E. de S.~Silva}, \bibinfo{person}{Tiago~B. Lacerda}, \bibinfo{person}{P{\'e}ricles Miranda}, \bibinfo{person}{Andr{\'e} C{\^a}mara}, \bibinfo{person}{Amerson Riley~Cabral Chagas}, {and} \bibinfo{person}{Ana Paula~C. Furtado}.} \bibinfo{year}{2023}\natexlab{}.
\newblock \showarticletitle{Federated Learning and Mel-Spectrograms for Physical Violence Detection in Audio}. In \bibinfo{booktitle}{\emph{Intelligent Systems}}, \bibfield{editor}{\bibinfo{person}{Murilo~C. Naldi} {and} \bibinfo{person}{Reinaldo A.~C. Bianchi}} (Eds.). \bibinfo{publisher}{Springer Nature Switzerland}, \bibinfo{address}{Cham}, \bibinfo{pages}{379--393}.
\newblock
\showISBNx{978-3-031-45392-2}


\bibitem[Diao et~al\mbox{.}(2023)]%
        {diao2023semi}
\bibfield{author}{\bibinfo{person}{Enmao Diao}, \bibinfo{person}{Eric~W Tramel}, \bibinfo{person}{Jie Ding}, {and} \bibinfo{person}{Tao Zhang}.} \bibinfo{year}{2023}\natexlab{}.
\newblock \showarticletitle{Semi-supervised federated learning for keyword spotting}.
\newblock \bibinfo{journal}{\emph{arXiv preprint arXiv:2305.05110}} (\bibinfo{year}{2023}).
\newblock


\bibitem[Dimitriadis et~al\mbox{.}(2020)]%
        {dimitriadis2020federated}
\bibfield{author}{\bibinfo{person}{Dimitrios Dimitriadis}, \bibinfo{person}{Robert~Gmyr Ken'ichi~Kumatani}, \bibinfo{person}{Robert Gmyr}, \bibinfo{person}{Yashesh Gaur}, {and} \bibinfo{person}{Sefik~Emre Eskimez}.} \bibinfo{year}{2020}\natexlab{}.
\newblock \showarticletitle{A Federated Approach in Training Acoustic Models.}. In \bibinfo{booktitle}{\emph{Interspeech}}. \bibinfo{pages}{981--985}.
\newblock


\bibitem[Doherty and Doherty(2018)]%
        {10.1145/3234149}
\bibfield{author}{\bibinfo{person}{Kevin Doherty} {and} \bibinfo{person}{Gavin Doherty}.} \bibinfo{year}{2018}\natexlab{}.
\newblock \showarticletitle{Engagement in HCI: Conception, Theory and Measurement}.
\newblock \bibinfo{journal}{\emph{ACM Comput. Surv.}} \bibinfo{volume}{51}, \bibinfo{number}{5}, Article \bibinfo{articleno}{99} (\bibinfo{date}{nov} \bibinfo{year}{2018}), \bibinfo{numpages}{39}~pages.
\newblock
\showISSN{0360-0300}
\urldef\tempurl%
\url{https://doi.org/10.1145/3234149}
\showDOI{\tempurl}


\bibitem[Doshi and Yilmaz(2022)]%
        {Doshi_2022_CVPR}
\bibfield{author}{\bibinfo{person}{Keval Doshi} {and} \bibinfo{person}{Yasin Yilmaz}.} \bibinfo{year}{2022}\natexlab{}.
\newblock \showarticletitle{Federated Learning-Based Driver Activity Recognition for Edge Devices}. In \bibinfo{booktitle}{\emph{Proceedings of the IEEE/CVF Conference on Computer Vision and Pattern Recognition (CVPR) Workshops}}. \bibinfo{pages}{3338--3346}.
\newblock


\bibitem[Dou et~al\mbox{.}(2023)]%
        {10214616}
\bibfield{author}{\bibinfo{person}{Fei Dou}, \bibinfo{person}{Jin Lu}, \bibinfo{person}{Tan Zhu}, {and} \bibinfo{person}{Jinbo Bi}.} \bibinfo{year}{2023}\natexlab{}.
\newblock \showarticletitle{On-Device Indoor Positioning: A Federated Reinforcement Learning Approach With Heterogeneous Devices}.
\newblock \bibinfo{journal}{\emph{IEEE Internet of Things Journal}} (\bibinfo{year}{2023}), \bibinfo{pages}{1--1}.
\newblock
\urldef\tempurl%
\url{https://doi.org/10.1109/JIOT.2023.3299262}
\showDOI{\tempurl}


\bibitem[Du et~al\mbox{.}(2023)]%
        {10186757}
\bibfield{author}{\bibinfo{person}{Runjia Du}, \bibinfo{person}{Kyungtae Han}, \bibinfo{person}{Rohit Gupta}, \bibinfo{person}{Sikai Chen}, \bibinfo{person}{Samuel Labi}, {and} \bibinfo{person}{Ziran Wang}.} \bibinfo{year}{2023}\natexlab{}.
\newblock \showarticletitle{Driver Monitoring-Based Lane-Change Prediction: A Personalized Federated Learning Framework}. In \bibinfo{booktitle}{\emph{2023 IEEE Intelligent Vehicles Symposium (IV)}}. \bibinfo{pages}{1--7}.
\newblock
\urldef\tempurl%
\url{https://doi.org/10.1109/IV55152.2023.10186757}
\showDOI{\tempurl}


\bibitem[Du et~al\mbox{.}(2020)]%
        {9086790}
\bibfield{author}{\bibinfo{person}{Zhaoyang Du}, \bibinfo{person}{Celimuge Wu}, \bibinfo{person}{Tsutomu Yoshinaga}, \bibinfo{person}{Kok-Lim~Alvin Yau}, \bibinfo{person}{Yusheng Ji}, {and} \bibinfo{person}{Jie Li}.} \bibinfo{year}{2020}\natexlab{}.
\newblock \showarticletitle{Federated Learning for Vehicular Internet of Things: Recent Advances and Open Issues}.
\newblock \bibinfo{journal}{\emph{IEEE Open Journal of the Computer Society}}  \bibinfo{volume}{1} (\bibinfo{year}{2020}), \bibinfo{pages}{45--61}.
\newblock
\urldef\tempurl%
\url{https://doi.org/10.1109/OJCS.2020.2992630}
\showDOI{\tempurl}


\bibitem[Ek et~al\mbox{.}(2020)]%
        {10.1145/3410530.3414321}
\bibfield{author}{\bibinfo{person}{Sannara Ek}, \bibinfo{person}{Fran\c{c}ois Portet}, \bibinfo{person}{Philippe Lalanda}, {and} \bibinfo{person}{German Vega}.} \bibinfo{year}{2020}\natexlab{}.
\newblock \showarticletitle{Evaluation of Federated Learning Aggregation Algorithms: Application to Human Activity Recognition}. In \bibinfo{booktitle}{\emph{Adjunct Proceedings of the 2020 ACM International Joint Conference on Pervasive and Ubiquitous Computing and Proceedings of the 2020 ACM International Symposium on Wearable Computers}} (Virtual Event, Mexico) \emph{(\bibinfo{series}{UbiComp/ISWC '20 Adjunct})}. \bibinfo{publisher}{Association for Computing Machinery}, \bibinfo{address}{New York, NY, USA}, \bibinfo{pages}{638–643}.
\newblock
\showISBNx{9781450380768}
\urldef\tempurl%
\url{https://doi.org/10.1145/3410530.3414321}
\showDOI{\tempurl}


\bibitem[EK et~al\mbox{.}(2021)]%
        {9439129}
\bibfield{author}{\bibinfo{person}{Sannara EK}, \bibinfo{person}{François PORTET}, \bibinfo{person}{Philippe LALANDA}, {and} \bibinfo{person}{German VEGA}.} \bibinfo{year}{2021}\natexlab{}.
\newblock \showarticletitle{A Federated Learning Aggregation Algorithm for Pervasive Computing: Evaluation and Comparison}. In \bibinfo{booktitle}{\emph{2021 IEEE International Conference on Pervasive Computing and Communications (PerCom)}}. \bibinfo{pages}{1--10}.
\newblock
\urldef\tempurl%
\url{https://doi.org/10.1109/PERCOM50583.2021.9439129}
\showDOI{\tempurl}


\bibitem[Ek et~al\mbox{.}(2022a)]%
        {ek2022evaluation}
\bibfield{author}{\bibinfo{person}{Sannara Ek}, \bibinfo{person}{Fran{\c{c}}ois Portet}, \bibinfo{person}{Philippe Lalanda}, {and} \bibinfo{person}{German Vega}.} \bibinfo{year}{2022}\natexlab{a}.
\newblock \showarticletitle{Evaluation and comparison of federated learning algorithms for Human Activity Recognition on smartphones}.
\newblock \bibinfo{journal}{\emph{Pervasive and Mobile Computing}}  \bibinfo{volume}{87} (\bibinfo{year}{2022}), \bibinfo{pages}{101714}.
\newblock


\bibitem[Ek et~al\mbox{.}(2022b)]%
        {9767369}
\bibfield{author}{\bibinfo{person}{Sannara Ek}, \bibinfo{person}{Romain Rombourg}, \bibinfo{person}{François Portet}, {and} \bibinfo{person}{Philippe Lalanda}.} \bibinfo{year}{2022}\natexlab{b}.
\newblock \showarticletitle{Federated Self-Supervised Learning in Heterogeneous Settings: Limits of a Baseline Approach on HAR}. In \bibinfo{booktitle}{\emph{2022 IEEE International Conference on Pervasive Computing and Communications Workshops and other Affiliated Events (PerCom Workshops)}}. \bibinfo{pages}{557--562}.
\newblock
\urldef\tempurl%
\url{https://doi.org/10.1109/PerComWorkshops53856.2022.9767369}
\showDOI{\tempurl}


\bibitem[Ekman et~al\mbox{.}(2013)]%
        {ekman2013emotion}
\bibfield{author}{\bibinfo{person}{Paul Ekman}, \bibinfo{person}{Wallace~V Friesen}, {and} \bibinfo{person}{Phoebe Ellsworth}.} \bibinfo{year}{2013}\natexlab{}.
\newblock \bibinfo{booktitle}{\emph{Emotion in the human face: Guidelines for research and an integration of findings}}. Vol.~\bibinfo{volume}{11}.
\newblock \bibinfo{publisher}{Elsevier}.
\newblock


\bibitem[El~Ayadi et~al\mbox{.}(2011)]%
        {el2011survey}
\bibfield{author}{\bibinfo{person}{Moataz El~Ayadi}, \bibinfo{person}{Mohamed~S Kamel}, {and} \bibinfo{person}{Fakhri Karray}.} \bibinfo{year}{2011}\natexlab{}.
\newblock \showarticletitle{Survey on speech emotion recognition: Features, classification schemes, and databases}.
\newblock \bibinfo{journal}{\emph{Pattern recognition}} \bibinfo{volume}{44}, \bibinfo{number}{3} (\bibinfo{year}{2011}), \bibinfo{pages}{572--587}.
\newblock


\bibitem[Emami et~al\mbox{.}(2023)]%
        {10223407}
\bibfield{author}{\bibinfo{person}{Negar Emami}, \bibinfo{person}{Antonio~Di Maio}, {and} \bibinfo{person}{Torsten Braun}.} \bibinfo{year}{2023}\natexlab{}.
\newblock \showarticletitle{FedForce: Network-adaptive Federated Learning for Reinforced Mobility Prediction}. In \bibinfo{booktitle}{\emph{2023 IEEE 48th Conference on Local Computer Networks (LCN)}}. \bibinfo{pages}{1--9}.
\newblock
\urldef\tempurl%
\url{https://doi.org/10.1109/LCN58197.2023.10223407}
\showDOI{\tempurl}


\bibitem[Errounda and Liu(2022)]%
        {errounda2022mobility}
\bibfield{author}{\bibinfo{person}{Fatima~Zahra Errounda} {and} \bibinfo{person}{Yan Liu}.} \bibinfo{year}{2022}\natexlab{}.
\newblock \showarticletitle{A Mobility Forecasting Framework with Vertical Federated Learning}. In \bibinfo{booktitle}{\emph{2022 IEEE 46th Annual Computers, Software, and Applications Conference (COMPSAC)}}. IEEE, \bibinfo{pages}{301--310}.
\newblock


\bibitem[Etiabi et~al\mbox{.}(2022)]%
        {etiabi2022federated}
\bibfield{author}{\bibinfo{person}{Yaya Etiabi}, \bibinfo{person}{Marwa Chafii}, {and} \bibinfo{person}{El~Mehdi Amhoud}.} \bibinfo{year}{2022}\natexlab{}.
\newblock \showarticletitle{Federated Distillation based Indoor Localization for IoT Networks}.
\newblock \bibinfo{journal}{\emph{arXiv preprint arXiv:2205.11440}} (\bibinfo{year}{2022}).
\newblock


\bibitem[Etiabi et~al\mbox{.}(2023)]%
        {10118848}
\bibfield{author}{\bibinfo{person}{Yaya Etiabi}, \bibinfo{person}{Wafa Njima}, {and} \bibinfo{person}{El~Mehdi Amhoud}.} \bibinfo{year}{2023}\natexlab{}.
\newblock \showarticletitle{Federated Learning based Hierarchical 3D Indoor Localization}. In \bibinfo{booktitle}{\emph{2023 IEEE Wireless Communications and Networking Conference (WCNC)}}. \bibinfo{pages}{1--6}.
\newblock
\urldef\tempurl%
\url{https://doi.org/10.1109/WCNC55385.2023.10118848}
\showDOI{\tempurl}


\bibitem[Ezequiel et~al\mbox{.}(2022)]%
        {ezequiel2022federated}
\bibfield{author}{\bibinfo{person}{Castro Elizondo~Jose Ezequiel}, \bibinfo{person}{Martin Gjoreski}, {and} \bibinfo{person}{Marc Langheinrich}.} \bibinfo{year}{2022}\natexlab{}.
\newblock \showarticletitle{Federated learning for privacy-aware human mobility modeling}.
\newblock \bibinfo{journal}{\emph{Frontiers in Artificial Intelligence}}  \bibinfo{volume}{5} (\bibinfo{year}{2022}), \bibinfo{pages}{867046}.
\newblock


\bibitem[Fallah et~al\mbox{.}(2020)]%
        {NEURIPS2020_24389bfe}
\bibfield{author}{\bibinfo{person}{Alireza Fallah}, \bibinfo{person}{Aryan Mokhtari}, {and} \bibinfo{person}{Asuman Ozdaglar}.} \bibinfo{year}{2020}\natexlab{}.
\newblock \showarticletitle{Personalized Federated Learning with Theoretical Guarantees: A Model-Agnostic Meta-Learning Approach}. In \bibinfo{booktitle}{\emph{Advances in Neural Information Processing Systems}}, \bibfield{editor}{\bibinfo{person}{H.~Larochelle}, \bibinfo{person}{M.~Ranzato}, \bibinfo{person}{R.~Hadsell}, \bibinfo{person}{M.F. Balcan}, {and} \bibinfo{person}{H.~Lin}} (Eds.), Vol.~\bibinfo{volume}{33}. \bibinfo{publisher}{Curran Associates, Inc.}, \bibinfo{pages}{3557--3568}.
\newblock
\urldef\tempurl%
\url{https://proceedings.neurips.cc/paper_files/paper/2020/file/24389bfe4fe2eba8bf9aa9203a44cdad-Paper.pdf}
\showURL{%
\tempurl}


\bibitem[Fan et~al\mbox{.}(2020)]%
        {10.1145/3369830}
\bibfield{author}{\bibinfo{person}{Zipei Fan}, \bibinfo{person}{Xuan Song}, \bibinfo{person}{Renhe Jiang}, \bibinfo{person}{Quanjun Chen}, {and} \bibinfo{person}{Ryosuke Shibasaki}.} \bibinfo{year}{2020}\natexlab{}.
\newblock \showarticletitle{Decentralized Attention-Based Personalized Human Mobility Prediction}.
\newblock \bibinfo{journal}{\emph{Proc. ACM Interact. Mob. Wearable Ubiquitous Technol.}} \bibinfo{volume}{3}, \bibinfo{number}{4}, Article \bibinfo{articleno}{133} (\bibinfo{date}{sep} \bibinfo{year}{2020}), \bibinfo{numpages}{26}~pages.
\newblock
\urldef\tempurl%
\url{https://doi.org/10.1145/3369830}
\showDOI{\tempurl}


\bibitem[Fang et~al\mbox{.}(2021)]%
        {fang2021bayesian}
\bibfield{author}{\bibinfo{person}{Lei Fang}, \bibinfo{person}{Xiaoli Liu}, \bibinfo{person}{Xiang Su}, \bibinfo{person}{Juan Ye}, \bibinfo{person}{Simon Dobson}, \bibinfo{person}{Pan Hui}, {and} \bibinfo{person}{Sasu Tarkoma}.} \bibinfo{year}{2021}\natexlab{}.
\newblock \showarticletitle{Bayesian inference federated learning for heart rate prediction}. In \bibinfo{booktitle}{\emph{Wireless Mobile Communication and Healthcare: 9th EAI International Conference, MobiHealth 2020, Virtual Event, November 19, 2020, Proceedings 9}}. Springer, \bibinfo{pages}{116--130}.
\newblock


\bibitem[Farahani et~al\mbox{.}(2023)]%
        {10175520}
\bibfield{author}{\bibinfo{person}{Bahar Farahani}, \bibinfo{person}{Shima Tabibian}, {and} \bibinfo{person}{Hamid Ebrahimi}.} \bibinfo{year}{2023}\natexlab{}.
\newblock \showarticletitle{Toward a Personalized Clustered Federated Learning: A Speech Recognition Case Study}.
\newblock \bibinfo{journal}{\emph{IEEE Internet of Things Journal}} \bibinfo{volume}{10}, \bibinfo{number}{21} (\bibinfo{year}{2023}), \bibinfo{pages}{18553--18562}.
\newblock
\urldef\tempurl%
\url{https://doi.org/10.1109/JIOT.2023.3292797}
\showDOI{\tempurl}


\bibitem[Fauzi et~al\mbox{.}(2022)]%
        {fauzi2022comparative}
\bibfield{author}{\bibinfo{person}{Muhammad~Ali Fauzi}, \bibinfo{person}{Bian Yang}, {and} \bibinfo{person}{Bernd Blobel}.} \bibinfo{year}{2022}\natexlab{}.
\newblock \showarticletitle{Comparative analysis between individual, centralized, and federated learning for smartwatch based stress detection}.
\newblock \bibinfo{journal}{\emph{Journal of Personalized Medicine}} \bibinfo{volume}{12}, \bibinfo{number}{10} (\bibinfo{year}{2022}), \bibinfo{pages}{1584}.
\newblock


\bibitem[Feng et~al\mbox{.}(2020)]%
        {10.1145/3381006}
\bibfield{author}{\bibinfo{person}{Jie Feng}, \bibinfo{person}{Can Rong}, \bibinfo{person}{Funing Sun}, \bibinfo{person}{Diansheng Guo}, {and} \bibinfo{person}{Yong Li}.} \bibinfo{year}{2020}\natexlab{}.
\newblock \showarticletitle{PMF: A Privacy-Preserving Human Mobility Prediction Framework via Federated Learning}.
\newblock \bibinfo{journal}{\emph{Proc. ACM Interact. Mob. Wearable Ubiquitous Technol.}} \bibinfo{volume}{4}, \bibinfo{number}{1}, Article \bibinfo{articleno}{10} (\bibinfo{date}{mar} \bibinfo{year}{2020}), \bibinfo{numpages}{21}~pages.
\newblock
\urldef\tempurl%
\url{https://doi.org/10.1145/3381006}
\showDOI{\tempurl}


\bibitem[Feng et~al\mbox{.}(2021)]%
        {feng2021attribute}
\bibfield{author}{\bibinfo{person}{Tiantian Feng}, \bibinfo{person}{Hanieh Hashemi}, \bibinfo{person}{Rajat Hebbar}, \bibinfo{person}{Murali Annavaram}, {and} \bibinfo{person}{Shrikanth~S Narayanan}.} \bibinfo{year}{2021}\natexlab{}.
\newblock \showarticletitle{Attribute inference attack of speech emotion recognition in federated learning settings}.
\newblock \bibinfo{journal}{\emph{arXiv preprint arXiv:2112.13416}} (\bibinfo{year}{2021}).
\newblock


\bibitem[Feng and Narayanan(2022)]%
        {feng2022semi}
\bibfield{author}{\bibinfo{person}{Tiantian Feng} {and} \bibinfo{person}{Shrikanth Narayanan}.} \bibinfo{year}{2022}\natexlab{}.
\newblock \showarticletitle{Semi-fedser: Semi-supervised learning for speech emotion recognition on federated learning using multiview pseudo-labeling}.
\newblock \bibinfo{journal}{\emph{arXiv preprint arXiv:2203.08810}} (\bibinfo{year}{2022}).
\newblock


\bibitem[Feng et~al\mbox{.}(2022)]%
        {feng2022user}
\bibfield{author}{\bibinfo{person}{Tiantian Feng}, \bibinfo{person}{Raghuveer Peri}, {and} \bibinfo{person}{Shrikanth Narayanan}.} \bibinfo{year}{2022}\natexlab{}.
\newblock \showarticletitle{User-level differential privacy against attribute inference attack of speech emotion recognition in federated learning}.
\newblock \bibinfo{journal}{\emph{arXiv preprint arXiv:2204.02500}} (\bibinfo{year}{2022}).
\newblock


\bibitem[Fenoglio et~al\mbox{.}(2023a)]%
        {10.1145/3626705.3631796}
\bibfield{author}{\bibinfo{person}{Dario Fenoglio}, \bibinfo{person}{Martin Gjoreski}, {and} \bibinfo{person}{Marc Langheinrich}.} \bibinfo{year}{2023}\natexlab{a}.
\newblock \showarticletitle{A Federated Unsupervised Personalisation for Cognitive Workload Estimation}. In \bibinfo{booktitle}{\emph{Proceedings of the 22nd International Conference on Mobile and Ubiquitous Multimedia}} (<conf-loc>, <city>Vienna</city>, <country>Austria</country>, </conf-loc>) \emph{(\bibinfo{series}{MUM '23})}. \bibinfo{publisher}{Association for Computing Machinery}, \bibinfo{address}{New York, NY, USA}, \bibinfo{pages}{526–528}.
\newblock
\showISBNx{9798400709210}
\urldef\tempurl%
\url{https://doi.org/10.1145/3626705.3631796}
\showDOI{\tempurl}


\bibitem[Fenoglio et~al\mbox{.}(2023b)]%
        {10.1145/3626705.3627783}
\bibfield{author}{\bibinfo{person}{Dario Fenoglio}, \bibinfo{person}{Daniel Josifovski}, \bibinfo{person}{Alessandro Gobbetti}, \bibinfo{person}{Mattias Formo}, \bibinfo{person}{Hristijan Gjoreski}, \bibinfo{person}{Martin Gjoreski}, {and} \bibinfo{person}{Marc Langheinrich}.} \bibinfo{year}{2023}\natexlab{b}.
\newblock \showarticletitle{Federated Learning for Privacy-Aware Cognitive Workload Estimation}. In \bibinfo{booktitle}{\emph{Proceedings of the 22nd International Conference on Mobile and Ubiquitous Multimedia}} (<conf-loc>, <city>Vienna</city>, <country>Austria</country>, </conf-loc>) \emph{(\bibinfo{series}{MUM '23})}. \bibinfo{publisher}{Association for Computing Machinery}, \bibinfo{address}{New York, NY, USA}, \bibinfo{pages}{25–36}.
\newblock
\showISBNx{9798400709210}
\urldef\tempurl%
\url{https://doi.org/10.1145/3626705.3627783}
\showDOI{\tempurl}


\bibitem[Fernandes et~al\mbox{.}(2022)]%
        {10.1145/3491208}
\bibfield{author}{\bibinfo{person}{Jos\'{e}~Marcelo Fernandes}, \bibinfo{person}{Jorge~S\'{a} Silva}, \bibinfo{person}{Andr\'{e} Rodrigues}, {and} \bibinfo{person}{Fernando Boavida}.} \bibinfo{year}{2022}\natexlab{}.
\newblock \showarticletitle{A Survey of Approaches to Unobtrusive Sensing of Humans}.
\newblock \bibinfo{journal}{\emph{ACM Comput. Surv.}} \bibinfo{volume}{55}, \bibinfo{number}{2}, Article \bibinfo{articleno}{41} (\bibinfo{date}{jan} \bibinfo{year}{2022}), \bibinfo{numpages}{28}~pages.
\newblock
\showISSN{0360-0300}
\urldef\tempurl%
\url{https://doi.org/10.1145/3491208}
\showDOI{\tempurl}


\bibitem[Fung et~al\mbox{.}(2018)]%
        {fung2018mitigating}
\bibfield{author}{\bibinfo{person}{Clement Fung}, \bibinfo{person}{Chris~JM Yoon}, {and} \bibinfo{person}{Ivan Beschastnikh}.} \bibinfo{year}{2018}\natexlab{}.
\newblock \showarticletitle{Mitigating sybils in federated learning poisoning}.
\newblock \bibinfo{journal}{\emph{arXiv preprint arXiv:1808.04866}} (\bibinfo{year}{2018}).
\newblock


\bibitem[Gahlan and Sethia(2023)]%
        {gahlan2023federated}
\bibfield{author}{\bibinfo{person}{Neha Gahlan} {and} \bibinfo{person}{Divyashikha Sethia}.} \bibinfo{year}{2023}\natexlab{}.
\newblock \showarticletitle{Federated learning inspired privacy sensitive emotion recognition based on multi-modal physiological sensors}.
\newblock \bibinfo{journal}{\emph{Cluster Computing}} (\bibinfo{year}{2023}), \bibinfo{pages}{1--23}.
\newblock


\bibitem[Ganju et~al\mbox{.}(2018)]%
        {10.1145/3243734.3243834}
\bibfield{author}{\bibinfo{person}{Karan Ganju}, \bibinfo{person}{Qi Wang}, \bibinfo{person}{Wei Yang}, \bibinfo{person}{Carl~A. Gunter}, {and} \bibinfo{person}{Nikita Borisov}.} \bibinfo{year}{2018}\natexlab{}.
\newblock \showarticletitle{Property Inference Attacks on Fully Connected Neural Networks using Permutation Invariant Representations}. In \bibinfo{booktitle}{\emph{Proceedings of the 2018 ACM SIGSAC Conference on Computer and Communications Security}} (Toronto, Canada) \emph{(\bibinfo{series}{CCS '18})}. \bibinfo{publisher}{Association for Computing Machinery}, \bibinfo{address}{New York, NY, USA}, \bibinfo{pages}{619–633}.
\newblock
\showISBNx{9781450356930}
\urldef\tempurl%
\url{https://doi.org/10.1145/3243734.3243834}
\showDOI{\tempurl}


\bibitem[Gao et~al\mbox{.}(2023)]%
        {9917443}
\bibfield{author}{\bibinfo{person}{Bo Gao}, \bibinfo{person}{Fan Yang}, \bibinfo{person}{Nan Cui}, \bibinfo{person}{Ke Xiong}, \bibinfo{person}{Yang Lu}, {and} \bibinfo{person}{Yuwei Wang}.} \bibinfo{year}{2023}\natexlab{}.
\newblock \showarticletitle{A Federated Learning Framework for Fingerprinting-Based Indoor Localization in Multibuilding and Multifloor Environments}.
\newblock \bibinfo{journal}{\emph{IEEE Internet of Things Journal}} \bibinfo{volume}{10}, \bibinfo{number}{3} (\bibinfo{year}{2023}), \bibinfo{pages}{2615--2629}.
\newblock
\urldef\tempurl%
\url{https://doi.org/10.1109/JIOT.2022.3214211}
\showDOI{\tempurl}


\bibitem[Gao and Konomi(2023)]%
        {10.1145/3594739.3610739}
\bibfield{author}{\bibinfo{person}{Lulu Gao} {and} \bibinfo{person}{Shin'ichi Konomi}.} \bibinfo{year}{2023}\natexlab{}.
\newblock \showarticletitle{Personalized Federated Human Activity Recognition through Semi-Supervised Learning and Enhanced Representation}. In \bibinfo{booktitle}{\emph{Adjunct Proceedings of the 2023 ACM International Joint Conference on Pervasive and Ubiquitous Computing \& the 2023 ACM International Symposium on Wearable Computing}} (<conf-loc>, <city>Cancun, Quintana Roo</city>, <country>Mexico</country>, </conf-loc>) \emph{(\bibinfo{series}{UbiComp/ISWC '23 Adjunct})}. \bibinfo{publisher}{Association for Computing Machinery}, \bibinfo{address}{New York, NY, USA}, \bibinfo{pages}{463–468}.
\newblock
\showISBNx{9798400702006}
\urldef\tempurl%
\url{https://doi.org/10.1145/3594739.3610739}
\showDOI{\tempurl}


\bibitem[Gao et~al\mbox{.}(2022)]%
        {9747161}
\bibfield{author}{\bibinfo{person}{Yan Gao}, \bibinfo{person}{Titouan Parcollet}, \bibinfo{person}{Salah Zaiem}, \bibinfo{person}{Javier Fernandez-Marques}, \bibinfo{person}{Pedro P.~B. de Gusmao}, \bibinfo{person}{Daniel~J. Beutel}, {and} \bibinfo{person}{Nicholas~D. Lane}.} \bibinfo{year}{2022}\natexlab{}.
\newblock \showarticletitle{End-to-End Speech Recognition from Federated Acoustic Models}. In \bibinfo{booktitle}{\emph{ICASSP 2022 - 2022 IEEE International Conference on Acoustics, Speech and Signal Processing (ICASSP)}}. \bibinfo{pages}{7227--7231}.
\newblock
\urldef\tempurl%
\url{https://doi.org/10.1109/ICASSP43922.2022.9747161}
\showDOI{\tempurl}


\bibitem[Ghosh et~al\mbox{.}(2020)]%
        {ghosh2020efficient}
\bibfield{author}{\bibinfo{person}{Avishek Ghosh}, \bibinfo{person}{Jichan Chung}, \bibinfo{person}{Dong Yin}, {and} \bibinfo{person}{Kannan Ramchandran}.} \bibinfo{year}{2020}\natexlab{}.
\newblock \showarticletitle{An efficient framework for clustered federated learning}.
\newblock \bibinfo{journal}{\emph{Advances in Neural Information Processing Systems}}  \bibinfo{volume}{33} (\bibinfo{year}{2020}), \bibinfo{pages}{19586--19597}.
\newblock


\bibitem[Girija et~al\mbox{.}(2023)]%
        {GIRIJA2023100793}
\bibfield{author}{\bibinfo{person}{Shini Girija}, \bibinfo{person}{Thar Baker}, \bibinfo{person}{Naveed Ahmed}, \bibinfo{person}{Ahmed~M. Khedr}, \bibinfo{person}{Zaher {Al Aghbari}}, \bibinfo{person}{Ashish Jha}, \bibinfo{person}{Konstantin Sobolev}, \bibinfo{person}{Salman~Ahmadi Asl}, {and} \bibinfo{person}{Anh-Huy Phan}.} \bibinfo{year}{2023}\natexlab{}.
\newblock \showarticletitle{Attribute recognition for person re-identification using federated learning at all-in-edge}.
\newblock \bibinfo{journal}{\emph{Internet of Things}}  \bibinfo{volume}{22} (\bibinfo{year}{2023}), \bibinfo{pages}{100793}.
\newblock
\showISSN{2542-6605}
\urldef\tempurl%
\url{https://doi.org/10.1016/j.iot.2023.100793}
\showDOI{\tempurl}


\bibitem[Gjoreski et~al\mbox{.}(2022)]%
        {gjoreski2022toward}
\bibfield{author}{\bibinfo{person}{Martin Gjoreski}, \bibinfo{person}{Mat{\'\i}as Laporte}, {and} \bibinfo{person}{Marc Langheinrich}.} \bibinfo{year}{2022}\natexlab{}.
\newblock \showarticletitle{Toward privacy-aware federated analytics of cohorts for smart mobility}.
\newblock \bibinfo{journal}{\emph{Frontiers in Computer Science}}  \bibinfo{volume}{4} (\bibinfo{year}{2022}), \bibinfo{pages}{891206}.
\newblock


\bibitem[G{\"o}n{\"u}l et~al\mbox{.}(2022)]%
        {gonul2022human}
\bibfield{author}{\bibinfo{person}{Tansel G{\"o}n{\"u}l}, \bibinfo{person}{Ozlem~Durmaz Incel}, {and} \bibinfo{person}{Gulfem Isiklar~Alptekin}.} \bibinfo{year}{2022}\natexlab{}.
\newblock \showarticletitle{Human Activity Recognition with Smart Watches Using Federated Learning}. In \bibinfo{booktitle}{\emph{International Conference on Intelligent and Fuzzy Systems}}. Springer, \bibinfo{pages}{77--85}.
\newblock


\bibitem[Granqvist et~al\mbox{.}(2020)]%
        {granqvist2020improving}
\bibfield{author}{\bibinfo{person}{Filip Granqvist}, \bibinfo{person}{Matt Seigel}, \bibinfo{person}{Rogier Van~Dalen}, \bibinfo{person}{Aine Cahill}, \bibinfo{person}{Stephen Shum}, {and} \bibinfo{person}{Matthias Paulik}.} \bibinfo{year}{2020}\natexlab{}.
\newblock \showarticletitle{Improving on-device speaker verification using federated learning with privacy}.
\newblock \bibinfo{journal}{\emph{arXiv preprint arXiv:2008.02651}} (\bibinfo{year}{2020}).
\newblock


\bibitem[Gudur and Perepu(2021a)]%
        {gudur2021resource}
\bibfield{author}{\bibinfo{person}{Gautham~Krishna Gudur} {and} \bibinfo{person}{Satheesh~Kumar Perepu}.} \bibinfo{year}{2021}\natexlab{a}.
\newblock \showarticletitle{Resource-constrained federated learning with heterogeneous labels and models for human activity recognition}. In \bibinfo{booktitle}{\emph{International Workshop on Deep Learning for Human Activity Recognition}}. Springer, \bibinfo{pages}{57--69}.
\newblock


\bibitem[Gudur and Perepu(2021b)]%
        {gudur2021zero}
\bibfield{author}{\bibinfo{person}{Gautham~Krishna Gudur} {and} \bibinfo{person}{Satheesh~K Perepu}.} \bibinfo{year}{2021}\natexlab{b}.
\newblock \showarticletitle{Zero-shot federated learning with new classes for audio classification}.
\newblock \bibinfo{journal}{\emph{arXiv preprint arXiv:2106.10019}} (\bibinfo{year}{2021}).
\newblock


\bibitem[Gufran and Pasricha(2023)]%
        {10.1145/3607919}
\bibfield{author}{\bibinfo{person}{Danish Gufran} {and} \bibinfo{person}{Sudeep Pasricha}.} \bibinfo{year}{2023}\natexlab{}.
\newblock \showarticletitle{FedHIL: Heterogeneity Resilient Federated Learning for Robust Indoor Localization with Mobile Devices}.
\newblock \bibinfo{journal}{\emph{ACM Trans. Embed. Comput. Syst.}} \bibinfo{volume}{22}, \bibinfo{number}{5s}, Article \bibinfo{articleno}{125} (\bibinfo{date}{sep} \bibinfo{year}{2023}), \bibinfo{numpages}{24}~pages.
\newblock
\showISSN{1539-9087}
\urldef\tempurl%
\url{https://doi.org/10.1145/3607919}
\showDOI{\tempurl}


\bibitem[Guliani et~al\mbox{.}(2021)]%
        {9413397}
\bibfield{author}{\bibinfo{person}{Dhruv Guliani}, \bibinfo{person}{Françoise Beaufays}, {and} \bibinfo{person}{Giovanni Motta}.} \bibinfo{year}{2021}\natexlab{}.
\newblock \showarticletitle{Training Speech Recognition Models with Federated Learning: A Quality/Cost Framework}. In \bibinfo{booktitle}{\emph{ICASSP 2021 - 2021 IEEE International Conference on Acoustics, Speech and Signal Processing (ICASSP)}}. \bibinfo{pages}{3080--3084}.
\newblock
\urldef\tempurl%
\url{https://doi.org/10.1109/ICASSP39728.2021.9413397}
\showDOI{\tempurl}


\bibitem[Guliani et~al\mbox{.}(2022)]%
        {9746226}
\bibfield{author}{\bibinfo{person}{Dhruv Guliani}, \bibinfo{person}{Lillian Zhou}, \bibinfo{person}{Changwan Ryu}, \bibinfo{person}{Tien-Ju Yang}, \bibinfo{person}{Harry Zhang}, \bibinfo{person}{Yonghui Xiao}, \bibinfo{person}{Françoise Beaufays}, {and} \bibinfo{person}{Giovanni Motta}.} \bibinfo{year}{2022}\natexlab{}.
\newblock \showarticletitle{Enabling On-Device Training of Speech Recognition Models With Federated Dropout}. In \bibinfo{booktitle}{\emph{ICASSP 2022 - 2022 IEEE International Conference on Acoustics, Speech and Signal Processing (ICASSP)}}. \bibinfo{pages}{8757--8761}.
\newblock
\urldef\tempurl%
\url{https://doi.org/10.1109/ICASSP43922.2022.9746226}
\showDOI{\tempurl}


\bibitem[Guo et~al\mbox{.}(2023a)]%
        {10005038}
\bibfield{author}{\bibinfo{person}{Jingtao Guo}, \bibinfo{person}{Ivan Wang-Hei Ho}, \bibinfo{person}{Yun Hou}, {and} \bibinfo{person}{Zijian Li}.} \bibinfo{year}{2023}\natexlab{a}.
\newblock \showarticletitle{FedPos: A Federated Transfer Learning Framework for CSI-Based Wi-Fi Indoor Positioning}.
\newblock \bibinfo{journal}{\emph{IEEE Systems Journal}} \bibinfo{volume}{17}, \bibinfo{number}{3} (\bibinfo{year}{2023}), \bibinfo{pages}{4579--4590}.
\newblock
\urldef\tempurl%
\url{https://doi.org/10.1109/JSYST.2022.3230425}
\showDOI{\tempurl}


\bibitem[Guo and Qin(2022)]%
        {10012859}
\bibfield{author}{\bibinfo{person}{Yiyu Guo} {and} \bibinfo{person}{Zhijin Qin}.} \bibinfo{year}{2022}\natexlab{}.
\newblock \showarticletitle{Federated Learning for Multi-view Synthesizing in Wireless Virtual Reality Networks}. In \bibinfo{booktitle}{\emph{2022 IEEE 96th Vehicular Technology Conference (VTC2022-Fall)}}. \bibinfo{pages}{1--5}.
\newblock
\urldef\tempurl%
\url{https://doi.org/10.1109/VTC2022-Fall57202.2022.10012859}
\showDOI{\tempurl}


\bibitem[Guo et~al\mbox{.}(2023b)]%
        {guo2023icmfed}
\bibfield{author}{\bibinfo{person}{Zihan Guo}, \bibinfo{person}{Linlin You}, \bibinfo{person}{Sheng Liu}, \bibinfo{person}{Junshu He}, {and} \bibinfo{person}{Bingran Zuo}.} \bibinfo{year}{2023}\natexlab{b}.
\newblock \showarticletitle{ICMFed: An Incremental and Cost-Efficient Mechanism of Federated Meta-Learning for Driver Distraction Detection}.
\newblock \bibinfo{journal}{\emph{Mathematics}} \bibinfo{volume}{11}, \bibinfo{number}{8} (\bibinfo{year}{2023}), \bibinfo{pages}{1867}.
\newblock


\bibitem[Gurukar et~al\mbox{.}(2022)]%
        {10.1145/3487351.3490964}
\bibfield{author}{\bibinfo{person}{Saket Gurukar}, \bibinfo{person}{Srinivasan Parthasarathy}, \bibinfo{person}{Rajiv Ramnath}, \bibinfo{person}{Catherine Calder}, {and} \bibinfo{person}{Sobhan Moosavi}.} \bibinfo{year}{2022}\natexlab{}.
\newblock \showarticletitle{LocationTrails: A Federated Approach to Learning Location Embeddings}. In \bibinfo{booktitle}{\emph{Proceedings of the 2021 IEEE/ACM International Conference on Advances in Social Networks Analysis and Mining}} (Virtual Event, Netherlands) \emph{(\bibinfo{series}{ASONAM '21})}. \bibinfo{publisher}{Association for Computing Machinery}, \bibinfo{address}{New York, NY, USA}, \bibinfo{pages}{377–384}.
\newblock
\showISBNx{9781450391283}
\urldef\tempurl%
\url{https://doi.org/10.1145/3487351.3490964}
\showDOI{\tempurl}


\bibitem[Hallaji et~al\mbox{.}(2024)]%
        {10420449}
\bibfield{author}{\bibinfo{person}{Ehsan Hallaji}, \bibinfo{person}{Roozbeh Razavi-Far}, \bibinfo{person}{Mehrdad Saif}, \bibinfo{person}{Boyu Wang}, {and} \bibinfo{person}{Qiang Yang}.} \bibinfo{year}{2024}\natexlab{}.
\newblock \showarticletitle{Decentralized Federated Learning: A Survey on Security and Privacy}.
\newblock \bibinfo{journal}{\emph{IEEE Transactions on Big Data}} \bibinfo{volume}{10}, \bibinfo{number}{2} (\bibinfo{year}{2024}), \bibinfo{pages}{194--213}.
\newblock
\urldef\tempurl%
\url{https://doi.org/10.1109/TBDATA.2024.3362191}
\showDOI{\tempurl}


\bibitem[Hamilton(1986)]%
        {hamilton1986hamilton}
\bibfield{author}{\bibinfo{person}{Max Hamilton}.} \bibinfo{year}{1986}\natexlab{}.
\newblock \showarticletitle{The Hamilton rating scale for depression}.
\newblock In \bibinfo{booktitle}{\emph{Assessment of depression}}. \bibinfo{publisher}{Springer}, \bibinfo{pages}{143--152}.
\newblock


\bibitem[Haouij et~al\mbox{.}(2018)]%
        {10.1145/3167132.3167395}
\bibfield{author}{\bibinfo{person}{Neska~El Haouij}, \bibinfo{person}{Jean-Michel Poggi}, \bibinfo{person}{Sylvie Sevestre-Ghalila}, \bibinfo{person}{Raja Ghozi}, {and} \bibinfo{person}{M\'{e}riem Ja\"{\i}dane}.} \bibinfo{year}{2018}\natexlab{}.
\newblock \showarticletitle{AffectiveROAD System and Database to Assess Driver's Attention}. In \bibinfo{booktitle}{\emph{Proceedings of the 33rd Annual ACM Symposium on Applied Computing}} (Pau, France) \emph{(\bibinfo{series}{SAC '18})}. \bibinfo{publisher}{Association for Computing Machinery}, \bibinfo{address}{New York, NY, USA}, \bibinfo{pages}{800–803}.
\newblock
\showISBNx{9781450351911}
\urldef\tempurl%
\url{https://doi.org/10.1145/3167132.3167395}
\showDOI{\tempurl}


\bibitem[Hard et~al\mbox{.}(2022)]%
        {hard2022production}
\bibfield{author}{\bibinfo{person}{Andrew Hard}, \bibinfo{person}{Kurt Partridge}, \bibinfo{person}{Neng Chen}, \bibinfo{person}{Sean Augenstein}, \bibinfo{person}{Aishanee Shah}, \bibinfo{person}{Hyun~Jin Park}, \bibinfo{person}{Alex Park}, \bibinfo{person}{Sara Ng}, \bibinfo{person}{Jessica Nguyen}, \bibinfo{person}{Ignacio~Lopez Moreno}, {et~al\mbox{.}}} \bibinfo{year}{2022}\natexlab{}.
\newblock \showarticletitle{Production federated keyword spotting via distillation, filtering, and joint federated-centralized training}.
\newblock \bibinfo{journal}{\emph{arXiv preprint arXiv:2204.06322}} (\bibinfo{year}{2022}).
\newblock


\bibitem[Hard et~al\mbox{.}(2020)]%
        {hard2020training}
\bibfield{author}{\bibinfo{person}{Andrew Hard}, \bibinfo{person}{Kurt Partridge}, \bibinfo{person}{Cameron Nguyen}, \bibinfo{person}{Niranjan Subrahmanya}, \bibinfo{person}{Aishanee Shah}, \bibinfo{person}{Pai Zhu}, \bibinfo{person}{Ignacio~Lopez Moreno}, {and} \bibinfo{person}{Rajiv Mathews}.} \bibinfo{year}{2020}\natexlab{}.
\newblock \showarticletitle{Training keyword spotting models on non-iid data with federated learning}.
\newblock \bibinfo{journal}{\emph{arXiv preprint arXiv:2005.10406}} (\bibinfo{year}{2020}).
\newblock


\bibitem[Hard et~al\mbox{.}(2018)]%
        {hard2018federated}
\bibfield{author}{\bibinfo{person}{Andrew Hard}, \bibinfo{person}{Kanishka Rao}, \bibinfo{person}{Rajiv Mathews}, \bibinfo{person}{Swaroop Ramaswamy}, \bibinfo{person}{Fran{\c{c}}oise Beaufays}, \bibinfo{person}{Sean Augenstein}, \bibinfo{person}{Hubert Eichner}, \bibinfo{person}{Chlo{\'e} Kiddon}, {and} \bibinfo{person}{Daniel Ramage}.} \bibinfo{year}{2018}\natexlab{}.
\newblock \showarticletitle{Federated learning for mobile keyboard prediction}.
\newblock \bibinfo{journal}{\emph{arXiv preprint arXiv:1811.03604}} (\bibinfo{year}{2018}).
\newblock


\bibitem[He et~al\mbox{.}(2023)]%
        {10.1145/3596599}
\bibfield{author}{\bibinfo{person}{Tongyue He}, \bibinfo{person}{Junxin Chen}, \bibinfo{person}{Ben-Guo He}, \bibinfo{person}{Wei Wang}, \bibinfo{person}{Zhi-Liang Zhu}, {and} \bibinfo{person}{Zhihan Lv}.} \bibinfo{year}{2023}\natexlab{}.
\newblock \showarticletitle{Toward Wearable Sensors: Advances, Trends, and Challenges}.
\newblock \bibinfo{journal}{\emph{ACM Comput. Surv.}} \bibinfo{volume}{55}, \bibinfo{number}{14s}, Article \bibinfo{articleno}{333} (\bibinfo{date}{jul} \bibinfo{year}{2023}), \bibinfo{numpages}{35}~pages.
\newblock
\showISSN{0360-0300}
\urldef\tempurl%
\url{https://doi.org/10.1145/3596599}
\showDOI{\tempurl}


\bibitem[He et~al\mbox{.}(2019)]%
        {10.1145/3359789.3359824}
\bibfield{author}{\bibinfo{person}{Zecheng He}, \bibinfo{person}{Tianwei Zhang}, {and} \bibinfo{person}{Ruby~B. Lee}.} \bibinfo{year}{2019}\natexlab{}.
\newblock \showarticletitle{Model inversion attacks against collaborative inference}. In \bibinfo{booktitle}{\emph{Proceedings of the 35th Annual Computer Security Applications Conference}} (San Juan, Puerto Rico, USA) \emph{(\bibinfo{series}{ACSAC '19})}. \bibinfo{publisher}{Association for Computing Machinery}, \bibinfo{address}{New York, NY, USA}, \bibinfo{pages}{148–162}.
\newblock
\showISBNx{9781450376280}
\urldef\tempurl%
\url{https://doi.org/10.1145/3359789.3359824}
\showDOI{\tempurl}


\bibitem[Heikenfeld et~al\mbox{.}(2018)]%
        {heikenfeld2018wearable}
\bibfield{author}{\bibinfo{person}{Jajack Heikenfeld}, \bibinfo{person}{Andrew Jajack}, \bibinfo{person}{Jim Rogers}, \bibinfo{person}{Philipp Gutruf}, \bibinfo{person}{Lei Tian}, \bibinfo{person}{Tingrui Pan}, \bibinfo{person}{Ruya Li}, \bibinfo{person}{Michelle Khine}, \bibinfo{person}{Jintae Kim}, {and} \bibinfo{person}{Juanhong Wang}.} \bibinfo{year}{2018}\natexlab{}.
\newblock \showarticletitle{Wearable sensors: modalities, challenges, and prospects}.
\newblock \bibinfo{journal}{\emph{Lab on a Chip}} \bibinfo{volume}{18}, \bibinfo{number}{2} (\bibinfo{year}{2018}), \bibinfo{pages}{217--248}.
\newblock


\bibitem[Host and Iva{\v{s}}i{\'c}-Kos(2022)]%
        {host2022overview}
\bibfield{author}{\bibinfo{person}{Kristina Host} {and} \bibinfo{person}{Marina Iva{\v{s}}i{\'c}-Kos}.} \bibinfo{year}{2022}\natexlab{}.
\newblock \showarticletitle{An overview of Human Action Recognition in sports based on Computer Vision}.
\newblock \bibinfo{journal}{\emph{Heliyon}} (\bibinfo{year}{2022}).
\newblock


\bibitem[Hu et~al\mbox{.}(2022a)]%
        {10.1145/3523273}
\bibfield{author}{\bibinfo{person}{Hongsheng Hu}, \bibinfo{person}{Zoran Salcic}, \bibinfo{person}{Lichao Sun}, \bibinfo{person}{Gillian Dobbie}, \bibinfo{person}{Philip~S. Yu}, {and} \bibinfo{person}{Xuyun Zhang}.} \bibinfo{year}{2022}\natexlab{a}.
\newblock \showarticletitle{Membership Inference Attacks on Machine Learning: A Survey}.
\newblock \bibinfo{journal}{\emph{ACM Comput. Surv.}} \bibinfo{volume}{54}, \bibinfo{number}{11s}, Article \bibinfo{articleno}{235} (\bibinfo{date}{sep} \bibinfo{year}{2022}), \bibinfo{numpages}{37}~pages.
\newblock
\showISSN{0360-0300}
\urldef\tempurl%
\url{https://doi.org/10.1145/3523273}
\showDOI{\tempurl}


\bibitem[Hu et~al\mbox{.}(2022b)]%
        {10.1145/3501814}
\bibfield{author}{\bibinfo{person}{Ziheng Hu}, \bibinfo{person}{Hongtao Xie}, \bibinfo{person}{Lingyun Yu}, \bibinfo{person}{Xingyu Gao}, \bibinfo{person}{Zhihua Shang}, {and} \bibinfo{person}{Yongdong Zhang}.} \bibinfo{year}{2022}\natexlab{b}.
\newblock \showarticletitle{Dynamic-Aware Federated Learning for Face Forgery Video Detection}.
\newblock \bibinfo{journal}{\emph{ACM Trans. Intell. Syst. Technol.}} \bibinfo{volume}{13}, \bibinfo{number}{4}, Article \bibinfo{articleno}{57} (\bibinfo{date}{jun} \bibinfo{year}{2022}), \bibinfo{numpages}{25}~pages.
\newblock
\showISSN{2157-6904}
\urldef\tempurl%
\url{https://doi.org/10.1145/3501814}
\showDOI{\tempurl}


\bibitem[Huang et~al\mbox{.}(2006)]%
        {HUANG2006489}
\bibfield{author}{\bibinfo{person}{Guang-Bin Huang}, \bibinfo{person}{Qin-Yu Zhu}, {and} \bibinfo{person}{Chee-Kheong Siew}.} \bibinfo{year}{2006}\natexlab{}.
\newblock \showarticletitle{Extreme learning machine: Theory and applications}.
\newblock \bibinfo{journal}{\emph{Neurocomputing}} \bibinfo{volume}{70}, \bibinfo{number}{1} (\bibinfo{year}{2006}), \bibinfo{pages}{489--501}.
\newblock
\showISSN{0925-2312}
\urldef\tempurl%
\url{https://doi.org/10.1016/j.neucom.2005.12.126}
\showDOI{\tempurl}
\newblock
\shownote{Neural Networks}.


\bibitem[Huang et~al\mbox{.}(2021)]%
        {huang2021personalized}
\bibfield{author}{\bibinfo{person}{Yutao Huang}, \bibinfo{person}{Lingyang Chu}, \bibinfo{person}{Zirui Zhou}, \bibinfo{person}{Lanjun Wang}, \bibinfo{person}{Jiangchuan Liu}, \bibinfo{person}{Jian Pei}, {and} \bibinfo{person}{Yong Zhang}.} \bibinfo{year}{2021}\natexlab{}.
\newblock \showarticletitle{Personalized cross-silo federated learning on non-iid data}. In \bibinfo{booktitle}{\emph{Proceedings of the AAAI conference on artificial intelligence}}, Vol.~\bibinfo{volume}{35}. \bibinfo{pages}{7865--7873}.
\newblock


\bibitem[Hwang et~al\mbox{.}(2023)]%
        {10301542}
\bibfield{author}{\bibinfo{person}{Tae-Ho Hwang}, \bibinfo{person}{Jingyao Shi}, {and} \bibinfo{person}{Kangyoon Lee}.} \bibinfo{year}{2023}\natexlab{}.
\newblock \showarticletitle{Enhancing Privacy-Preserving Personal Identification Through Federated Learning With Multimodal Vital Signs Data}.
\newblock \bibinfo{journal}{\emph{IEEE Access}}  \bibinfo{volume}{11} (\bibinfo{year}{2023}), \bibinfo{pages}{121556--121566}.
\newblock
\urldef\tempurl%
\url{https://doi.org/10.1109/ACCESS.2023.3328641}
\showDOI{\tempurl}


\bibitem[Iacob et~al\mbox{.}(2023)]%
        {iacob2023privacy}
\bibfield{author}{\bibinfo{person}{Alex Iacob}, \bibinfo{person}{Pedro~PB Gusm{\~a}o}, \bibinfo{person}{Nicholas~D Lane}, \bibinfo{person}{Armand~K Koupai}, \bibinfo{person}{Mohammud~J Bocus}, \bibinfo{person}{Ra{\'u}l Santos-Rodr{\'\i}guez}, \bibinfo{person}{Robert~J Piechocki}, {and} \bibinfo{person}{Ryan McConville}.} \bibinfo{year}{2023}\natexlab{}.
\newblock \showarticletitle{Privacy in Multimodal Federated Human Activity Recognition}.
\newblock \bibinfo{journal}{\emph{arXiv preprint arXiv:2305.12134}} (\bibinfo{year}{2023}).
\newblock


\bibitem[Imteaj et~al\mbox{.}(2021)]%
        {imteaj2021exploiting}
\bibfield{author}{\bibinfo{person}{Ahmed Imteaj}, \bibinfo{person}{Raghad Alabagi}, {and} \bibinfo{person}{M~Hadi Amini}.} \bibinfo{year}{2021}\natexlab{}.
\newblock \showarticletitle{Exploiting federated learning technique to recognize human activities in resource-constrained environment}. In \bibinfo{booktitle}{\emph{International Conference on Intelligent Human Computer Interaction}}. Springer, \bibinfo{pages}{659--672}.
\newblock


\bibitem[Imteaj et~al\mbox{.}(2022)]%
        {9475501}
\bibfield{author}{\bibinfo{person}{Ahmed Imteaj}, \bibinfo{person}{Urmish Thakker}, \bibinfo{person}{Shiqiang Wang}, \bibinfo{person}{Jian Li}, {and} \bibinfo{person}{M.~Hadi Amini}.} \bibinfo{year}{2022}\natexlab{}.
\newblock \showarticletitle{A Survey on Federated Learning for Resource-Constrained IoT Devices}.
\newblock \bibinfo{journal}{\emph{IEEE Internet of Things Journal}} \bibinfo{volume}{9}, \bibinfo{number}{1} (\bibinfo{year}{2022}), \bibinfo{pages}{1--24}.
\newblock
\urldef\tempurl%
\url{https://doi.org/10.1109/JIOT.2021.3095077}
\showDOI{\tempurl}


\bibitem[Ishimaru et~al\mbox{.}(2017)]%
        {10.1145/3123024.3129271}
\bibfield{author}{\bibinfo{person}{Shoya Ishimaru}, \bibinfo{person}{Kensuke Hoshika}, \bibinfo{person}{Kai Kunze}, \bibinfo{person}{Koichi Kise}, {and} \bibinfo{person}{Andreas Dengel}.} \bibinfo{year}{2017}\natexlab{}.
\newblock \showarticletitle{Towards Reading Trackers in the Wild: Detecting Reading Activities by EOG Glasses and Deep Neural Networks}. In \bibinfo{booktitle}{\emph{Proceedings of the 2017 ACM International Joint Conference on Pervasive and Ubiquitous Computing and Proceedings of the 2017 ACM International Symposium on Wearable Computers}} (Maui, Hawaii) \emph{(\bibinfo{series}{UbiComp '17})}. \bibinfo{publisher}{Association for Computing Machinery}, \bibinfo{address}{New York, NY, USA}, \bibinfo{pages}{704–711}.
\newblock
\showISBNx{9781450351904}
\urldef\tempurl%
\url{https://doi.org/10.1145/3123024.3129271}
\showDOI{\tempurl}


\bibitem[Javeed et~al\mbox{.}(2023)]%
        {10262054}
\bibfield{author}{\bibinfo{person}{Danish Javeed}, \bibinfo{person}{Muhammad~Shahid Saeed}, \bibinfo{person}{Prabhat Kumar}, \bibinfo{person}{Alireza Jolfaei}, \bibinfo{person}{Shareeful Islam}, {and} \bibinfo{person}{A.~K. M.~Najmul Islam}.} \bibinfo{year}{2023}\natexlab{}.
\newblock \showarticletitle{Federated Learning-based Personalized Recommendation Systems: An Overview on Security and Privacy Challenges}.
\newblock \bibinfo{journal}{\emph{IEEE Transactions on Consumer Electronics}} (\bibinfo{year}{2023}), \bibinfo{pages}{1--1}.
\newblock
\urldef\tempurl%
\url{https://doi.org/10.1109/TCE.2023.3318754}
\showDOI{\tempurl}


\bibitem[Jeong et~al\mbox{.}(2020)]%
        {jeong2020federated}
\bibfield{author}{\bibinfo{person}{Wonyong Jeong}, \bibinfo{person}{Jaehong Yoon}, \bibinfo{person}{Eunho Yang}, {and} \bibinfo{person}{Sung~Ju Hwang}.} \bibinfo{year}{2020}\natexlab{}.
\newblock \showarticletitle{Federated semi-supervised learning with inter-client consistency \& disjoint learning}.
\newblock \bibinfo{journal}{\emph{arXiv preprint arXiv:2006.12097}} (\bibinfo{year}{2020}).
\newblock


\bibitem[Jiang et~al\mbox{.}(2021a)]%
        {10.1145/3447687}
\bibfield{author}{\bibinfo{person}{Di Jiang}, \bibinfo{person}{Conghui Tan}, \bibinfo{person}{Jinhua Peng}, \bibinfo{person}{Chaotao Chen}, \bibinfo{person}{Xueyang Wu}, \bibinfo{person}{Weiwei Zhao}, \bibinfo{person}{Yuanfeng Song}, \bibinfo{person}{Yongxin Tong}, \bibinfo{person}{Chang Liu}, \bibinfo{person}{Qian Xu}, \bibinfo{person}{Qiang Yang}, {and} \bibinfo{person}{Li Deng}.} \bibinfo{year}{2021}\natexlab{a}.
\newblock \showarticletitle{A GDPR-Compliant Ecosystem for Speech Recognition with Transfer, Federated, and Evolutionary Learning}.
\newblock \bibinfo{journal}{\emph{ACM Trans. Intell. Syst. Technol.}} \bibinfo{volume}{12}, \bibinfo{number}{3}, Article \bibinfo{articleno}{30} (\bibinfo{date}{may} \bibinfo{year}{2021}), \bibinfo{numpages}{19}~pages.
\newblock
\showISSN{2157-6904}
\urldef\tempurl%
\url{https://doi.org/10.1145/3447687}
\showDOI{\tempurl}


\bibitem[Jiang et~al\mbox{.}(2023a)]%
        {10.1145/3550486}
\bibfield{author}{\bibinfo{person}{Linli Jiang}, \bibinfo{person}{Chao-Xiong Chen}, {and} \bibinfo{person}{Chao Chen}.} \bibinfo{year}{2023}\natexlab{a}.
\newblock \showarticletitle{L2MM: Learning to Map Matching with Deep Models for Low-Quality GPS Trajectory Data}.
\newblock \bibinfo{journal}{\emph{ACM Trans. Knowl. Discov. Data}} \bibinfo{volume}{17}, \bibinfo{number}{3}, Article \bibinfo{articleno}{39} (\bibinfo{date}{feb} \bibinfo{year}{2023}), \bibinfo{numpages}{25}~pages.
\newblock
\showISSN{1556-4681}
\urldef\tempurl%
\url{https://doi.org/10.1145/3550486}
\showDOI{\tempurl}


\bibitem[Jiang et~al\mbox{.}(2023b)]%
        {10197159}
\bibfield{author}{\bibinfo{person}{Shiyi Jiang}, \bibinfo{person}{Farshad Firouzi}, {and} \bibinfo{person}{Krishnendu Chakrabarty}.} \bibinfo{year}{2023}\natexlab{b}.
\newblock \showarticletitle{Low-Overhead Clustered Federated Learning for Personalized Stress Monitoring}.
\newblock \bibinfo{journal}{\emph{IEEE Internet of Things Journal}} (\bibinfo{year}{2023}), \bibinfo{pages}{1--1}.
\newblock
\urldef\tempurl%
\url{https://doi.org/10.1109/JIOT.2023.3299736}
\showDOI{\tempurl}


\bibitem[Jiang et~al\mbox{.}(2021b)]%
        {9671447}
\bibfield{author}{\bibinfo{person}{Xiaopeng Jiang}, \bibinfo{person}{Shuai Zhao}, \bibinfo{person}{Guy Jacobson}, \bibinfo{person}{Rittwik Jana}, \bibinfo{person}{Wen-Ling Hsu}, \bibinfo{person}{Manoop Talasila}, \bibinfo{person}{Syed~Anwar Aftab}, \bibinfo{person}{Yi Chen}, {and} \bibinfo{person}{Cristian Borcea}.} \bibinfo{year}{2021}\natexlab{b}.
\newblock \showarticletitle{Federated Meta-Location Learning for Fine-Grained Location Prediction}. In \bibinfo{booktitle}{\emph{2021 IEEE International Conference on Big Data (Big Data)}}. \bibinfo{pages}{446--456}.
\newblock
\urldef\tempurl%
\url{https://doi.org/10.1109/BigData52589.2021.9671447}
\showDOI{\tempurl}


\bibitem[Jin et~al\mbox{.}(2023)]%
        {jin2023federated}
\bibfield{author}{\bibinfo{person}{Yilun Jin}, \bibinfo{person}{Yang Liu}, \bibinfo{person}{Kai Chen}, {and} \bibinfo{person}{Qiang Yang}.} \bibinfo{year}{2023}\natexlab{}.
\newblock \showarticletitle{Federated Learning without Full Labels: A Survey}.
\newblock \bibinfo{journal}{\emph{arXiv preprint arXiv:2303.14453}} (\bibinfo{year}{2023}).
\newblock


\bibitem[Kairouz et~al\mbox{.}(2021)]%
        {kairouz2021advances}
\bibfield{author}{\bibinfo{person}{Peter Kairouz}, \bibinfo{person}{H~Brendan McMahan}, \bibinfo{person}{Brendan Avent}, \bibinfo{person}{Aur{\'e}lien Bellet}, \bibinfo{person}{Mehdi Bennis}, \bibinfo{person}{Arjun~Nitin Bhagoji}, \bibinfo{person}{Kallista Bonawitz}, \bibinfo{person}{Zachary Charles}, \bibinfo{person}{Graham Cormode}, \bibinfo{person}{Rachel Cummings}, {et~al\mbox{.}}} \bibinfo{year}{2021}\natexlab{}.
\newblock \showarticletitle{Advances and open problems in federated learning}.
\newblock \bibinfo{journal}{\emph{Foundations and Trends{\textregistered} in Machine Learning}} \bibinfo{volume}{14}, \bibinfo{number}{1--2} (\bibinfo{year}{2021}), \bibinfo{pages}{1--210}.
\newblock


\bibitem[Kalabakov et~al\mbox{.}(2023)]%
        {10162197}
\bibfield{author}{\bibinfo{person}{Stefan Kalabakov}, \bibinfo{person}{Borche Jovanovski}, \bibinfo{person}{Daniel Denkovski}, \bibinfo{person}{Valentin Rakovic}, \bibinfo{person}{Bjarne Pfitzner}, \bibinfo{person}{Orhan Konak}, \bibinfo{person}{Bert Arnrich}, {and} \bibinfo{person}{Hristijan Gjoreski}.} \bibinfo{year}{2023}\natexlab{}.
\newblock \showarticletitle{Federated Learning for Activity Recognition: A System Level Perspective}.
\newblock \bibinfo{journal}{\emph{IEEE Access}}  \bibinfo{volume}{11} (\bibinfo{year}{2023}), \bibinfo{pages}{64442--64457}.
\newblock
\urldef\tempurl%
\url{https://doi.org/10.1109/ACCESS.2023.3289220}
\showDOI{\tempurl}


\bibitem[Katsigiannis and Ramzan(2018)]%
        {7887697}
\bibfield{author}{\bibinfo{person}{Stamos Katsigiannis} {and} \bibinfo{person}{Naeem Ramzan}.} \bibinfo{year}{2018}\natexlab{}.
\newblock \showarticletitle{DREAMER: A Database for Emotion Recognition Through EEG and ECG Signals From Wireless Low-cost Off-the-Shelf Devices}.
\newblock \bibinfo{journal}{\emph{IEEE Journal of Biomedical and Health Informatics}} \bibinfo{volume}{22}, \bibinfo{number}{1} (\bibinfo{year}{2018}), \bibinfo{pages}{98--107}.
\newblock
\urldef\tempurl%
\url{https://doi.org/10.1109/JBHI.2017.2688239}
\showDOI{\tempurl}


\bibitem[Khan et~al\mbox{.}({[n.\,d.]})]%
        {khan4395564federated}
\bibfield{author}{\bibinfo{person}{Ahsan~Raza Khan}, \bibinfo{person}{Syed~Mohsin Bokhari}, \bibinfo{person}{Sarmad Sohaib}, \bibinfo{person}{Olaoluwa Popoola}, \bibinfo{person}{Kamran Arshad}, \bibinfo{person}{Khaled Assaleh}, \bibinfo{person}{Muhammad~Ali Imran}, {and} \bibinfo{person}{Ahmed Zoha}.} \bibinfo{year}{[n.\,d.]}\natexlab{}.
\newblock \showarticletitle{Federated Learning Based Non-Invasive Human Activity Recognition Using Channel State Information}.
\newblock \bibinfo{journal}{\emph{Available at SSRN 4395564}} (\bibinfo{year}{[n.\,d.]}).
\newblock


\bibitem[Khan et~al\mbox{.}(2021)]%
        {9460016}
\bibfield{author}{\bibinfo{person}{Latif~U. Khan}, \bibinfo{person}{Walid Saad}, \bibinfo{person}{Zhu Han}, \bibinfo{person}{Ekram Hossain}, {and} \bibinfo{person}{Choong~Seon Hong}.} \bibinfo{year}{2021}\natexlab{}.
\newblock \showarticletitle{Federated Learning for Internet of Things: Recent Advances, Taxonomy, and Open Challenges}.
\newblock \bibinfo{journal}{\emph{IEEE Communications Surveys \& Tutorials}} \bibinfo{volume}{23}, \bibinfo{number}{3} (\bibinfo{year}{2021}), \bibinfo{pages}{1759--1799}.
\newblock
\urldef\tempurl%
\url{https://doi.org/10.1109/COMST.2021.3090430}
\showDOI{\tempurl}


\bibitem[Khoa et~al\mbox{.}(2022)]%
        {10.1145/3512731.3534207}
\bibfield{author}{\bibinfo{person}{Tran~Anh Khoa}, \bibinfo{person}{Do-Van Nguyen}, \bibinfo{person}{Phuoc~Van Nguyen~Thi}, {and} \bibinfo{person}{Koji Zettsu}.} \bibinfo{year}{2022}\natexlab{}.
\newblock \showarticletitle{FedMCRNN: Federated Learning Using Multiple Convolutional Recurrent Neural Networks for Sleep Quality Prediction}. In \bibinfo{booktitle}{\emph{Proceedings of the 3rd ACM Workshop on Intelligent Cross-Data Analysis and Retrieval}} (Newark, NJ, USA) \emph{(\bibinfo{series}{ICDAR '22})}. \bibinfo{publisher}{Association for Computing Machinery}, \bibinfo{address}{New York, NY, USA}, \bibinfo{pages}{63–69}.
\newblock
\showISBNx{9781450392419}
\urldef\tempurl%
\url{https://doi.org/10.1145/3512731.3534207}
\showDOI{\tempurl}


\bibitem[Khoa et~al\mbox{.}(2023)]%
        {10146344}
\bibfield{author}{\bibinfo{person}{Tran~Anh Khoa}, \bibinfo{person}{Nguyen~Dang Trac}, \bibinfo{person}{Vo~Phuc Tinh}, \bibinfo{person}{Nguyen~Hoang Nam}, \bibinfo{person}{Duc Ngoc~Minh Dang}, \bibinfo{person}{Hoang~Hai Son}, {and} \bibinfo{person}{Pham~Duc Lam}.} \bibinfo{year}{2023}\natexlab{}.
\newblock \showarticletitle{Safety Is Our Friend: A Federated Learning Framework Toward Driver’s State and Behavior Detection}.
\newblock \bibinfo{journal}{\emph{IEEE Transactions on Computational Social Systems}} (\bibinfo{year}{2023}), \bibinfo{pages}{1--19}.
\newblock
\urldef\tempurl%
\url{https://doi.org/10.1109/TCSS.2023.3273727}
\showDOI{\tempurl}


\bibitem[Kitchenham et~al\mbox{.}(2009)]%
        {KITCHENHAM20097}
\bibfield{author}{\bibinfo{person}{Barbara Kitchenham}, \bibinfo{person}{O. {Pearl Brereton}}, \bibinfo{person}{David Budgen}, \bibinfo{person}{Mark Turner}, \bibinfo{person}{John Bailey}, {and} \bibinfo{person}{Stephen Linkman}.} \bibinfo{year}{2009}\natexlab{}.
\newblock \showarticletitle{Systematic literature reviews in software engineering – A systematic literature review}.
\newblock \bibinfo{journal}{\emph{Information and Software Technology}} \bibinfo{volume}{51}, \bibinfo{number}{1} (\bibinfo{year}{2009}), \bibinfo{pages}{7--15}.
\newblock
\showISSN{0950-5849}
\urldef\tempurl%
\url{https://doi.org/10.1016/j.infsof.2008.09.009}
\showDOI{\tempurl}


\bibitem[Kong et~al\mbox{.}(2023)]%
        {10266762}
\bibfield{author}{\bibinfo{person}{Xiangjie Kong}, \bibinfo{person}{Wenyi Zhang}, \bibinfo{person}{Youyang Qu}, \bibinfo{person}{Xinwei Yao}, {and} \bibinfo{person}{Guojiang Shen}.} \bibinfo{year}{2023}\natexlab{}.
\newblock \showarticletitle{FedAWR : An Interactive Federated Active Learning Framework for Air Writing Recognition}.
\newblock \bibinfo{journal}{\emph{IEEE Transactions on Mobile Computing}} (\bibinfo{year}{2023}), \bibinfo{pages}{1--15}.
\newblock
\urldef\tempurl%
\url{https://doi.org/10.1109/TMC.2023.3320147}
\showDOI{\tempurl}


\bibitem[Kosch et~al\mbox{.}(2023)]%
        {10.1145/3582272}
\bibfield{author}{\bibinfo{person}{Thomas Kosch}, \bibinfo{person}{Jakob Karolus}, \bibinfo{person}{Johannes Zagermann}, \bibinfo{person}{Harald Reiterer}, \bibinfo{person}{Albrecht Schmidt}, {and} \bibinfo{person}{Pawe\l{}~W. Wo\'{z}niak}.} \bibinfo{year}{2023}\natexlab{}.
\newblock \showarticletitle{A Survey on Measuring Cognitive Workload in Human-Computer Interaction}.
\newblock \bibinfo{journal}{\emph{ACM Comput. Surv.}} \bibinfo{volume}{55}, \bibinfo{number}{13s}, Article \bibinfo{articleno}{283} (\bibinfo{date}{jul} \bibinfo{year}{2023}), \bibinfo{numpages}{39}~pages.
\newblock
\showISSN{0360-0300}
\urldef\tempurl%
\url{https://doi.org/10.1145/3582272}
\showDOI{\tempurl}


\bibitem[Kurakin et~al\mbox{.}(2016)]%
        {kurakin2016adversarial}
\bibfield{author}{\bibinfo{person}{Alexey Kurakin}, \bibinfo{person}{Ian Goodfellow}, {and} \bibinfo{person}{Samy Bengio}.} \bibinfo{year}{2016}\natexlab{}.
\newblock \showarticletitle{Adversarial machine learning at scale}.
\newblock \bibinfo{journal}{\emph{arXiv preprint arXiv:1611.01236}} (\bibinfo{year}{2016}).
\newblock


\bibitem[Lang(1995)]%
        {lang1995emotion}
\bibfield{author}{\bibinfo{person}{Peter~J Lang}.} \bibinfo{year}{1995}\natexlab{}.
\newblock \showarticletitle{The emotion probe: Studies of motivation and attention.}
\newblock \bibinfo{journal}{\emph{American psychologist}} \bibinfo{volume}{50}, \bibinfo{number}{5} (\bibinfo{year}{1995}), \bibinfo{pages}{372}.
\newblock


\bibitem[Langheinrich(2001)]%
        {langheinrich2001privacy}
\bibfield{author}{\bibinfo{person}{Marc Langheinrich}.} \bibinfo{year}{2001}\natexlab{}.
\newblock \showarticletitle{Privacy by design—principles of privacy-aware ubiquitous systems}. In \bibinfo{booktitle}{\emph{International conference on ubiquitous computing}}. Springer, \bibinfo{pages}{273--291}.
\newblock


\bibitem[Lecun et~al\mbox{.}(1998)]%
        {726791}
\bibfield{author}{\bibinfo{person}{Y. Lecun}, \bibinfo{person}{L. Bottou}, \bibinfo{person}{Y. Bengio}, {and} \bibinfo{person}{P. Haffner}.} \bibinfo{year}{1998}\natexlab{}.
\newblock \showarticletitle{Gradient-based learning applied to document recognition}.
\newblock \bibinfo{journal}{\emph{Proc. IEEE}} \bibinfo{volume}{86}, \bibinfo{number}{11} (\bibinfo{year}{1998}), \bibinfo{pages}{2278--2324}.
\newblock
\urldef\tempurl%
\url{https://doi.org/10.1109/5.726791}
\showDOI{\tempurl}


\bibitem[Leroy et~al\mbox{.}(2019)]%
        {8683546}
\bibfield{author}{\bibinfo{person}{David Leroy}, \bibinfo{person}{Alice Coucke}, \bibinfo{person}{Thibaut Lavril}, \bibinfo{person}{Thibault Gisselbrecht}, {and} \bibinfo{person}{Joseph Dureau}.} \bibinfo{year}{2019}\natexlab{}.
\newblock \showarticletitle{Federated Learning for Keyword Spotting}. In \bibinfo{booktitle}{\emph{ICASSP 2019 - 2019 IEEE International Conference on Acoustics, Speech and Signal Processing (ICASSP)}}. \bibinfo{pages}{6341--6345}.
\newblock
\urldef\tempurl%
\url{https://doi.org/10.1109/ICASSP.2019.8683546}
\showDOI{\tempurl}


\bibitem[Li et~al\mbox{.}(2020b)]%
        {10.1145/3397536.3422270}
\bibfield{author}{\bibinfo{person}{Anliang Li}, \bibinfo{person}{Shuang Wang}, \bibinfo{person}{Wenzhu Li}, \bibinfo{person}{Shengnan Liu}, {and} \bibinfo{person}{Siyuan Zhang}.} \bibinfo{year}{2020}\natexlab{b}.
\newblock \showarticletitle{Predicting Human Mobility with Federated Learning}. In \bibinfo{booktitle}{\emph{Proceedings of the 28th International Conference on Advances in Geographic Information Systems}} (Seattle, WA, USA) \emph{(\bibinfo{series}{SIGSPATIAL '20})}. \bibinfo{publisher}{Association for Computing Machinery}, \bibinfo{address}{New York, NY, USA}, \bibinfo{pages}{441–444}.
\newblock
\showISBNx{9781450380195}
\urldef\tempurl%
\url{https://doi.org/10.1145/3397536.3422270}
\showDOI{\tempurl}


\bibitem[Li et~al\mbox{.}(2021b)]%
        {10.1145/3442381.3450006}
\bibfield{author}{\bibinfo{person}{Chenglin Li}, \bibinfo{person}{Di Niu}, \bibinfo{person}{Bei Jiang}, \bibinfo{person}{Xiao Zuo}, {and} \bibinfo{person}{Jianming Yang}.} \bibinfo{year}{2021}\natexlab{b}.
\newblock \showarticletitle{Meta-HAR: Federated Representation Learning for Human Activity Recognition}. In \bibinfo{booktitle}{\emph{Proceedings of the Web Conference 2021}} (Ljubljana, Slovenia) \emph{(\bibinfo{series}{WWW '21})}. \bibinfo{publisher}{Association for Computing Machinery}, \bibinfo{address}{New York, NY, USA}, \bibinfo{pages}{912–922}.
\newblock
\showISBNx{9781450383127}
\urldef\tempurl%
\url{https://doi.org/10.1145/3442381.3450006}
\showDOI{\tempurl}


\bibitem[Li et~al\mbox{.}(2021c)]%
        {9724416}
\bibfield{author}{\bibinfo{person}{Jinli Li}, \bibinfo{person}{Ran Zhang}, \bibinfo{person}{Mingcan Cen}, \bibinfo{person}{Xunao Wang}, {and} \bibinfo{person}{M. Jiang}.} \bibinfo{year}{2021}\natexlab{c}.
\newblock \showarticletitle{Depression Detection Using Asynchronous Federated Optimization}. In \bibinfo{booktitle}{\emph{2021 IEEE 20th International Conference on Trust, Security and Privacy in Computing and Communications (TrustCom)}}. \bibinfo{pages}{758--765}.
\newblock
\urldef\tempurl%
\url{https://doi.org/10.1109/TrustCom53373.2021.00110}
\showDOI{\tempurl}


\bibitem[Li et~al\mbox{.}(2021a)]%
        {9482149}
\bibfield{author}{\bibinfo{person}{Lin Li}, \bibinfo{person}{Mai Li}, \bibinfo{person}{Fang Qin}, {and} \bibinfo{person}{Weijia Zeng}.} \bibinfo{year}{2021}\natexlab{a}.
\newblock \showarticletitle{Evolutionary-based Federated Ensemble Learning on Face Recognition}. In \bibinfo{booktitle}{\emph{2021 IEEE 4th Advanced Information Management, Communicates, Electronic and Automation Control Conference (IMCEC)}}, Vol.~\bibinfo{volume}{4}. \bibinfo{pages}{815--819}.
\newblock
\urldef\tempurl%
\url{https://doi.org/10.1109/IMCEC51613.2021.9482149}
\showDOI{\tempurl}


\bibitem[Li et~al\mbox{.}(2022a)]%
        {9835537}
\bibfield{author}{\bibinfo{person}{Qinbin Li}, \bibinfo{person}{Yiqun Diao}, \bibinfo{person}{Quan Chen}, {and} \bibinfo{person}{Bingsheng He}.} \bibinfo{year}{2022}\natexlab{a}.
\newblock \showarticletitle{Federated Learning on Non-IID Data Silos: An Experimental Study}. In \bibinfo{booktitle}{\emph{2022 IEEE 38th International Conference on Data Engineering (ICDE)}}. \bibinfo{pages}{965--978}.
\newblock
\urldef\tempurl%
\url{https://doi.org/10.1109/ICDE53745.2022.00077}
\showDOI{\tempurl}


\bibitem[Li et~al\mbox{.}(2020a)]%
        {9084352}
\bibfield{author}{\bibinfo{person}{Tian Li}, \bibinfo{person}{Anit~Kumar Sahu}, \bibinfo{person}{Ameet Talwalkar}, {and} \bibinfo{person}{Virginia Smith}.} \bibinfo{year}{2020}\natexlab{a}.
\newblock \showarticletitle{Federated Learning: Challenges, Methods, and Future Directions}.
\newblock \bibinfo{journal}{\emph{IEEE Signal Processing Magazine}} \bibinfo{volume}{37}, \bibinfo{number}{3} (\bibinfo{year}{2020}), \bibinfo{pages}{50--60}.
\newblock
\urldef\tempurl%
\url{https://doi.org/10.1109/MSP.2020.2975749}
\showDOI{\tempurl}


\bibitem[Li et~al\mbox{.}(2019)]%
        {li2019privacy}
\bibfield{author}{\bibinfo{person}{Wenqi Li}, \bibinfo{person}{Fausto Milletar{\`\i}}, \bibinfo{person}{Daguang Xu}, \bibinfo{person}{Nicola Rieke}, \bibinfo{person}{Jonny Hancox}, \bibinfo{person}{Wentao Zhu}, \bibinfo{person}{Maximilian Baust}, \bibinfo{person}{Yan Cheng}, \bibinfo{person}{S{\'e}bastien Ourselin}, \bibinfo{person}{M~Jorge Cardoso}, {et~al\mbox{.}}} \bibinfo{year}{2019}\natexlab{}.
\newblock \showarticletitle{Privacy-preserving federated brain tumour segmentation}. In \bibinfo{booktitle}{\emph{Machine Learning in Medical Imaging: 10th International Workshop, MLMI 2019, Held in Conjunction with MICCAI 2019, Shenzhen, China, October 13, 2019, Proceedings 10}}. Springer, \bibinfo{pages}{133--141}.
\newblock


\bibitem[Li et~al\mbox{.}(2020c)]%
        {9103044}
\bibfield{author}{\bibinfo{person}{Wei Li}, \bibinfo{person}{Cheng Zhang}, {and} \bibinfo{person}{Yoshiaki Tanaka}.} \bibinfo{year}{2020}\natexlab{c}.
\newblock \showarticletitle{Pseudo Label-Driven Federated Learning-Based Decentralized Indoor Localization via Mobile Crowdsourcing}.
\newblock \bibinfo{journal}{\emph{IEEE Sensors Journal}} \bibinfo{volume}{20}, \bibinfo{number}{19} (\bibinfo{year}{2020}), \bibinfo{pages}{11556--11565}.
\newblock
\urldef\tempurl%
\url{https://doi.org/10.1109/JSEN.2020.2998116}
\showDOI{\tempurl}


\bibitem[Li et~al\mbox{.}(2022c)]%
        {li2022avoid}
\bibfield{author}{\bibinfo{person}{Xin-Chun Li}, \bibinfo{person}{Jin-Lin Tang}, \bibinfo{person}{Shaoming Song}, \bibinfo{person}{Bingshuai Li}, \bibinfo{person}{Yinchuan Li}, \bibinfo{person}{Yunfeng Shao}, \bibinfo{person}{Le Gan}, {and} \bibinfo{person}{De-Chuan Zhan}.} \bibinfo{year}{2022}\natexlab{c}.
\newblock \showarticletitle{Avoid overfitting user specific information in federated keyword spotting}.
\newblock \bibinfo{journal}{\emph{arXiv preprint arXiv:2206.08864}} (\bibinfo{year}{2022}).
\newblock


\bibitem[Li et~al\mbox{.}(2023)]%
        {10.1145/3580795}
\bibfield{author}{\bibinfo{person}{Youpeng Li}, \bibinfo{person}{Xuyu Wang}, {and} \bibinfo{person}{Lingling An}.} \bibinfo{year}{2023}\natexlab{}.
\newblock \showarticletitle{Hierarchical Clustering-Based Personalized Federated Learning for Robust and Fair Human Activity Recognition}.
\newblock \bibinfo{journal}{\emph{Proc. ACM Interact. Mob. Wearable Ubiquitous Technol.}} \bibinfo{volume}{7}, \bibinfo{number}{1}, Article \bibinfo{articleno}{20} (\bibinfo{date}{mar} \bibinfo{year}{2023}), \bibinfo{numpages}{38}~pages.
\newblock
\urldef\tempurl%
\url{https://doi.org/10.1145/3580795}
\showDOI{\tempurl}


\bibitem[Li et~al\mbox{.}(2022b)]%
        {9956474}
\bibfield{author}{\bibinfo{person}{Ziqiong Li}, \bibinfo{person}{Yan-Ran Li}, {and} \bibinfo{person}{Shiqi Yu}.} \bibinfo{year}{2022}\natexlab{b}.
\newblock \showarticletitle{FedGait: A Benchmark for Federated Gait Recognition}. In \bibinfo{booktitle}{\emph{2022 26th International Conference on Pattern Recognition (ICPR)}}. \bibinfo{pages}{1371--1377}.
\newblock
\urldef\tempurl%
\url{https://doi.org/10.1109/ICPR56361.2022.9956474}
\showDOI{\tempurl}


\bibitem[Lieskovsk{\'a} et~al\mbox{.}(2021)]%
        {lieskovska2021review}
\bibfield{author}{\bibinfo{person}{Eva Lieskovsk{\'a}}, \bibinfo{person}{Maro{\v{s}} Jakubec}, \bibinfo{person}{Roman Jarina}, {and} \bibinfo{person}{Michal Chmul{\'\i}k}.} \bibinfo{year}{2021}\natexlab{}.
\newblock \showarticletitle{A review on speech emotion recognition using deep learning and attention mechanism}.
\newblock \bibinfo{journal}{\emph{Electronics}} \bibinfo{volume}{10}, \bibinfo{number}{10} (\bibinfo{year}{2021}), \bibinfo{pages}{1163}.
\newblock


\bibitem[Lin et~al\mbox{.}(2023)]%
        {LIN2023110396}
\bibfield{author}{\bibinfo{person}{Zhentao Lin}, \bibinfo{person}{Bi Zeng}, \bibinfo{person}{Huiting Hu}, \bibinfo{person}{Yuting Huang}, \bibinfo{person}{Linwen Xu}, {and} \bibinfo{person}{Zhuangze Yao}.} \bibinfo{year}{2023}\natexlab{}.
\newblock \showarticletitle{SASE: Self-Adaptive noise distribution network for Speech Enhancement with Federated Learning using heterogeneous data}.
\newblock \bibinfo{journal}{\emph{Knowledge-Based Systems}}  \bibinfo{volume}{266} (\bibinfo{year}{2023}), \bibinfo{pages}{110396}.
\newblock
\showISSN{0950-7051}
\urldef\tempurl%
\url{https://doi.org/10.1016/j.knosys.2023.110396}
\showDOI{\tempurl}


\bibitem[Liu et~al\mbox{.}(2022b)]%
        {liu2022fedfr}
\bibfield{author}{\bibinfo{person}{Chih-Ting Liu}, \bibinfo{person}{Chien-Yi Wang}, \bibinfo{person}{Shao-Yi Chien}, {and} \bibinfo{person}{Shang-Hong Lai}.} \bibinfo{year}{2022}\natexlab{b}.
\newblock \showarticletitle{FedFR: Joint optimization federated framework for generic and personalized face recognition}. In \bibinfo{booktitle}{\emph{Proceedings of the AAAI Conference on Artificial Intelligence}}, Vol.~\bibinfo{volume}{36}. \bibinfo{pages}{1656--1664}.
\newblock


\bibitem[Liu et~al\mbox{.}(2023a)]%
        {10177776}
\bibfield{author}{\bibinfo{person}{Decheng Liu}, \bibinfo{person}{Zhan Dang}, \bibinfo{person}{Chunlei Peng}, \bibinfo{person}{Yu Zheng}, \bibinfo{person}{Shuang Li}, \bibinfo{person}{Nannan Wang}, {and} \bibinfo{person}{Xinbo Gao}.} \bibinfo{year}{2023}\natexlab{a}.
\newblock \showarticletitle{FedForgery: Generalized Face Forgery Detection With Residual Federated Learning}.
\newblock \bibinfo{journal}{\emph{IEEE Transactions on Information Forensics and Security}}  \bibinfo{volume}{18} (\bibinfo{year}{2023}), \bibinfo{pages}{4272--4284}.
\newblock
\urldef\tempurl%
\url{https://doi.org/10.1109/TIFS.2023.3293951}
\showDOI{\tempurl}


\bibitem[Liu et~al\mbox{.}(2020)]%
        {8794643}
\bibfield{author}{\bibinfo{person}{Jian Liu}, \bibinfo{person}{Hongbo Liu}, \bibinfo{person}{Yingying Chen}, \bibinfo{person}{Yan Wang}, {and} \bibinfo{person}{Chen Wang}.} \bibinfo{year}{2020}\natexlab{}.
\newblock \showarticletitle{Wireless Sensing for Human Activity: A Survey}.
\newblock \bibinfo{journal}{\emph{IEEE Communications Surveys \& Tutorials}} \bibinfo{volume}{22}, \bibinfo{number}{3} (\bibinfo{year}{2020}), \bibinfo{pages}{1629--1645}.
\newblock
\urldef\tempurl%
\url{https://doi.org/10.1109/COMST.2019.2934489}
\showDOI{\tempurl}


\bibitem[Liu et~al\mbox{.}(2022c)]%
        {9949249}
\bibfield{author}{\bibinfo{person}{Jiabei Liu}, \bibinfo{person}{Weiming Zhuang}, \bibinfo{person}{Yonggang Wen}, \bibinfo{person}{Jun Huang}, {and} \bibinfo{person}{Wei Lin}.} \bibinfo{year}{2022}\natexlab{c}.
\newblock \showarticletitle{Optimizing Federated Unsupervised Person Re-identification via Camera-aware Clustering}. In \bibinfo{booktitle}{\emph{2022 IEEE 24th International Workshop on Multimedia Signal Processing (MMSP)}}. \bibinfo{pages}{1--6}.
\newblock
\urldef\tempurl%
\url{https://doi.org/10.1109/MMSP55362.2022.9949249}
\showDOI{\tempurl}


\bibitem[Liu et~al\mbox{.}(2022a)]%
        {liu2022federated}
\bibfield{author}{\bibinfo{person}{Songfeng Liu}, \bibinfo{person}{Jinyan Wang}, {and} \bibinfo{person}{Wenliang Zhang}.} \bibinfo{year}{2022}\natexlab{a}.
\newblock \showarticletitle{Federated personalized random forest for human activity recognition}.
\newblock \bibinfo{journal}{\emph{Math. Biosci. Eng}} \bibinfo{volume}{19}, \bibinfo{number}{1} (\bibinfo{year}{2022}), \bibinfo{pages}{953--971}.
\newblock


\bibitem[Liu et~al\mbox{.}(2023b)]%
        {9462394}
\bibfield{author}{\bibinfo{person}{Xiao Liu}, \bibinfo{person}{Fanjin Zhang}, \bibinfo{person}{Zhenyu Hou}, \bibinfo{person}{Li Mian}, \bibinfo{person}{Zhaoyu Wang}, \bibinfo{person}{Jing Zhang}, {and} \bibinfo{person}{Jie Tang}.} \bibinfo{year}{2023}\natexlab{b}.
\newblock \showarticletitle{Self-Supervised Learning: Generative or Contrastive}.
\newblock \bibinfo{journal}{\emph{IEEE Transactions on Knowledge and Data Engineering}} \bibinfo{volume}{35}, \bibinfo{number}{1} (\bibinfo{year}{2023}), \bibinfo{pages}{857--876}.
\newblock
\urldef\tempurl%
\url{https://doi.org/10.1109/TKDE.2021.3090866}
\showDOI{\tempurl}


\bibitem[Liu et~al\mbox{.}(2019)]%
        {9066152}
\bibfield{author}{\bibinfo{person}{Yuxiang Liu}, \bibinfo{person}{Huichuwu Li}, \bibinfo{person}{Jiang Xiao}, {and} \bibinfo{person}{Hai Jin}.} \bibinfo{year}{2019}\natexlab{}.
\newblock \showarticletitle{FLoc: Fingerprint-Based Indoor Localization System under a Federated Learning Updating Framework}. In \bibinfo{booktitle}{\emph{2019 15th International Conference on Mobile Ad-Hoc and Sensor Networks (MSN)}}. \bibinfo{pages}{113--118}.
\newblock
\urldef\tempurl%
\url{https://doi.org/10.1109/MSN48538.2019.00033}
\showDOI{\tempurl}


\bibitem[Lu et~al\mbox{.}(2023)]%
        {10274418}
\bibfield{author}{\bibinfo{person}{Huali Lu}, \bibinfo{person}{Feng Lyu}, \bibinfo{person}{Huaqing Wu}, \bibinfo{person}{Jie Zhang}, \bibinfo{person}{Ju Ren}, \bibinfo{person}{Yaoxue Zhang}, {and} \bibinfo{person}{Xuemin Shen}.} \bibinfo{year}{2023}\natexlab{}.
\newblock \showarticletitle{FL-AMM: Federated Learning Augmented Map Matching With Heterogeneous Cellular Moving Trajectories}.
\newblock \bibinfo{journal}{\emph{IEEE Journal on Selected Areas in Communications}} \bibinfo{volume}{41}, \bibinfo{number}{12} (\bibinfo{year}{2023}), \bibinfo{pages}{3878--3892}.
\newblock
\urldef\tempurl%
\url{https://doi.org/10.1109/JSAC.2023.3322841}
\showDOI{\tempurl}


\bibitem[Luca et~al\mbox{.}(2021)]%
        {10.1145/3485125}
\bibfield{author}{\bibinfo{person}{Massimiliano Luca}, \bibinfo{person}{Gianni Barlacchi}, \bibinfo{person}{Bruno Lepri}, {and} \bibinfo{person}{Luca Pappalardo}.} \bibinfo{year}{2021}\natexlab{}.
\newblock \showarticletitle{A Survey on Deep Learning for Human Mobility}.
\newblock \bibinfo{journal}{\emph{ACM Comput. Surv.}} \bibinfo{volume}{55}, \bibinfo{number}{1}, Article \bibinfo{articleno}{7} (\bibinfo{date}{nov} \bibinfo{year}{2021}), \bibinfo{numpages}{44}~pages.
\newblock
\showISSN{0360-0300}
\urldef\tempurl%
\url{https://doi.org/10.1145/3485125}
\showDOI{\tempurl}


\bibitem[López-Espejo et~al\mbox{.}(2022)]%
        {9665775}
\bibfield{author}{\bibinfo{person}{Iván López-Espejo}, \bibinfo{person}{Zheng-Hua Tan}, \bibinfo{person}{John H.~L. Hansen}, {and} \bibinfo{person}{Jesper Jensen}.} \bibinfo{year}{2022}\natexlab{}.
\newblock \showarticletitle{Deep Spoken Keyword Spotting: An Overview}.
\newblock \bibinfo{journal}{\emph{IEEE Access}}  \bibinfo{volume}{10} (\bibinfo{year}{2022}), \bibinfo{pages}{4169--4199}.
\newblock
\urldef\tempurl%
\url{https://doi.org/10.1109/ACCESS.2021.3139508}
\showDOI{\tempurl}


\bibitem[Ma et~al\mbox{.}(2022)]%
        {ma2022state}
\bibfield{author}{\bibinfo{person}{Xiaodong Ma}, \bibinfo{person}{Jia Zhu}, \bibinfo{person}{Zhihao Lin}, \bibinfo{person}{Shanxuan Chen}, {and} \bibinfo{person}{Yangjie Qin}.} \bibinfo{year}{2022}\natexlab{}.
\newblock \showarticletitle{A state-of-the-art survey on solving non-IID data in Federated Learning}.
\newblock \bibinfo{journal}{\emph{Future Generation Computer Systems}}  \bibinfo{volume}{135} (\bibinfo{year}{2022}), \bibinfo{pages}{244--258}.
\newblock


\bibitem[Madan et~al\mbox{.}(2022)]%
        {10.1145/3571732}
\bibfield{author}{\bibinfo{person}{Chetan Madan}, \bibinfo{person}{Harshita Diddee}, \bibinfo{person}{Deepika Kumar}, {and} \bibinfo{person}{Mamta Mittal}.} \bibinfo{year}{2022}\natexlab{}.
\newblock \showarticletitle{CodeFed: Federated Speech Recognition for Low-Resource Code-Switching Detection}.
\newblock \bibinfo{journal}{\emph{ACM Trans. Asian Low-Resour. Lang. Inf. Process.}} (\bibinfo{date}{nov} \bibinfo{year}{2022}).
\newblock
\showISSN{2375-4699}
\urldef\tempurl%
\url{https://doi.org/10.1145/3571732}
\showDOI{\tempurl}
\newblock
\shownote{Just Accepted}.


\bibitem[Madry et~al\mbox{.}(2017)]%
        {madry2017towards}
\bibfield{author}{\bibinfo{person}{Aleksander Madry}, \bibinfo{person}{Aleksandar Makelov}, \bibinfo{person}{Ludwig Schmidt}, \bibinfo{person}{Dimitris Tsipras}, {and} \bibinfo{person}{Adrian Vladu}.} \bibinfo{year}{2017}\natexlab{}.
\newblock \showarticletitle{Towards deep learning models resistant to adversarial attacks}.
\newblock \bibinfo{journal}{\emph{arXiv preprint arXiv:1706.06083}} (\bibinfo{year}{2017}).
\newblock


\bibitem[McMahan et~al\mbox{.}(2017a)]%
        {pmlr-v54-mcmahan17a}
\bibfield{author}{\bibinfo{person}{Brendan McMahan}, \bibinfo{person}{Eider Moore}, \bibinfo{person}{Daniel Ramage}, \bibinfo{person}{Seth Hampson}, {and} \bibinfo{person}{Blaise Aguera~y Arcas}.} \bibinfo{year}{2017}\natexlab{a}.
\newblock \showarticletitle{{Communication-Efficient Learning of Deep Networks from Decentralized Data}}. In \bibinfo{booktitle}{\emph{Proceedings of the 20th International Conference on Artificial Intelligence and Statistics}} \emph{(\bibinfo{series}{Proceedings of Machine Learning Research}, Vol.~\bibinfo{volume}{54})}, \bibfield{editor}{\bibinfo{person}{Aarti Singh} {and} \bibinfo{person}{Jerry Zhu}} (Eds.). \bibinfo{publisher}{PMLR}, \bibinfo{pages}{1273--1282}.
\newblock
\urldef\tempurl%
\url{https://proceedings.mlr.press/v54/mcmahan17a.html}
\showURL{%
\tempurl}


\bibitem[McMahan et~al\mbox{.}(2017b)]%
        {mcmahan2017learning}
\bibfield{author}{\bibinfo{person}{H~Brendan McMahan}, \bibinfo{person}{Daniel Ramage}, \bibinfo{person}{Kunal Talwar}, {and} \bibinfo{person}{Li Zhang}.} \bibinfo{year}{2017}\natexlab{b}.
\newblock \showarticletitle{Learning differentially private recurrent language models}.
\newblock \bibinfo{journal}{\emph{arXiv preprint arXiv:1710.06963}} (\bibinfo{year}{2017}).
\newblock


\bibitem[McQuire et~al\mbox{.}(2021)]%
        {9669395}
\bibfield{author}{\bibinfo{person}{Jamie McQuire}, \bibinfo{person}{Paul Watson}, \bibinfo{person}{Nick Wright}, \bibinfo{person}{Hugo Hiden}, {and} \bibinfo{person}{Michael Catt}.} \bibinfo{year}{2021}\natexlab{}.
\newblock \showarticletitle{Uneven and Irregular Surface Condition Prediction from Human Walking Data using both Centralized and Decentralized Machine Learning Approaches}. In \bibinfo{booktitle}{\emph{2021 IEEE International Conference on Bioinformatics and Biomedicine (BIBM)}}. \bibinfo{pages}{1449--1452}.
\newblock
\urldef\tempurl%
\url{https://doi.org/10.1109/BIBM52615.2021.9669395}
\showDOI{\tempurl}


\bibitem[Meng et~al\mbox{.}(2023)]%
        {10094675}
\bibfield{author}{\bibinfo{person}{Dan Meng}, \bibinfo{person}{Xue Wang}, {and} \bibinfo{person}{Jun Wang}.} \bibinfo{year}{2023}\natexlab{}.
\newblock \showarticletitle{Backdoor Attack Against Automatic Speaker Verification Models in Federated Learning}. In \bibinfo{booktitle}{\emph{ICASSP 2023 - 2023 IEEE International Conference on Acoustics, Speech and Signal Processing (ICASSP)}}. \bibinfo{pages}{1--5}.
\newblock
\urldef\tempurl%
\url{https://doi.org/10.1109/ICASSP49357.2023.10094675}
\showDOI{\tempurl}


\bibitem[Meng et~al\mbox{.}(2022)]%
        {meng2022improving}
\bibfield{author}{\bibinfo{person}{Qiang Meng}, \bibinfo{person}{Feng Zhou}, \bibinfo{person}{Hainan Ren}, \bibinfo{person}{Tianshu Feng}, \bibinfo{person}{Guochao Liu}, {and} \bibinfo{person}{Yuanqing Lin}.} \bibinfo{year}{2022}\natexlab{}.
\newblock \showarticletitle{Improving federated learning face recognition via privacy-agnostic clusters}.
\newblock \bibinfo{journal}{\emph{arXiv preprint arXiv:2201.12467}} (\bibinfo{year}{2022}).
\newblock


\bibitem[Mensah et~al\mbox{.}(2022)]%
        {9922022}
\bibfield{author}{\bibinfo{person}{Daniel~Opoku Mensah}, \bibinfo{person}{Godwin Badu-Marfo}, \bibinfo{person}{Ranwa Al~Mallah}, {and} \bibinfo{person}{Bilal Farooq}.} \bibinfo{year}{2022}\natexlab{}.
\newblock \showarticletitle{eFedDNN: Ensemble based Federated Deep Neural Networks for Trajectory Mode Inference}. In \bibinfo{booktitle}{\emph{2022 IEEE International Smart Cities Conference (ISC2)}}. \bibinfo{pages}{1--7}.
\newblock
\urldef\tempurl%
\url{https://doi.org/10.1109/ISC255366.2022.9922022}
\showDOI{\tempurl}


\bibitem[Mistry et~al\mbox{.}(2023)]%
        {10196311}
\bibfield{author}{\bibinfo{person}{Durjoy Mistry}, \bibinfo{person}{M.~F. Mridha}, \bibinfo{person}{Mejdl Safran}, \bibinfo{person}{Sultan Alfarhood}, \bibinfo{person}{Aloke~Kumar Saha}, {and} \bibinfo{person}{Dunren Che}.} \bibinfo{year}{2023}\natexlab{}.
\newblock \showarticletitle{Privacy-Preserving On-Screen Activity Tracking and Classification in E-Learning Using Federated Learning}.
\newblock \bibinfo{journal}{\emph{IEEE Access}}  \bibinfo{volume}{11} (\bibinfo{year}{2023}), \bibinfo{pages}{79315--79329}.
\newblock
\urldef\tempurl%
\url{https://doi.org/10.1109/ACCESS.2023.3299331}
\showDOI{\tempurl}


\bibitem[Mohammadi et~al\mbox{.}(2023a)]%
        {10224953}
\bibfield{author}{\bibinfo{person}{Mohammadreza Mohammadi}, \bibinfo{person}{Roberto Allocca}, \bibinfo{person}{David Eklund}, \bibinfo{person}{Rakesh Shrestha}, {and} \bibinfo{person}{Sima Sinaei}.} \bibinfo{year}{2023}\natexlab{a}.
\newblock \showarticletitle{Privacy-preserving Federated Learning System for Fatigue Detection}. In \bibinfo{booktitle}{\emph{2023 IEEE International Conference on Cyber Security and Resilience (CSR)}}. \bibinfo{pages}{624--629}.
\newblock
\urldef\tempurl%
\url{https://doi.org/10.1109/CSR57506.2023.10224953}
\showDOI{\tempurl}


\bibitem[Mohammadi et~al\mbox{.}(2023b)]%
        {10306049}
\bibfield{author}{\bibinfo{person}{Samaneh Mohammadi}, \bibinfo{person}{Mohammadreza Mohammadi}, \bibinfo{person}{Sima Sinaei}, \bibinfo{person}{Ali Balador}, \bibinfo{person}{Ehsan Nowroozi}, \bibinfo{person}{Francesco Flammini}, {and} \bibinfo{person}{Mauro Conti}.} \bibinfo{year}{2023}\natexlab{b}.
\newblock \showarticletitle{Balancing Privacy and Accuracy in Federated Learning for Speech Emotion Recognition}. In \bibinfo{booktitle}{\emph{2023 18th Conference on Computer Science and Intelligence Systems (FedCSIS)}}. \bibinfo{pages}{191--199}.
\newblock
\urldef\tempurl%
\url{https://doi.org/10.15439/2023F444}
\showDOI{\tempurl}


\bibitem[Mohammadi et~al\mbox{.}(2023c)]%
        {10196918}
\bibfield{author}{\bibinfo{person}{Samaneh Mohammadi}, \bibinfo{person}{Sima Sinaei}, \bibinfo{person}{Ali Balador}, {and} \bibinfo{person}{Francesco Flammini}.} \bibinfo{year}{2023}\natexlab{c}.
\newblock \showarticletitle{Optimized Paillier Homomorphic Encryption in Federated Learning for Speech Emotion Recognition}. In \bibinfo{booktitle}{\emph{2023 IEEE 47th Annual Computers, Software, and Applications Conference (COMPSAC)}}. \bibinfo{pages}{1021--1022}.
\newblock
\urldef\tempurl%
\url{https://doi.org/10.1109/COMPSAC57700.2023.00156}
\showDOI{\tempurl}


\bibitem[Mohammadi et~al\mbox{.}(2023d)]%
        {mohammadi2023secure}
\bibfield{author}{\bibinfo{person}{Samaneh Mohammadi}, \bibinfo{person}{Sima Sinaei}, \bibinfo{person}{Ali Balador}, {and} \bibinfo{person}{Francesco Flammini}.} \bibinfo{year}{2023}\natexlab{d}.
\newblock \showarticletitle{Secure and efficient federated learning by combining homomorphic encryption and gradient pruning in speech emotion recognition}. In \bibinfo{booktitle}{\emph{International Conference on Information Security Practice and Experience}}. Springer, \bibinfo{pages}{1--16}.
\newblock


\bibitem[Moher et~al\mbox{.}(2010)]%
        {MOHER2010336}
\bibfield{author}{\bibinfo{person}{David Moher}, \bibinfo{person}{Alessandro Liberati}, \bibinfo{person}{Jennifer Tetzlaff}, {and} \bibinfo{person}{Douglas~G. Altman}.} \bibinfo{year}{2010}\natexlab{}.
\newblock \showarticletitle{Preferred reporting items for systematic reviews and meta-analyses: The PRISMA statement}.
\newblock \bibinfo{journal}{\emph{International Journal of Surgery}} \bibinfo{volume}{8}, \bibinfo{number}{5} (\bibinfo{year}{2010}), \bibinfo{pages}{336--341}.
\newblock
\showISSN{1743-9191}
\urldef\tempurl%
\url{https://doi.org/10.1016/j.ijsu.2010.02.007}
\showDOI{\tempurl}


\bibitem[Mothukuri et~al\mbox{.}(2021)]%
        {mothukuri2021survey}
\bibfield{author}{\bibinfo{person}{Viraaji Mothukuri}, \bibinfo{person}{Reza~M Parizi}, \bibinfo{person}{Seyedamin Pouriyeh}, \bibinfo{person}{Yan Huang}, \bibinfo{person}{Ali Dehghantanha}, {and} \bibinfo{person}{Gautam Srivastava}.} \bibinfo{year}{2021}\natexlab{}.
\newblock \showarticletitle{A survey on security and privacy of federated learning}.
\newblock \bibinfo{journal}{\emph{Future Generation Computer Systems}}  \bibinfo{volume}{115} (\bibinfo{year}{2021}), \bibinfo{pages}{619--640}.
\newblock


\bibitem[Nandi and Xhafa(2022)]%
        {nandi2022federated}
\bibfield{author}{\bibinfo{person}{Arijit Nandi} {and} \bibinfo{person}{Fatos Xhafa}.} \bibinfo{year}{2022}\natexlab{}.
\newblock \showarticletitle{A federated learning method for real-time emotion state classification from multi-modal streaming}.
\newblock \bibinfo{journal}{\emph{Methods}}  \bibinfo{volume}{204} (\bibinfo{year}{2022}), \bibinfo{pages}{340--347}.
\newblock


\bibitem[Nguyen et~al\mbox{.}(2021)]%
        {9415623}
\bibfield{author}{\bibinfo{person}{Dinh~C. Nguyen}, \bibinfo{person}{Ming Ding}, \bibinfo{person}{Pubudu~N. Pathirana}, \bibinfo{person}{Aruna Seneviratne}, \bibinfo{person}{Jun Li}, {and} \bibinfo{person}{H. Vincent~Poor}.} \bibinfo{year}{2021}\natexlab{}.
\newblock \showarticletitle{Federated Learning for Internet of Things: A Comprehensive Survey}.
\newblock \bibinfo{journal}{\emph{IEEE Communications Surveys \& Tutorials}} \bibinfo{volume}{23}, \bibinfo{number}{3} (\bibinfo{year}{2021}), \bibinfo{pages}{1622--1658}.
\newblock
\urldef\tempurl%
\url{https://doi.org/10.1109/COMST.2021.3075439}
\showDOI{\tempurl}


\bibitem[Nguyen et~al\mbox{.}(2022)]%
        {10.1145/3501296}
\bibfield{author}{\bibinfo{person}{Dinh~C. Nguyen}, \bibinfo{person}{Quoc-Viet Pham}, \bibinfo{person}{Pubudu~N. Pathirana}, \bibinfo{person}{Ming Ding}, \bibinfo{person}{Aruna Seneviratne}, \bibinfo{person}{Zihuai Lin}, \bibinfo{person}{Octavia Dobre}, {and} \bibinfo{person}{Won-Joo Hwang}.} \bibinfo{year}{2022}\natexlab{}.
\newblock \showarticletitle{Federated Learning for Smart Healthcare: A Survey}.
\newblock \bibinfo{journal}{\emph{ACM Comput. Surv.}} \bibinfo{volume}{55}, \bibinfo{number}{3}, Article \bibinfo{articleno}{60} (\bibinfo{date}{feb} \bibinfo{year}{2022}), \bibinfo{numpages}{37}~pages.
\newblock
\showISSN{0360-0300}
\urldef\tempurl%
\url{https://doi.org/10.1145/3501296}
\showDOI{\tempurl}


\bibitem[Nguyen et~al\mbox{.}(2023)]%
        {10096426}
\bibfield{author}{\bibinfo{person}{Tuan Nguyen}, \bibinfo{person}{Salima Mdhaffar}, \bibinfo{person}{Natalia Tomashenko}, \bibinfo{person}{Jean-François Bonastre}, {and} \bibinfo{person}{Yannick Estève}.} \bibinfo{year}{2023}\natexlab{}.
\newblock \showarticletitle{Federated Learning for ASR Based on wav2vec 2.0}. In \bibinfo{booktitle}{\emph{ICASSP 2023 - 2023 IEEE International Conference on Acoustics, Speech and Signal Processing (ICASSP)}}. \bibinfo{pages}{1--5}.
\newblock
\urldef\tempurl%
\url{https://doi.org/10.1109/ICASSP49357.2023.10096426}
\showDOI{\tempurl}


\bibitem[Nirmal et~al\mbox{.}(2021)]%
        {9353723}
\bibfield{author}{\bibinfo{person}{Isura Nirmal}, \bibinfo{person}{Abdelwahed Khamis}, \bibinfo{person}{Mahbub Hassan}, \bibinfo{person}{Wen Hu}, {and} \bibinfo{person}{Xiaoqing Zhu}.} \bibinfo{year}{2021}\natexlab{}.
\newblock \showarticletitle{Deep Learning for Radio-Based Human Sensing: Recent Advances and Future Directions}.
\newblock \bibinfo{journal}{\emph{IEEE Communications Surveys \& Tutorials}} \bibinfo{volume}{23}, \bibinfo{number}{2} (\bibinfo{year}{2021}), \bibinfo{pages}{995--1019}.
\newblock
\urldef\tempurl%
\url{https://doi.org/10.1109/COMST.2021.3058333}
\showDOI{\tempurl}


\bibitem[Niu and Deng(2022)]%
        {niu2022federated}
\bibfield{author}{\bibinfo{person}{Yifan Niu} {and} \bibinfo{person}{Weihong Deng}.} \bibinfo{year}{2022}\natexlab{}.
\newblock \showarticletitle{Federated learning for face recognition with gradient correction}. In \bibinfo{booktitle}{\emph{Proceedings of the AAAI Conference on Artificial Intelligence}}, Vol.~\bibinfo{volume}{36}. \bibinfo{pages}{1999--2007}.
\newblock


\bibitem[Novikova et~al\mbox{.}(2022)]%
        {novikova2022analysis}
\bibfield{author}{\bibinfo{person}{Evgenia Novikova}, \bibinfo{person}{Dmitry Fomichov}, \bibinfo{person}{Ivan Kholod}, {and} \bibinfo{person}{Evgeny Filippov}.} \bibinfo{year}{2022}\natexlab{}.
\newblock \showarticletitle{Analysis of privacy-enhancing technologies in open-source federated learning frameworks for driver activity recognition}.
\newblock \bibinfo{journal}{\emph{Sensors}} \bibinfo{volume}{22}, \bibinfo{number}{8} (\bibinfo{year}{2022}), \bibinfo{pages}{2983}.
\newblock


\bibitem[Ouyang et~al\mbox{.}(2021)]%
        {10.1145/3458864.3467681}
\bibfield{author}{\bibinfo{person}{Xiaomin Ouyang}, \bibinfo{person}{Zhiyuan Xie}, \bibinfo{person}{Jiayu Zhou}, \bibinfo{person}{Jianwei Huang}, {and} \bibinfo{person}{Guoliang Xing}.} \bibinfo{year}{2021}\natexlab{}.
\newblock \showarticletitle{ClusterFL: A Similarity-Aware Federated Learning System for Human Activity Recognition}. In \bibinfo{booktitle}{\emph{Proceedings of the 19th Annual International Conference on Mobile Systems, Applications, and Services}} (Virtual Event, Wisconsin) \emph{(\bibinfo{series}{MobiSys '21})}. \bibinfo{publisher}{Association for Computing Machinery}, \bibinfo{address}{New York, NY, USA}, \bibinfo{pages}{54–66}.
\newblock
\showISBNx{9781450384438}
\urldef\tempurl%
\url{https://doi.org/10.1145/3458864.3467681}
\showDOI{\tempurl}


\bibitem[Ouyang et~al\mbox{.}(2022)]%
        {10.1145/3554980}
\bibfield{author}{\bibinfo{person}{Xiaomin Ouyang}, \bibinfo{person}{Zhiyuan Xie}, \bibinfo{person}{Jiayu Zhou}, \bibinfo{person}{Guoliang Xing}, {and} \bibinfo{person}{Jianwei Huang}.} \bibinfo{year}{2022}\natexlab{}.
\newblock \showarticletitle{ClusterFL: A Clustering-Based Federated Learning System for Human Activity Recognition}.
\newblock \bibinfo{journal}{\emph{ACM Trans. Sen. Netw.}} \bibinfo{volume}{19}, \bibinfo{number}{1}, Article \bibinfo{articleno}{17} (\bibinfo{date}{dec} \bibinfo{year}{2022}), \bibinfo{numpages}{32}~pages.
\newblock
\showISSN{1550-4859}
\urldef\tempurl%
\url{https://doi.org/10.1145/3554980}
\showDOI{\tempurl}


\bibitem[Oza and Patel(2021)]%
        {9484338}
\bibfield{author}{\bibinfo{person}{Poojan Oza} {and} \bibinfo{person}{Vishal~M. Patel}.} \bibinfo{year}{2021}\natexlab{}.
\newblock \showarticletitle{Federated Learning-based Active Authentication on Mobile Devices}. In \bibinfo{booktitle}{\emph{2021 IEEE International Joint Conference on Biometrics (IJCB)}}. \bibinfo{pages}{1--8}.
\newblock
\urldef\tempurl%
\url{https://doi.org/10.1109/IJCB52358.2021.9484338}
\showDOI{\tempurl}


\bibitem[Pan and Yang(2010)]%
        {5288526}
\bibfield{author}{\bibinfo{person}{Sinno~Jialin Pan} {and} \bibinfo{person}{Qiang Yang}.} \bibinfo{year}{2010}\natexlab{}.
\newblock \showarticletitle{A Survey on Transfer Learning}.
\newblock \bibinfo{journal}{\emph{IEEE Transactions on Knowledge and Data Engineering}} \bibinfo{volume}{22}, \bibinfo{number}{10} (\bibinfo{year}{2010}), \bibinfo{pages}{1345--1359}.
\newblock
\urldef\tempurl%
\url{https://doi.org/10.1109/TKDE.2009.191}
\showDOI{\tempurl}


\bibitem[Panayotov et~al\mbox{.}(2015)]%
        {7178964}
\bibfield{author}{\bibinfo{person}{Vassil Panayotov}, \bibinfo{person}{Guoguo Chen}, \bibinfo{person}{Daniel Povey}, {and} \bibinfo{person}{Sanjeev Khudanpur}.} \bibinfo{year}{2015}\natexlab{}.
\newblock \showarticletitle{Librispeech: An ASR corpus based on public domain audio books}. In \bibinfo{booktitle}{\emph{2015 IEEE International Conference on Acoustics, Speech and Signal Processing (ICASSP)}}. \bibinfo{pages}{5206--5210}.
\newblock
\urldef\tempurl%
\url{https://doi.org/10.1109/ICASSP.2015.7178964}
\showDOI{\tempurl}


\bibitem[Pardau(2018)]%
        {pardau2018california}
\bibfield{author}{\bibinfo{person}{Stuart~L Pardau}.} \bibinfo{year}{2018}\natexlab{}.
\newblock \showarticletitle{The california consumer privacy act: Towards a european-style privacy regime in the united states}.
\newblock \bibinfo{journal}{\emph{J. Tech. L. \& Pol'y}}  \bibinfo{volume}{23} (\bibinfo{year}{2018}), \bibinfo{pages}{68}.
\newblock


\bibitem[Parisi et~al\mbox{.}(2019)]%
        {parisi2019continual}
\bibfield{author}{\bibinfo{person}{German~I Parisi}, \bibinfo{person}{Ronald Kemker}, \bibinfo{person}{Jose~L Part}, \bibinfo{person}{Christopher Kanan}, {and} \bibinfo{person}{Stefan Wermter}.} \bibinfo{year}{2019}\natexlab{}.
\newblock \showarticletitle{Continual lifelong learning with neural networks: A review}.
\newblock \bibinfo{journal}{\emph{Neural networks}}  \bibinfo{volume}{113} (\bibinfo{year}{2019}), \bibinfo{pages}{54--71}.
\newblock


\bibitem[Park et~al\mbox{.}(2022)]%
        {9764747}
\bibfield{author}{\bibinfo{person}{Junha Park}, \bibinfo{person}{Jiseon Moon}, \bibinfo{person}{Taekyoon Kim}, \bibinfo{person}{Peng Wu}, \bibinfo{person}{Tales Imbiriba}, \bibinfo{person}{Pau Closas}, {and} \bibinfo{person}{Sunwoo Kim}.} \bibinfo{year}{2022}\natexlab{}.
\newblock \showarticletitle{Federated Learning for Indoor Localization via Model Reliability With Dropout}.
\newblock \bibinfo{journal}{\emph{IEEE Communications Letters}} \bibinfo{volume}{26}, \bibinfo{number}{7} (\bibinfo{year}{2022}), \bibinfo{pages}{1553--1557}.
\newblock
\urldef\tempurl%
\url{https://doi.org/10.1109/LCOMM.2022.3170878}
\showDOI{\tempurl}


\bibitem[Peng et~al\mbox{.}(2019)]%
        {peng2019federated}
\bibfield{author}{\bibinfo{person}{Xingchao Peng}, \bibinfo{person}{Zijun Huang}, \bibinfo{person}{Yizhe Zhu}, {and} \bibinfo{person}{Kate Saenko}.} \bibinfo{year}{2019}\natexlab{}.
\newblock \showarticletitle{Federated adversarial domain adaptation}.
\newblock \bibinfo{journal}{\emph{arXiv preprint arXiv:1911.02054}} (\bibinfo{year}{2019}).
\newblock


\bibitem[Pfitzmann and Hansen(2010)]%
        {pfitzmann2010terminology}
\bibfield{author}{\bibinfo{person}{Andreas Pfitzmann} {and} \bibinfo{person}{Marit Hansen}.} \bibinfo{year}{2010}\natexlab{}.
\newblock \bibinfo{title}{A terminology for talking about privacy by data minimization: Anonymity, unlinkability, undetectability, unobservability, pseudonymity, and identity management}.
\newblock
\newblock


\bibitem[Pfitzner et~al\mbox{.}(2021)]%
        {10.1145/3412357}
\bibfield{author}{\bibinfo{person}{Bjarne Pfitzner}, \bibinfo{person}{Nico Steckhan}, {and} \bibinfo{person}{Bert Arnrich}.} \bibinfo{year}{2021}\natexlab{}.
\newblock \showarticletitle{Federated Learning in a Medical Context: A Systematic Literature Review}.
\newblock \bibinfo{journal}{\emph{ACM Trans. Internet Technol.}} \bibinfo{volume}{21}, \bibinfo{number}{2}, Article \bibinfo{articleno}{50} (\bibinfo{date}{jun} \bibinfo{year}{2021}), \bibinfo{numpages}{31}~pages.
\newblock
\showISSN{1533-5399}
\urldef\tempurl%
\url{https://doi.org/10.1145/3412357}
\showDOI{\tempurl}


\bibitem[Pham et~al\mbox{.}(2023)]%
        {pham2023personalized}
\bibfield{author}{\bibinfo{person}{Vinh Pham}, \bibinfo{person}{Yongho Lee}, {and} \bibinfo{person}{Tai-Myoung Chung}.} \bibinfo{year}{2023}\natexlab{}.
\newblock \showarticletitle{Personalized Stress Detection System Using Physiological Data from Wearable Sensors}. In \bibinfo{booktitle}{\emph{International Conference on Future Data and Security Engineering}}. Springer, \bibinfo{pages}{433--441}.
\newblock


\bibitem[Povey et~al\mbox{.}(2011)]%
        {povey2011kaldi}
\bibfield{author}{\bibinfo{person}{Daniel Povey}, \bibinfo{person}{Arnab Ghoshal}, \bibinfo{person}{Gilles Boulianne}, \bibinfo{person}{Lukas Burget}, \bibinfo{person}{Ondrej Glembek}, \bibinfo{person}{Nagendra Goel}, \bibinfo{person}{Mirko Hannemann}, \bibinfo{person}{Petr Motlicek}, \bibinfo{person}{Yanmin Qian}, \bibinfo{person}{Petr Schwarz}, {et~al\mbox{.}}} \bibinfo{year}{2011}\natexlab{}.
\newblock \showarticletitle{The Kaldi speech recognition toolkit}. In \bibinfo{booktitle}{\emph{IEEE 2011 workshop on automatic speech recognition and understanding}}. IEEE Signal Processing Society.
\newblock


\bibitem[Pranto and Al~Asad(2021)]%
        {9885430}
\bibfield{author}{\bibinfo{person}{Md. Appel~Mahmud Pranto} {and} \bibinfo{person}{Nafiz Al~Asad}.} \bibinfo{year}{2021}\natexlab{}.
\newblock \showarticletitle{A Comprehensive Model to Monitor Mental Health based on Federated Learning and Deep Learning}. In \bibinfo{booktitle}{\emph{2021 IEEE International Conference on Signal Processing, Information, Communication \& Systems (SPICSCON)}}. \bibinfo{pages}{18--21}.
\newblock
\urldef\tempurl%
\url{https://doi.org/10.1109/SPICSCON54707.2021.9885430}
\showDOI{\tempurl}


\bibitem[Presotto et~al\mbox{.}(2022a)]%
        {9762352}
\bibfield{author}{\bibinfo{person}{Riccardo Presotto}, \bibinfo{person}{Gabriele Civitarese}, {and} \bibinfo{person}{Claudio Bettini}.} \bibinfo{year}{2022}\natexlab{a}.
\newblock \showarticletitle{FedCLAR: Federated Clustering for Personalized Sensor-Based Human Activity Recognition}. In \bibinfo{booktitle}{\emph{2022 IEEE International Conference on Pervasive Computing and Communications (PerCom)}}. \bibinfo{pages}{227--236}.
\newblock
\urldef\tempurl%
\url{https://doi.org/10.1109/PerCom53586.2022.9762352}
\showDOI{\tempurl}


\bibitem[Presotto et~al\mbox{.}(2022b)]%
        {presotto2022federated}
\bibfield{author}{\bibinfo{person}{Riccardo Presotto}, \bibinfo{person}{Gabriele Civitarese}, {and} \bibinfo{person}{Claudio Bettini}.} \bibinfo{year}{2022}\natexlab{b}.
\newblock \showarticletitle{Federated clustering and semi-supervised learning: a new partnership for personalized human activity recognition}.
\newblock \bibinfo{journal}{\emph{Pervasive and Mobile Computing}}  \bibinfo{volume}{88} (\bibinfo{year}{2022}), \bibinfo{pages}{101726}.
\newblock


\bibitem[Presotto et~al\mbox{.}(2022c)]%
        {9767215}
\bibfield{author}{\bibinfo{person}{Riccardo Presotto}, \bibinfo{person}{Gabriele Civitarese}, {and} \bibinfo{person}{Claudio Bettini}.} \bibinfo{year}{2022}\natexlab{c}.
\newblock \showarticletitle{Preliminary Results on Sensitive Data Leakage in Federated Human Activity Recognition}. In \bibinfo{booktitle}{\emph{2022 IEEE International Conference on Pervasive Computing and Communications Workshops and other Affiliated Events (PerCom Workshops)}}. \bibinfo{pages}{304--309}.
\newblock
\urldef\tempurl%
\url{https://doi.org/10.1109/PerComWorkshops53856.2022.9767215}
\showDOI{\tempurl}


\bibitem[Presotto et~al\mbox{.}(2022d)]%
        {presotto2022semi}
\bibfield{author}{\bibinfo{person}{Riccardo Presotto}, \bibinfo{person}{Gabriele Civitarese}, {and} \bibinfo{person}{Claudio Bettini}.} \bibinfo{year}{2022}\natexlab{d}.
\newblock \showarticletitle{Semi-supervised and personalized federated activity recognition based on active learning and label propagation}.
\newblock \bibinfo{journal}{\emph{Personal and Ubiquitous Computing}} \bibinfo{volume}{26}, \bibinfo{number}{5} (\bibinfo{year}{2022}), \bibinfo{pages}{1281--1298}.
\newblock


\bibitem[Proteasa et~al\mbox{.}(2023)]%
        {10257271}
\bibfield{author}{\bibinfo{person}{Vlad-Alexandru Proteasa}, \bibinfo{person}{Radu-Ioan Ciobanu}, \bibinfo{person}{Ciprian Dobre}, {and} \bibinfo{person}{Radu-Corneliu Marin}.} \bibinfo{year}{2023}\natexlab{}.
\newblock \showarticletitle{Federated Learning for Human Mobility}. In \bibinfo{booktitle}{\emph{2023 19th International Conference on Distributed Computing in Smart Systems and the Internet of Things (DCOSS-IoT)}}. \bibinfo{pages}{780--785}.
\newblock
\urldef\tempurl%
\url{https://doi.org/10.1109/DCOSS-IoT58021.2023.00122}
\showDOI{\tempurl}


\bibitem[Psaltis et~al\mbox{.}(2022)]%
        {psaltis2022deep}
\bibfield{author}{\bibinfo{person}{Athanasios Psaltis}, \bibinfo{person}{Charalampos~Z Patrikakis}, {and} \bibinfo{person}{Petros Daras}.} \bibinfo{year}{2022}\natexlab{}.
\newblock \showarticletitle{Deep Multi-Modal Representation Schemes For Federated 3D Human Action Recognition}. In \bibinfo{booktitle}{\emph{European Conference on Computer Vision}}. Springer, \bibinfo{pages}{334--352}.
\newblock


\bibitem[Puri et~al\mbox{.}(2021)]%
        {9630505}
\bibfield{author}{\bibinfo{person}{Chetanya Puri}, \bibinfo{person}{Koustabh Dolui}, \bibinfo{person}{Gerben Kooijman}, \bibinfo{person}{Felipe Masculo}, \bibinfo{person}{Shannon Van~Sambeek}, \bibinfo{person}{Sebastiaan Den~Boer}, \bibinfo{person}{Sam Michiels}, \bibinfo{person}{Hans Hallez}, \bibinfo{person}{Stijn Luca}, {and} \bibinfo{person}{Bart Vanrumste}.} \bibinfo{year}{2021}\natexlab{}.
\newblock \showarticletitle{Gestational weight gain prediction using privacy preserving federated learning}. In \bibinfo{booktitle}{\emph{2021 43rd Annual International Conference of the IEEE Engineering in Medicine \& Biology Society (EMBC)}}. \bibinfo{pages}{2170--2174}.
\newblock
\urldef\tempurl%
\url{https://doi.org/10.1109/EMBC46164.2021.9630505}
\showDOI{\tempurl}


\bibitem[Qayyum et~al\mbox{.}(2021)]%
        {9153891}
\bibfield{author}{\bibinfo{person}{Adnan Qayyum}, \bibinfo{person}{Junaid Qadir}, \bibinfo{person}{Muhammad Bilal}, {and} \bibinfo{person}{Ala Al-Fuqaha}.} \bibinfo{year}{2021}\natexlab{}.
\newblock \showarticletitle{Secure and Robust Machine Learning for Healthcare: A Survey}.
\newblock \bibinfo{journal}{\emph{IEEE Reviews in Biomedical Engineering}}  \bibinfo{volume}{14} (\bibinfo{year}{2021}), \bibinfo{pages}{156--180}.
\newblock
\urldef\tempurl%
\url{https://doi.org/10.1109/RBME.2020.3013489}
\showDOI{\tempurl}


\bibitem[Qi et~al\mbox{.}(2022b)]%
        {10.1145/3503161.3548278}
\bibfield{author}{\bibinfo{person}{Fan Qi}, \bibinfo{person}{Zixin Zhang}, \bibinfo{person}{Xianshan Yang}, \bibinfo{person}{Huaiwen Zhang}, {and} \bibinfo{person}{Changsheng Xu}.} \bibinfo{year}{2022}\natexlab{b}.
\newblock \showarticletitle{Feeling Without Sharing: A Federated Video Emotion Recognition Framework Via Privacy-Agnostic Hybrid Aggregation}. In \bibinfo{booktitle}{\emph{Proceedings of the 30th ACM International Conference on Multimedia}} (<conf-loc>, <city>Lisboa</city>, <country>Portugal</country>, </conf-loc>) \emph{(\bibinfo{series}{MM '22})}. \bibinfo{publisher}{Association for Computing Machinery}, \bibinfo{address}{New York, NY, USA}, \bibinfo{pages}{151–160}.
\newblock
\showISBNx{9781450392037}
\urldef\tempurl%
\url{https://doi.org/10.1145/3503161.3548278}
\showDOI{\tempurl}


\bibitem[Qi et~al\mbox{.}(2022a)]%
        {10.1145/3556562.3558575}
\bibfield{author}{\bibinfo{person}{Wanbin Qi}, \bibinfo{person}{Yanxi Xie}, \bibinfo{person}{Hao Zhang}, \bibinfo{person}{Jiaen Zhou}, \bibinfo{person}{Ronghui Zhang}, {and} \bibinfo{person}{Xiaojun Jing}.} \bibinfo{year}{2022}\natexlab{a}.
\newblock \showarticletitle{An Efficient Cross-Domain Device-Free Gesture Recognition Method for ISAC with Federated Transfer Learning}. In \bibinfo{booktitle}{\emph{Proceedings of the 1st ACM MobiCom Workshop on Integrated Sensing and Communications Systems}} (Sydney, NSW, Australia) \emph{(\bibinfo{series}{ISACom '22})}. \bibinfo{publisher}{Association for Computing Machinery}, \bibinfo{address}{New York, NY, USA}, \bibinfo{pages}{67–71}.
\newblock
\showISBNx{9781450395250}
\urldef\tempurl%
\url{https://doi.org/10.1145/3556562.3558575}
\showDOI{\tempurl}


\bibitem[Qi et~al\mbox{.}(2023)]%
        {10005134}
\bibfield{author}{\bibinfo{person}{Wanbin Qi}, \bibinfo{person}{Ronghui Zhang}, \bibinfo{person}{Jiaen Zhou}, \bibinfo{person}{Hao Zhang}, \bibinfo{person}{Yanxi Xie}, {and} \bibinfo{person}{Xiaojun Jing}.} \bibinfo{year}{2023}\natexlab{}.
\newblock \showarticletitle{A Resource-Efficient Cross-Domain Sensing Method for Device-Free Gesture Recognition With Federated Transfer Learning}.
\newblock \bibinfo{journal}{\emph{IEEE Transactions on Green Communications and Networking}} \bibinfo{volume}{7}, \bibinfo{number}{1} (\bibinfo{year}{2023}), \bibinfo{pages}{393--400}.
\newblock
\urldef\tempurl%
\url{https://doi.org/10.1109/TGCN.2022.3233825}
\showDOI{\tempurl}


\bibitem[Qirtas et~al\mbox{.}(2022)]%
        {9861090}
\bibfield{author}{\bibinfo{person}{Malik~Muhammad Qirtas}, \bibinfo{person}{Dirk Pesch}, \bibinfo{person}{Evi Zafeiridi}, {and} \bibinfo{person}{Eleanor~Bantry White}.} \bibinfo{year}{2022}\natexlab{}.
\newblock \showarticletitle{Privacy Preserving Loneliness Detection: A Federated Learning Approach}. In \bibinfo{booktitle}{\emph{2022 IEEE International Conference on Digital Health (ICDH)}}. \bibinfo{pages}{157--162}.
\newblock
\urldef\tempurl%
\url{https://doi.org/10.1109/ICDH55609.2022.00032}
\showDOI{\tempurl}


\bibitem[Qiu et~al\mbox{.}(2022)]%
        {qiu2022multi}
\bibfield{author}{\bibinfo{person}{Sen Qiu}, \bibinfo{person}{Hongkai Zhao}, \bibinfo{person}{Nan Jiang}, \bibinfo{person}{Zhelong Wang}, \bibinfo{person}{Long Liu}, \bibinfo{person}{Yi An}, \bibinfo{person}{Hongyu Zhao}, \bibinfo{person}{Xin Miao}, \bibinfo{person}{Ruichen Liu}, {and} \bibinfo{person}{Giancarlo Fortino}.} \bibinfo{year}{2022}\natexlab{}.
\newblock \showarticletitle{Multi-sensor information fusion based on machine learning for real applications in human activity recognition: State-of-the-art and research challenges}.
\newblock \bibinfo{journal}{\emph{Information Fusion}}  \bibinfo{volume}{80} (\bibinfo{year}{2022}), \bibinfo{pages}{241--265}.
\newblock


\bibitem[Rafi et~al\mbox{.}(2022)]%
        {rafi2022personalization}
\bibfield{author}{\bibinfo{person}{Houda Rafi}, \bibinfo{person}{Yannick Benezeth}, \bibinfo{person}{Philippe Reynaud}, \bibinfo{person}{Emmanuel Arnoux}, \bibinfo{person}{Fan~Yang Song}, {and} \bibinfo{person}{Cedric Demonceaux}.} \bibinfo{year}{2022}\natexlab{}.
\newblock \showarticletitle{Personalization of AI Models Based on Federated Learning for Driver Stress Monitoring}. In \bibinfo{booktitle}{\emph{European Conference on Computer Vision}}. Springer, \bibinfo{pages}{575--585}.
\newblock


\bibitem[Ramaswamy et~al\mbox{.}(2019)]%
        {ramaswamy2019federated}
\bibfield{author}{\bibinfo{person}{Swaroop Ramaswamy}, \bibinfo{person}{Rajiv Mathews}, \bibinfo{person}{Kanishka Rao}, {and} \bibinfo{person}{Fran{\c{c}}oise Beaufays}.} \bibinfo{year}{2019}\natexlab{}.
\newblock \showarticletitle{Federated learning for emoji prediction in a mobile keyboard}.
\newblock \bibinfo{journal}{\emph{arXiv preprint arXiv:1906.04329}} (\bibinfo{year}{2019}).
\newblock


\bibitem[Rani et~al\mbox{.}(2023)]%
        {rani2023federated}
\bibfield{author}{\bibinfo{person}{Sita Rani}, \bibinfo{person}{Aman Kataria}, \bibinfo{person}{Sachin Kumar}, {and} \bibinfo{person}{Prayag Tiwari}.} \bibinfo{year}{2023}\natexlab{}.
\newblock \showarticletitle{Federated learning for secure IoMT-applications in smart healthcare systems: A comprehensive review}.
\newblock \bibinfo{journal}{\emph{Knowledge-Based Systems}} (\bibinfo{year}{2023}), \bibinfo{pages}{110658}.
\newblock


\bibitem[Regulation(2016)]%
        {regulation2016regulation}
\bibfield{author}{\bibinfo{person}{Protection Regulation}.} \bibinfo{year}{2016}\natexlab{}.
\newblock \showarticletitle{Regulation (EU) 2016/679 of the European Parliament and of the Council}.
\newblock \bibinfo{journal}{\emph{Regulation (eu)}}  \bibinfo{volume}{679} (\bibinfo{year}{2016}), \bibinfo{pages}{2016}.
\newblock


\bibitem[Rieke et~al\mbox{.}(2020)]%
        {rieke2020future}
\bibfield{author}{\bibinfo{person}{Nicola Rieke}, \bibinfo{person}{Jonny Hancox}, \bibinfo{person}{Wenqi Li}, \bibinfo{person}{Fausto Milletari}, \bibinfo{person}{Holger~R Roth}, \bibinfo{person}{Shadi Albarqouni}, \bibinfo{person}{Spyridon Bakas}, \bibinfo{person}{Mathieu~N Galtier}, \bibinfo{person}{Bennett~A Landman}, \bibinfo{person}{Klaus Maier-Hein}, {et~al\mbox{.}}} \bibinfo{year}{2020}\natexlab{}.
\newblock \showarticletitle{The future of digital health with federated learning}.
\newblock \bibinfo{journal}{\emph{NPJ digital medicine}} \bibinfo{volume}{3}, \bibinfo{number}{1} (\bibinfo{year}{2020}), \bibinfo{pages}{119}.
\newblock


\bibitem[Roh and Fang(2022)]%
        {10185525}
\bibfield{author}{\bibinfo{person}{Jaechul Roh} {and} \bibinfo{person}{Yajun Fang}.} \bibinfo{year}{2022}\natexlab{}.
\newblock \showarticletitle{Robust Smart Home Face Recognition Under Starving Federated Data}. In \bibinfo{booktitle}{\emph{2022 6th International Conference on Universal Village (UV)}}. \bibinfo{pages}{1--11}.
\newblock
\urldef\tempurl%
\url{https://doi.org/10.1109/UV56588.2022.10185525}
\showDOI{\tempurl}


\bibitem[Roy et~al\mbox{.}(2023)]%
        {10.1145/3594739.3610675}
\bibfield{author}{\bibinfo{person}{Debaditya Roy}, \bibinfo{person}{Ahmed Lekssays}, \bibinfo{person}{Sarunas Girdzijauskas}, \bibinfo{person}{Barbara Carminati}, {and} \bibinfo{person}{Elena Ferrari}.} \bibinfo{year}{2023}\natexlab{}.
\newblock \showarticletitle{Private, Fair and Secure Collaborative Learning Framework for Human Activity Recognition}. In \bibinfo{booktitle}{\emph{Adjunct Proceedings of the 2023 ACM International Joint Conference on Pervasive and Ubiquitous Computing \& the 2023 ACM International Symposium on Wearable Computing}} \emph{(\bibinfo{series}{UbiComp/ISWC '23 Adjunct})}. \bibinfo{publisher}{Association for Computing Machinery}, \bibinfo{address}{New York, NY, USA}, \bibinfo{pages}{352–358}.
\newblock
\showISBNx{9798400702006}
\urldef\tempurl%
\url{https://doi.org/10.1145/3594739.3610675}
\showDOI{\tempurl}


\bibitem[Sachin et~al\mbox{.}(2022)]%
        {sachin2022federated}
\bibfield{author}{\bibinfo{person}{DN Sachin}, \bibinfo{person}{B Annappa}, {and} \bibinfo{person}{Sateesh Ambesenge}.} \bibinfo{year}{2022}\natexlab{}.
\newblock \showarticletitle{Federated learning for wearable sensor-based human activity recognition}. In \bibinfo{booktitle}{\emph{International Conference on Intelligent Technologies}}. Springer, \bibinfo{pages}{131--139}.
\newblock


\bibitem[Salamon et~al\mbox{.}(2014)]%
        {10.1145/2647868.2655045}
\bibfield{author}{\bibinfo{person}{Justin Salamon}, \bibinfo{person}{Christopher Jacoby}, {and} \bibinfo{person}{Juan~Pablo Bello}.} \bibinfo{year}{2014}\natexlab{}.
\newblock \showarticletitle{A Dataset and Taxonomy for Urban Sound Research}. In \bibinfo{booktitle}{\emph{Proceedings of the 22nd ACM International Conference on Multimedia}} (Orlando, Florida, USA) \emph{(\bibinfo{series}{MM '14})}. \bibinfo{publisher}{Association for Computing Machinery}, \bibinfo{address}{New York, NY, USA}, \bibinfo{pages}{1041–1044}.
\newblock
\showISBNx{9781450330633}
\urldef\tempurl%
\url{https://doi.org/10.1145/2647868.2655045}
\showDOI{\tempurl}


\bibitem[Salman and Busso(2022)]%
        {10.1145/3536221.3556614}
\bibfield{author}{\bibinfo{person}{Ali Salman} {and} \bibinfo{person}{Carlos Busso}.} \bibinfo{year}{2022}\natexlab{}.
\newblock \showarticletitle{Privacy Preserving Personalization for Video Facial Expression Recognition Using Federated Learning}. In \bibinfo{booktitle}{\emph{Proceedings of the 2022 International Conference on Multimodal Interaction}} (Bengaluru, India) \emph{(\bibinfo{series}{ICMI '22})}. \bibinfo{publisher}{Association for Computing Machinery}, \bibinfo{address}{New York, NY, USA}, \bibinfo{pages}{495–503}.
\newblock
\showISBNx{9781450393904}
\urldef\tempurl%
\url{https://doi.org/10.1145/3536221.3556614}
\showDOI{\tempurl}


\bibitem[Sanchez et~al\mbox{.}(2023)]%
        {sanchez2023federated}
\bibfield{author}{\bibinfo{person}{Sergio Sanchez}, \bibinfo{person}{Javier Machacuay}, {and} \bibinfo{person}{Mario Quinde}.} \bibinfo{year}{2023}\natexlab{}.
\newblock \showarticletitle{Federated Learning for Human Activity Recognition on the MHealth Dataset}. In \bibinfo{booktitle}{\emph{International Conference on Artificial Intelligence and Soft Computing}}. Springer, \bibinfo{pages}{215--225}.
\newblock


\bibitem[Sarkar et~al\mbox{.}(2021)]%
        {9680185}
\bibfield{author}{\bibinfo{person}{Abhishek Sarkar}, \bibinfo{person}{Tanmay Sen}, {and} \bibinfo{person}{Ashis~Kumar Roy}.} \bibinfo{year}{2021}\natexlab{}.
\newblock \showarticletitle{GraFeHTy: Graph Neural Network using Federated Learning for Human Activity Recognition}. In \bibinfo{booktitle}{\emph{2021 20th IEEE International Conference on Machine Learning and Applications (ICMLA)}}. \bibinfo{pages}{1124--1129}.
\newblock
\urldef\tempurl%
\url{https://doi.org/10.1109/ICMLA52953.2021.00184}
\showDOI{\tempurl}


\bibitem[Sazdov et~al\mbox{.}(2023)]%
        {10.1145/3594739.3611323}
\bibfield{author}{\bibinfo{person}{Borjan Sazdov}, \bibinfo{person}{Bojan Jakimovski}, \bibinfo{person}{Simon Stankoski}, \bibinfo{person}{Ivana Kiprijanovska}, \bibinfo{person}{Bojan Sofronievski}, \bibinfo{person}{Martin Gjoreski}, \bibinfo{person}{Charles Nduka}, {and} \bibinfo{person}{Hristijan Gjoreski}.} \bibinfo{year}{2023}\natexlab{}.
\newblock \showarticletitle{Privacy-Aware Human Activity Recognition with Smart Glasses for Digital Therapeutics}. In \bibinfo{booktitle}{\emph{Adjunct Proceedings of the 2023 ACM International Joint Conference on Pervasive and Ubiquitous Computing \& the 2023 ACM International Symposium on Wearable Computing}} (Cancun, Quintana Roo, Mexico) \emph{(\bibinfo{series}{UbiComp/ISWC '23 Adjunct})}. \bibinfo{publisher}{Association for Computing Machinery}, \bibinfo{address}{New York, NY, USA}, \bibinfo{pages}{592–596}.
\newblock
\showISBNx{9798400702006}
\urldef\tempurl%
\url{https://doi.org/10.1145/3594739.3611323}
\showDOI{\tempurl}


\bibitem[Schmidt et~al\mbox{.}(2018)]%
        {10.1145/3242969.3242985}
\bibfield{author}{\bibinfo{person}{Philip Schmidt}, \bibinfo{person}{Attila Reiss}, \bibinfo{person}{Robert Duerichen}, \bibinfo{person}{Claus Marberger}, {and} \bibinfo{person}{Kristof Van~Laerhoven}.} \bibinfo{year}{2018}\natexlab{}.
\newblock \showarticletitle{Introducing WESAD, a Multimodal Dataset for Wearable Stress and Affect Detection}. In \bibinfo{booktitle}{\emph{Proceedings of the 20th ACM International Conference on Multimodal Interaction}} (Boulder, CO, USA) \emph{(\bibinfo{series}{ICMI '18})}. \bibinfo{publisher}{Association for Computing Machinery}, \bibinfo{address}{New York, NY, USA}, \bibinfo{pages}{400–408}.
\newblock
\showISBNx{9781450356923}
\urldef\tempurl%
\url{https://doi.org/10.1145/3242969.3242985}
\showDOI{\tempurl}


\bibitem[Shaik et~al\mbox{.}(2022)]%
        {SHAIK2022109929}
\bibfield{author}{\bibinfo{person}{Thanveer Shaik}, \bibinfo{person}{Xiaohui Tao}, \bibinfo{person}{Niall Higgins}, \bibinfo{person}{Raj Gururajan}, \bibinfo{person}{Yuefeng Li}, \bibinfo{person}{Xujuan Zhou}, {and} \bibinfo{person}{U.~Rajendra Acharya}.} \bibinfo{year}{2022}\natexlab{}.
\newblock \showarticletitle{FedStack: Personalized activity monitoring using stacked federated learning}.
\newblock \bibinfo{journal}{\emph{Knowledge-Based Systems}}  \bibinfo{volume}{257} (\bibinfo{year}{2022}), \bibinfo{pages}{109929}.
\newblock
\showISSN{0950-7051}
\urldef\tempurl%
\url{https://doi.org/10.1016/j.knosys.2022.109929}
\showDOI{\tempurl}


\bibitem[Shang et~al\mbox{.}(2023)]%
        {10041112}
\bibfield{author}{\bibinfo{person}{Ertong Shang}, \bibinfo{person}{Hui Liu}, \bibinfo{person}{Zhuo Yang}, \bibinfo{person}{Junzhao Du}, {and} \bibinfo{person}{Yiming Ge}.} \bibinfo{year}{2023}\natexlab{}.
\newblock \showarticletitle{FedBiKD: Federated Bidirectional Knowledge Distillation for Distracted Driving Detection}.
\newblock \bibinfo{journal}{\emph{IEEE Internet of Things Journal}} \bibinfo{volume}{10}, \bibinfo{number}{13} (\bibinfo{year}{2023}), \bibinfo{pages}{11643--11654}.
\newblock
\urldef\tempurl%
\url{https://doi.org/10.1109/JIOT.2023.3243622}
\showDOI{\tempurl}


\bibitem[Shao et~al\mbox{.}(2022)]%
        {9780603}
\bibfield{author}{\bibinfo{person}{Rui Shao}, \bibinfo{person}{Pramuditha Perera}, \bibinfo{person}{Pong~C. Yuen}, {and} \bibinfo{person}{Vishal~M. Patel}.} \bibinfo{year}{2022}\natexlab{}.
\newblock \showarticletitle{Federated Generalized Face Presentation Attack Detection}.
\newblock \bibinfo{journal}{\emph{IEEE Transactions on Neural Networks and Learning Systems}} (\bibinfo{year}{2022}), \bibinfo{pages}{1--14}.
\newblock
\urldef\tempurl%
\url{https://doi.org/10.1109/TNNLS.2022.3172316}
\showDOI{\tempurl}


\bibitem[Shen et~al\mbox{.}(2023)]%
        {shen2023federated}
\bibfield{author}{\bibinfo{person}{Qiang Shen}, \bibinfo{person}{Haotian Feng}, \bibinfo{person}{Rui Song}, \bibinfo{person}{Donglei Song}, {and} \bibinfo{person}{Hao Xu}.} \bibinfo{year}{2023}\natexlab{}.
\newblock \showarticletitle{Federated Meta-Learning with Attention for Diversity-Aware Human Activity Recognition}.
\newblock \bibinfo{journal}{\emph{Sensors}} \bibinfo{volume}{23}, \bibinfo{number}{3} (\bibinfo{year}{2023}), \bibinfo{pages}{1083}.
\newblock


\bibitem[Shen et~al\mbox{.}(2022)]%
        {shen2022federated}
\bibfield{author}{\bibinfo{person}{Qiang Shen}, \bibinfo{person}{Haotian Feng}, \bibinfo{person}{Rui Song}, \bibinfo{person}{Stefano Teso}, \bibinfo{person}{Fausto Giunchiglia}, \bibinfo{person}{Hao Xu}, {et~al\mbox{.}}} \bibinfo{year}{2022}\natexlab{}.
\newblock \showarticletitle{Federated Multi-Task Attention for Cross-Individual Human Activity Recognition}. In \bibinfo{booktitle}{\emph{IJCAI}}. IJCAI, \bibinfo{pages}{3423--3429}.
\newblock


\bibitem[Shome and Kar(2021)]%
        {Shome_2021_ICCV}
\bibfield{author}{\bibinfo{person}{Debaditya Shome} {and} \bibinfo{person}{Tejaswini Kar}.} \bibinfo{year}{2021}\natexlab{}.
\newblock \showarticletitle{FedAffect: Few-Shot Federated Learning for Facial Expression Recognition}. In \bibinfo{booktitle}{\emph{Proceedings of the IEEE/CVF International Conference on Computer Vision (ICCV) Workshops}}. \bibinfo{pages}{4168--4175}.
\newblock


\bibitem[Siddiqui et~al\mbox{.}(2022)]%
        {9765165}
\bibfield{author}{\bibinfo{person}{Md. Saiful~Bari Siddiqui}, \bibinfo{person}{Sanjida~Ali Shusmita}, \bibinfo{person}{Shareea Sabreen}, {and} \bibinfo{person}{Md. Golam~Rabiul Alam}.} \bibinfo{year}{2022}\natexlab{}.
\newblock \showarticletitle{FedNet: Federated Implementation of Neural Networks for Facial Expression Recognition}. In \bibinfo{booktitle}{\emph{2022 International Conference on Decision Aid Sciences and Applications (DASA)}}. \bibinfo{pages}{82--87}.
\newblock
\urldef\tempurl%
\url{https://doi.org/10.1109/DASA54658.2022.9765165}
\showDOI{\tempurl}


\bibitem[Silva et~al\mbox{.}(2022)]%
        {9892150}
\bibfield{author}{\bibinfo{person}{Victor E. De~S. Silva}, \bibinfo{person}{Tiago~B. Lacerda}, \bibinfo{person}{Péricles~B.C. Miranda}, \bibinfo{person}{André~C.A. Nascimento}, {and} \bibinfo{person}{Ana Paula~C. Furtado}.} \bibinfo{year}{2022}\natexlab{}.
\newblock \showarticletitle{Federated Learning for Physical Violence Detection in Videos}. In \bibinfo{booktitle}{\emph{2022 International Joint Conference on Neural Networks (IJCNN)}}. \bibinfo{pages}{1--8}.
\newblock
\urldef\tempurl%
\url{https://doi.org/10.1109/IJCNN55064.2022.9892150}
\showDOI{\tempurl}


\bibitem[Singh et~al\mbox{.}(2022)]%
        {singh2022privacy}
\bibfield{author}{\bibinfo{person}{Ankit~Kumar Singh}, \bibinfo{person}{Ajit Kumar}, {and} \bibinfo{person}{Bong~Jun Choi}.} \bibinfo{year}{2022}\natexlab{}.
\newblock \showarticletitle{Privacy-Preserving Digital Intervention for Mental Health Using Federated Learning}. In \bibinfo{booktitle}{\emph{International Conference on Intelligent Human Computer Interaction}}. Springer, \bibinfo{pages}{213--224}.
\newblock


\bibitem[Singhal et~al\mbox{.}(2023)]%
        {10101124}
\bibfield{author}{\bibinfo{person}{Rijul Singhal}, \bibinfo{person}{Hardik Modi}, \bibinfo{person}{S Srihari}, \bibinfo{person}{Advit Gandhi}, \bibinfo{person}{C~O Prakash}, {and} \bibinfo{person}{Sivaraman Eswaran}.} \bibinfo{year}{2023}\natexlab{}.
\newblock \showarticletitle{Body Posture Correction and Hand Gesture Detection Using Federated Learning and Mediapipe}. In \bibinfo{booktitle}{\emph{2023 2nd International Conference for Innovation in Technology (INOCON)}}. \bibinfo{pages}{1--6}.
\newblock
\urldef\tempurl%
\url{https://doi.org/10.1109/INOCON57975.2023.10101124}
\showDOI{\tempurl}


\bibitem[Sozinov et~al\mbox{.}(2018)]%
        {8672262}
\bibfield{author}{\bibinfo{person}{Konstantin Sozinov}, \bibinfo{person}{Vladimir Vlassov}, {and} \bibinfo{person}{Sarunas Girdzijauskas}.} \bibinfo{year}{2018}\natexlab{}.
\newblock \showarticletitle{Human Activity Recognition Using Federated Learning}. In \bibinfo{booktitle}{\emph{2018 IEEE Intl Conf on Parallel \& Distributed Processing with Applications, Ubiquitous Computing \& Communications, Big Data \& Cloud Computing, Social Computing \& Networking, Sustainable Computing \& Communications (ISPA/IUCC/BDCloud/SocialCom/SustainCom)}}. \bibinfo{pages}{1103--1111}.
\newblock
\urldef\tempurl%
\url{https://doi.org/10.1109/BDCloud.2018.00164}
\showDOI{\tempurl}


\bibitem[Sun et~al\mbox{.}(2022)]%
        {9618642}
\bibfield{author}{\bibinfo{person}{Gan Sun}, \bibinfo{person}{Yang Cong}, \bibinfo{person}{Jiahua Dong}, \bibinfo{person}{Qiang Wang}, \bibinfo{person}{Lingjuan Lyu}, {and} \bibinfo{person}{Ji Liu}.} \bibinfo{year}{2022}\natexlab{}.
\newblock \showarticletitle{Data Poisoning Attacks on Federated Machine Learning}.
\newblock \bibinfo{journal}{\emph{IEEE Internet of Things Journal}} \bibinfo{volume}{9}, \bibinfo{number}{13} (\bibinfo{year}{2022}), \bibinfo{pages}{11365--11375}.
\newblock
\urldef\tempurl%
\url{https://doi.org/10.1109/JIOT.2021.3128646}
\showDOI{\tempurl}


\bibitem[Suruliraj and Orji(2022)]%
        {9978600}
\bibfield{author}{\bibinfo{person}{Banuchitra Suruliraj} {and} \bibinfo{person}{Rita Orji}.} \bibinfo{year}{2022}\natexlab{}.
\newblock \showarticletitle{Federated Learning Framework for Mobile Sensing Apps in Mental Health}. In \bibinfo{booktitle}{\emph{2022 IEEE 10th International Conference on Serious Games and Applications for Health(SeGAH)}}. \bibinfo{pages}{1--7}.
\newblock
\urldef\tempurl%
\url{https://doi.org/10.1109/SEGAH54908.2022.9978600}
\showDOI{\tempurl}


\bibitem[Tabatabaie and He(2024)]%
        {10.1145/3631421}
\bibfield{author}{\bibinfo{person}{Mahan Tabatabaie} {and} \bibinfo{person}{Suining He}.} \bibinfo{year}{2024}\natexlab{}.
\newblock \showarticletitle{Driver Maneuver Interaction Identification with Anomaly-Aware Federated Learning on Heterogeneous Feature Representations}.
\newblock \bibinfo{journal}{\emph{Proc. ACM Interact. Mob. Wearable Ubiquitous Technol.}} \bibinfo{volume}{7}, \bibinfo{number}{4}, Article \bibinfo{articleno}{180} (\bibinfo{date}{Jan.} \bibinfo{year}{2024}), \bibinfo{numpages}{28}~pages.
\newblock
\urldef\tempurl%
\url{https://doi.org/10.1145/3631421}
\showDOI{\tempurl}


\bibitem[Tan et~al\mbox{.}(2020)]%
        {10.1007/978-3-030-59419-0_54}
\bibfield{author}{\bibinfo{person}{Conghui Tan}, \bibinfo{person}{Di Jiang}, \bibinfo{person}{Huaxiao Mo}, \bibinfo{person}{Jinhua Peng}, \bibinfo{person}{Yongxin Tong}, \bibinfo{person}{Weiwei Zhao}, \bibinfo{person}{Chaotao Chen}, \bibinfo{person}{Rongzhong Lian}, \bibinfo{person}{Yuanfeng Song}, {and} \bibinfo{person}{Qian Xu}.} \bibinfo{year}{2020}\natexlab{}.
\newblock \showarticletitle{Federated Acoustic Model Optimization for Automatic Speech Recognition}. In \bibinfo{booktitle}{\emph{Database Systems for Advanced Applications}}, \bibfield{editor}{\bibinfo{person}{Yunmook Nah}, \bibinfo{person}{Bin Cui}, \bibinfo{person}{Sang-Won Lee}, \bibinfo{person}{Jeffrey~Xu Yu}, \bibinfo{person}{Yang-Sae Moon}, {and} \bibinfo{person}{Steven~Euijong Whang}} (Eds.). \bibinfo{publisher}{Springer International Publishing}, \bibinfo{address}{Cham}, \bibinfo{pages}{771--774}.
\newblock
\showISBNx{978-3-030-59419-0}


\bibitem[Tasbaz et~al\mbox{.}(2022)]%
        {9960135}
\bibfield{author}{\bibinfo{person}{Omid Tasbaz}, \bibinfo{person}{Vahideh Moghtadaiee}, {and} \bibinfo{person}{Bahar Farahani}.} \bibinfo{year}{2022}\natexlab{}.
\newblock \showarticletitle{Zone-Based Federated Learning in Indoor Positioning}. In \bibinfo{booktitle}{\emph{2022 12th International Conference on Computer and Knowledge Engineering (ICCKE)}}. \bibinfo{pages}{163--168}.
\newblock
\urldef\tempurl%
\url{https://doi.org/10.1109/ICCKE57176.2022.9960135}
\showDOI{\tempurl}


\bibitem[Teixeira et~al\mbox{.}(2010)]%
        {teixeira2010survey}
\bibfield{author}{\bibinfo{person}{Thiago Teixeira}, \bibinfo{person}{Gershon Dublon}, {and} \bibinfo{person}{Andreas Savvides}.} \bibinfo{year}{2010}\natexlab{}.
\newblock \showarticletitle{A survey of human-sensing: Methods for detecting presence, count, location, track, and identity}.
\newblock \bibinfo{journal}{\emph{Comput. Surveys}} \bibinfo{volume}{5}, \bibinfo{number}{1} (\bibinfo{year}{2010}), \bibinfo{pages}{59--69}.
\newblock


\bibitem[Thieme et~al\mbox{.}(2020)]%
        {10.1145/3398069}
\bibfield{author}{\bibinfo{person}{Anja Thieme}, \bibinfo{person}{Danielle Belgrave}, {and} \bibinfo{person}{Gavin Doherty}.} \bibinfo{year}{2020}\natexlab{}.
\newblock \showarticletitle{Machine Learning in Mental Health: A Systematic Review of the HCI Literature to Support the Development of Effective and Implementable ML Systems}.
\newblock \bibinfo{journal}{\emph{ACM Trans. Comput.-Hum. Interact.}} \bibinfo{volume}{27}, \bibinfo{number}{5}, Article \bibinfo{articleno}{34} (\bibinfo{date}{aug} \bibinfo{year}{2020}), \bibinfo{numpages}{53}~pages.
\newblock
\showISSN{1073-0516}
\urldef\tempurl%
\url{https://doi.org/10.1145/3398069}
\showDOI{\tempurl}


\bibitem[Tirumala et~al\mbox{.}(2017)]%
        {TIRUMALA2017250}
\bibfield{author}{\bibinfo{person}{Sreenivas~Sremath Tirumala}, \bibinfo{person}{Seyed~Reza Shahamiri}, \bibinfo{person}{Abhimanyu~Singh Garhwal}, {and} \bibinfo{person}{Ruili Wang}.} \bibinfo{year}{2017}\natexlab{}.
\newblock \showarticletitle{Speaker identification features extraction methods: A systematic review}.
\newblock \bibinfo{journal}{\emph{Expert Systems with Applications}}  \bibinfo{volume}{90} (\bibinfo{year}{2017}), \bibinfo{pages}{250--271}.
\newblock
\showISSN{0957-4174}
\urldef\tempurl%
\url{https://doi.org/10.1016/j.eswa.2017.08.015}
\showDOI{\tempurl}


\bibitem[Tomashenko et~al\mbox{.}(2022)]%
        {9746541}
\bibfield{author}{\bibinfo{person}{Natalia Tomashenko}, \bibinfo{person}{Salima Mdhaffar}, \bibinfo{person}{Marc Tommasi}, \bibinfo{person}{Yannick Estève}, {and} \bibinfo{person}{Jean-François Bonastre}.} \bibinfo{year}{2022}\natexlab{}.
\newblock \showarticletitle{Privacy Attacks for Automatic Speech Recognition Acoustic Models in A Federated Learning Framework}. In \bibinfo{booktitle}{\emph{ICASSP 2022 - 2022 IEEE International Conference on Acoustics, Speech and Signal Processing (ICASSP)}}. \bibinfo{pages}{6972--6976}.
\newblock
\urldef\tempurl%
\url{https://doi.org/10.1109/ICASSP43922.2022.9746541}
\showDOI{\tempurl}


\bibitem[Tsouvalas et~al\mbox{.}(2022a)]%
        {9767445}
\bibfield{author}{\bibinfo{person}{Vasileios Tsouvalas}, \bibinfo{person}{Tanir Ozcelebi}, {and} \bibinfo{person}{Nirvana Meratnia}.} \bibinfo{year}{2022}\natexlab{a}.
\newblock \showarticletitle{Privacy-preserving Speech Emotion Recognition through Semi-Supervised Federated Learning}. In \bibinfo{booktitle}{\emph{2022 IEEE International Conference on Pervasive Computing and Communications Workshops and other Affiliated Events (PerCom Workshops)}}. \bibinfo{pages}{359--364}.
\newblock
\urldef\tempurl%
\url{https://doi.org/10.1109/PerComWorkshops53856.2022.9767445}
\showDOI{\tempurl}


\bibitem[Tsouvalas et~al\mbox{.}(2022b)]%
        {9746356}
\bibfield{author}{\bibinfo{person}{Vasileios Tsouvalas}, \bibinfo{person}{Aaqib Saeed}, {and} \bibinfo{person}{Tanir Ozcelebi}.} \bibinfo{year}{2022}\natexlab{b}.
\newblock \showarticletitle{Federated Self-Training for Data-Efficient Audio Recognition}. In \bibinfo{booktitle}{\emph{ICASSP 2022 - 2022 IEEE International Conference on Acoustics, Speech and Signal Processing (ICASSP)}}. \bibinfo{pages}{476--480}.
\newblock
\urldef\tempurl%
\url{https://doi.org/10.1109/ICASSP43922.2022.9746356}
\showDOI{\tempurl}


\bibitem[Tsouvalas et~al\mbox{.}(2022c)]%
        {10.1145/3520128}
\bibfield{author}{\bibinfo{person}{Vasileios Tsouvalas}, \bibinfo{person}{Aaqib Saeed}, {and} \bibinfo{person}{Tanir Ozcelebi}.} \bibinfo{year}{2022}\natexlab{c}.
\newblock \showarticletitle{Federated Self-Training for Semi-Supervised Audio Recognition}.
\newblock \bibinfo{journal}{\emph{ACM Trans. Embed. Comput. Syst.}} \bibinfo{volume}{21}, \bibinfo{number}{6}, Article \bibinfo{articleno}{74} (\bibinfo{date}{oct} \bibinfo{year}{2022}), \bibinfo{numpages}{26}~pages.
\newblock
\showISSN{1539-9087}
\urldef\tempurl%
\url{https://doi.org/10.1145/3520128}
\showDOI{\tempurl}


\bibitem[Tu et~al\mbox{.}(2021)]%
        {10.1145/3485730.3485946}
\bibfield{author}{\bibinfo{person}{Linlin Tu}, \bibinfo{person}{Xiaomin Ouyang}, \bibinfo{person}{Jiayu Zhou}, \bibinfo{person}{Yuze He}, {and} \bibinfo{person}{Guoliang Xing}.} \bibinfo{year}{2021}\natexlab{}.
\newblock \showarticletitle{FedDL: Federated Learning via Dynamic Layer Sharing for Human Activity Recognition}. In \bibinfo{booktitle}{\emph{Proceedings of the 19th ACM Conference on Embedded Networked Sensor Systems}} (Coimbra, Portugal) \emph{(\bibinfo{series}{SenSys '21})}. \bibinfo{publisher}{Association for Computing Machinery}, \bibinfo{address}{New York, NY, USA}, \bibinfo{pages}{15–28}.
\newblock
\showISBNx{9781450390972}
\urldef\tempurl%
\url{https://doi.org/10.1145/3485730.3485946}
\showDOI{\tempurl}


\bibitem[Tzinis et~al\mbox{.}(2021)]%
        {9632783}
\bibfield{author}{\bibinfo{person}{Efthymios Tzinis}, \bibinfo{person}{Jonah Casebeer}, \bibinfo{person}{Zhepei Wang}, {and} \bibinfo{person}{Paris Smaragdis}.} \bibinfo{year}{2021}\natexlab{}.
\newblock \showarticletitle{Separate But Together: Unsupervised Federated Learning for Speech Enhancement from Non-IID Data}. In \bibinfo{booktitle}{\emph{2021 IEEE Workshop on Applications of Signal Processing to Audio and Acoustics (WASPAA)}}. \bibinfo{pages}{46--50}.
\newblock
\urldef\tempurl%
\url{https://doi.org/10.1109/WASPAA52581.2021.9632783}
\showDOI{\tempurl}


\bibitem[Vyas et~al\mbox{.}(2023)]%
        {vyas2023federated}
\bibfield{author}{\bibinfo{person}{Jayant Vyas}, \bibinfo{person}{Debasis Das}, \bibinfo{person}{Santanu Chaudhury}, {et~al\mbox{.}}} \bibinfo{year}{2023}\natexlab{}.
\newblock \showarticletitle{Federated learning based driver recommendation for next generation transportation system}.
\newblock \bibinfo{journal}{\emph{Expert Systems with Applications}}  \bibinfo{volume}{225} (\bibinfo{year}{2023}), \bibinfo{pages}{119951}.
\newblock


\bibitem[Wang et~al\mbox{.}(2022a)]%
        {Wang_2022_CVPR}
\bibfield{author}{\bibinfo{person}{Chunnan Wang}, \bibinfo{person}{Xiang Chen}, \bibinfo{person}{Junzhe Wang}, {and} \bibinfo{person}{Hongzhi Wang}.} \bibinfo{year}{2022}\natexlab{a}.
\newblock \showarticletitle{ATPFL: Automatic Trajectory Prediction Model Design Under Federated Learning Framework}. In \bibinfo{booktitle}{\emph{Proceedings of the IEEE/CVF Conference on Computer Vision and Pattern Recognition (CVPR)}}. \bibinfo{pages}{6563--6572}.
\newblock


\bibitem[Wang and Deng(2021)]%
        {WANG2021215}
\bibfield{author}{\bibinfo{person}{Mei Wang} {and} \bibinfo{person}{Weihong Deng}.} \bibinfo{year}{2021}\natexlab{}.
\newblock \showarticletitle{Deep face recognition: A survey}.
\newblock \bibinfo{journal}{\emph{Neurocomputing}}  \bibinfo{volume}{429} (\bibinfo{year}{2021}), \bibinfo{pages}{215--244}.
\newblock
\showISSN{0925-2312}
\urldef\tempurl%
\url{https://doi.org/10.1016/j.neucom.2020.10.081}
\showDOI{\tempurl}


\bibitem[Wang et~al\mbox{.}(2014)]%
        {10.1145/2632048.2632054}
\bibfield{author}{\bibinfo{person}{Rui Wang}, \bibinfo{person}{Fanglin Chen}, \bibinfo{person}{Zhenyu Chen}, \bibinfo{person}{Tianxing Li}, \bibinfo{person}{Gabriella Harari}, \bibinfo{person}{Stefanie Tignor}, \bibinfo{person}{Xia Zhou}, \bibinfo{person}{Dror Ben-Zeev}, {and} \bibinfo{person}{Andrew~T. Campbell}.} \bibinfo{year}{2014}\natexlab{}.
\newblock \showarticletitle{StudentLife: Assessing Mental Health, Academic Performance and Behavioral Trends of College Students Using Smartphones}. In \bibinfo{booktitle}{\emph{Proceedings of the 2014 ACM International Joint Conference on Pervasive and Ubiquitous Computing}} (Seattle, Washington) \emph{(\bibinfo{series}{UbiComp '14})}. \bibinfo{publisher}{Association for Computing Machinery}, \bibinfo{address}{New York, NY, USA}, \bibinfo{pages}{3–14}.
\newblock
\showISBNx{9781450329682}
\urldef\tempurl%
\url{https://doi.org/10.1145/2632048.2632054}
\showDOI{\tempurl}


\bibitem[Wang et~al\mbox{.}(2020)]%
        {wang2020fed}
\bibfield{author}{\bibinfo{person}{Yaojie Wang}, \bibinfo{person}{Xiaolong Cui}, \bibinfo{person}{Zhiqiang Gao}, {and} \bibinfo{person}{Bo Gan}.} \bibinfo{year}{2020}\natexlab{}.
\newblock \showarticletitle{Fed-SCNN: a federated shallow-cnn recognition framework for distracted driving}.
\newblock \bibinfo{journal}{\emph{Security and Communication Networks}}  \bibinfo{volume}{2020} (\bibinfo{year}{2020}), \bibinfo{pages}{1--10}.
\newblock


\bibitem[Wang et~al\mbox{.}(2022b)]%
        {10.1007/978-3-030-95391-1_29}
\bibfield{author}{\bibinfo{person}{Yangqian Wang}, \bibinfo{person}{Yuanfeng Song}, \bibinfo{person}{Di Jiang}, \bibinfo{person}{Ye Ding}, \bibinfo{person}{Xuan Wang}, \bibinfo{person}{Yang Liu}, {and} \bibinfo{person}{Qing Liao}.} \bibinfo{year}{2022}\natexlab{b}.
\newblock \showarticletitle{FedSP: Federated Speaker Verification with Personal Privacy Preservation}. In \bibinfo{booktitle}{\emph{Algorithms and Architectures for Parallel Processing}}, \bibfield{editor}{\bibinfo{person}{Yongxuan Lai}, \bibinfo{person}{Tian Wang}, \bibinfo{person}{Min Jiang}, \bibinfo{person}{Guangquan Xu}, \bibinfo{person}{Wei Liang}, {and} \bibinfo{person}{Aniello Castiglione}} (Eds.). \bibinfo{publisher}{Springer International Publishing}, \bibinfo{address}{Cham}, \bibinfo{pages}{462--478}.
\newblock
\showISBNx{978-3-030-95391-1}


\bibitem[Wang et~al\mbox{.}(2021)]%
        {9546898}
\bibfield{author}{\bibinfo{person}{Zexin Wang}, \bibinfo{person}{Weidong Zhang}, \bibinfo{person}{Xuangou Wu}, {and} \bibinfo{person}{Xiujun Wang}.} \bibinfo{year}{2021}\natexlab{}.
\newblock \showarticletitle{Matched Averaging Federated Learning Gesture Recognition with WiFi Signals}. In \bibinfo{booktitle}{\emph{2021 7th International Conference on Big Data Computing and Communications (BigCom)}}. \bibinfo{pages}{38--43}.
\newblock
\urldef\tempurl%
\url{https://doi.org/10.1109/BigCom53800.2021.00018}
\showDOI{\tempurl}


\bibitem[Warden(2018)]%
        {warden2018speech}
\bibfield{author}{\bibinfo{person}{Pete Warden}.} \bibinfo{year}{2018}\natexlab{}.
\newblock \showarticletitle{Speech commands: A dataset for limited-vocabulary speech recognition}.
\newblock \bibinfo{journal}{\emph{arXiv preprint arXiv:1804.03209}} (\bibinfo{year}{2018}).
\newblock


\bibitem[Waref and Salem(2022)]%
        {9942417}
\bibfield{author}{\bibinfo{person}{Dinah Waref} {and} \bibinfo{person}{Mohammed Salem}.} \bibinfo{year}{2022}\natexlab{}.
\newblock \showarticletitle{Split Federated Learning for Emotion Detection}. In \bibinfo{booktitle}{\emph{2022 4th Novel Intelligent and Leading Emerging Sciences Conference (NILES)}}. \bibinfo{pages}{112--115}.
\newblock
\urldef\tempurl%
\url{https://doi.org/10.1109/NILES56402.2022.9942417}
\showDOI{\tempurl}


\bibitem[Watanabe et~al\mbox{.}(2018)]%
        {watanabe2018espnet}
\bibfield{author}{\bibinfo{person}{Shinji Watanabe}, \bibinfo{person}{Takaaki Hori}, \bibinfo{person}{Shigeki Karita}, \bibinfo{person}{Tomoki Hayashi}, \bibinfo{person}{Jiro Nishitoba}, \bibinfo{person}{Yuya Unno}, \bibinfo{person}{Nelson Enrique~Yalta Soplin}, \bibinfo{person}{Jahn Heymann}, \bibinfo{person}{Matthew Wiesner}, \bibinfo{person}{Nanxin Chen}, {et~al\mbox{.}}} \bibinfo{year}{2018}\natexlab{}.
\newblock \showarticletitle{Espnet: End-to-end speech processing toolkit}.
\newblock \bibinfo{journal}{\emph{arXiv preprint arXiv:1804.00015}} (\bibinfo{year}{2018}).
\newblock


\bibitem[Weng et~al\mbox{.}(2023)]%
        {WENG2023103831}
\bibfield{author}{\bibinfo{person}{Jianfeng Weng}, \bibinfo{person}{Kun Hu}, \bibinfo{person}{Tingting Yao}, \bibinfo{person}{Jingya Wang}, {and} \bibinfo{person}{Zhiyong Wang}.} \bibinfo{year}{2023}\natexlab{}.
\newblock \showarticletitle{Federated Unsupervised Cluster-Contrastive learning for person Re-identification: A coarse-to-fine approach}.
\newblock \bibinfo{journal}{\emph{Computer Vision and Image Understanding}}  \bibinfo{volume}{237} (\bibinfo{year}{2023}), \bibinfo{pages}{103831}.
\newblock
\showISSN{1077-3142}
\urldef\tempurl%
\url{https://doi.org/10.1016/j.cviu.2023.103831}
\showDOI{\tempurl}


\bibitem[Woubie and Bäckström(2021)]%
        {9592761}
\bibfield{author}{\bibinfo{person}{Abraham Woubie} {and} \bibinfo{person}{Tom Bäckström}.} \bibinfo{year}{2021}\natexlab{}.
\newblock \showarticletitle{Federated Learning for Privacy-Preserving Speaker Recognition}.
\newblock \bibinfo{journal}{\emph{IEEE Access}}  \bibinfo{volume}{9} (\bibinfo{year}{2021}), \bibinfo{pages}{149477--149485}.
\newblock
\urldef\tempurl%
\url{https://doi.org/10.1109/ACCESS.2021.3124029}
\showDOI{\tempurl}


\bibitem[Wu and Gong(2021)]%
        {Wu_Gong_2021}
\bibfield{author}{\bibinfo{person}{Guile Wu} {and} \bibinfo{person}{Shaogang Gong}.} \bibinfo{year}{2021}\natexlab{}.
\newblock \showarticletitle{Decentralised Learning from Independent Multi-Domain Labels for Person Re-Identification}.
\newblock \bibinfo{journal}{\emph{Proceedings of the AAAI Conference on Artificial Intelligence}} \bibinfo{volume}{35}, \bibinfo{number}{4} (\bibinfo{date}{May} \bibinfo{year}{2021}), \bibinfo{pages}{2898--2906}.
\newblock
\urldef\tempurl%
\url{https://doi.org/10.1609/aaai.v35i4.16396}
\showDOI{\tempurl}


\bibitem[Wu et~al\mbox{.}(2021a)]%
        {9593115}
\bibfield{author}{\bibinfo{person}{Peng Wu}, \bibinfo{person}{Tales Imbiriba}, \bibinfo{person}{Junha Park}, \bibinfo{person}{Sunwoo Kim}, {and} \bibinfo{person}{Pau Closas}.} \bibinfo{year}{2021}\natexlab{a}.
\newblock \showarticletitle{Personalized Federated Learning over non-IID Data for Indoor Localization}. In \bibinfo{booktitle}{\emph{2021 IEEE 22nd International Workshop on Signal Processing Advances in Wireless Communications (SPAWC)}}. \bibinfo{pages}{421--425}.
\newblock
\urldef\tempurl%
\url{https://doi.org/10.1109/SPAWC51858.2021.9593115}
\showDOI{\tempurl}


\bibitem[Wu et~al\mbox{.}(2021b)]%
        {9658722}
\bibfield{author}{\bibinfo{person}{Zheshun Wu}, \bibinfo{person}{Xiaoping Wu}, \bibinfo{person}{Xiaoli Long}, {and} \bibinfo{person}{Yunliang Long}.} \bibinfo{year}{2021}\natexlab{b}.
\newblock \showarticletitle{A Privacy-Preserved Online Personalized Federated Learning Framework for Indoor Localization}. In \bibinfo{booktitle}{\emph{2021 IEEE International Conference on Systems, Man, and Cybernetics (SMC)}}. \bibinfo{pages}{2834--2839}.
\newblock
\urldef\tempurl%
\url{https://doi.org/10.1109/SMC52423.2021.9658722}
\showDOI{\tempurl}


\bibitem[Wu et~al\mbox{.}(2022a)]%
        {9734052}
\bibfield{author}{\bibinfo{person}{Zheshun Wu}, \bibinfo{person}{Xiaoping Wu}, {and} \bibinfo{person}{Yunliang Long}.} \bibinfo{year}{2022}\natexlab{a}.
\newblock \showarticletitle{Multi-Level Federated Graph Learning and Self-Attention Based Personalized Wi-Fi Indoor Fingerprint Localization}.
\newblock \bibinfo{journal}{\emph{IEEE Communications Letters}} \bibinfo{volume}{26}, \bibinfo{number}{8} (\bibinfo{year}{2022}), \bibinfo{pages}{1794--1798}.
\newblock
\urldef\tempurl%
\url{https://doi.org/10.1109/LCOMM.2022.3159504}
\showDOI{\tempurl}


\bibitem[Wu et~al\mbox{.}(2022b)]%
        {9749277}
\bibfield{author}{\bibinfo{person}{Zheshun Wu}, \bibinfo{person}{Xiaoping Wu}, {and} \bibinfo{person}{Yunliang Long}.} \bibinfo{year}{2022}\natexlab{b}.
\newblock \showarticletitle{Prediction Based Semi-Supervised Online Personalized Federated Learning for Indoor Localization}.
\newblock \bibinfo{journal}{\emph{IEEE Sensors Journal}} \bibinfo{volume}{22}, \bibinfo{number}{11} (\bibinfo{year}{2022}), \bibinfo{pages}{10640--10654}.
\newblock
\urldef\tempurl%
\url{https://doi.org/10.1109/JSEN.2022.3165042}
\showDOI{\tempurl}


\bibitem[Xiao et~al\mbox{.}(2021)]%
        {XIAO2021107338}
\bibfield{author}{\bibinfo{person}{Zhiwen Xiao}, \bibinfo{person}{Xin Xu}, \bibinfo{person}{Huanlai Xing}, \bibinfo{person}{Fuhong Song}, \bibinfo{person}{Xinhan Wang}, {and} \bibinfo{person}{Bowen Zhao}.} \bibinfo{year}{2021}\natexlab{}.
\newblock \showarticletitle{A federated learning system with enhanced feature extraction for human activity recognition}.
\newblock \bibinfo{journal}{\emph{Knowledge-Based Systems}}  \bibinfo{volume}{229} (\bibinfo{year}{2021}), \bibinfo{pages}{107338}.
\newblock
\showISSN{0950-7051}
\urldef\tempurl%
\url{https://doi.org/10.1016/j.knosys.2021.107338}
\showDOI{\tempurl}


\bibitem[Xu et~al\mbox{.}(2021a)]%
        {xu2021federated}
\bibfield{author}{\bibinfo{person}{Jie Xu}, \bibinfo{person}{Benjamin~S Glicksberg}, \bibinfo{person}{Chang Su}, \bibinfo{person}{Peter Walker}, \bibinfo{person}{Jiang Bian}, {and} \bibinfo{person}{Fei Wang}.} \bibinfo{year}{2021}\natexlab{a}.
\newblock \showarticletitle{Federated learning for healthcare informatics}.
\newblock \bibinfo{journal}{\emph{Journal of Healthcare Informatics Research}}  \bibinfo{volume}{5} (\bibinfo{year}{2021}), \bibinfo{pages}{1--19}.
\newblock


\bibitem[Xu et~al\mbox{.}(2022a)]%
        {9853484}
\bibfield{author}{\bibinfo{person}{Shuzhen Xu}, \bibinfo{person}{Yanhong Liu}, {and} \bibinfo{person}{Xin He}.} \bibinfo{year}{2022}\natexlab{a}.
\newblock \showarticletitle{Studies on Human Recognition Activities Based on Federated Learning}. In \bibinfo{booktitle}{\emph{2022 International Conference on Computer Engineering and Artificial Intelligence (ICCEAI)}}. \bibinfo{pages}{372--377}.
\newblock
\urldef\tempurl%
\url{https://doi.org/10.1109/ICCEAI55464.2022.00084}
\showDOI{\tempurl}


\bibitem[Xu et~al\mbox{.}(2022b)]%
        {9540999}
\bibfield{author}{\bibinfo{person}{Xiaohang Xu}, \bibinfo{person}{Hao Peng}, \bibinfo{person}{Md~Zakirul~Alam Bhuiyan}, \bibinfo{person}{Zhifeng Hao}, \bibinfo{person}{Lianzhong Liu}, \bibinfo{person}{Lichao Sun}, {and} \bibinfo{person}{Lifang He}.} \bibinfo{year}{2022}\natexlab{b}.
\newblock \showarticletitle{Privacy-Preserving Federated Depression Detection From Multisource Mobile Health Data}.
\newblock \bibinfo{journal}{\emph{IEEE Transactions on Industrial Informatics}} \bibinfo{volume}{18}, \bibinfo{number}{7} (\bibinfo{year}{2022}), \bibinfo{pages}{4788--4797}.
\newblock
\urldef\tempurl%
\url{https://doi.org/10.1109/TII.2021.3113708}
\showDOI{\tempurl}


\bibitem[Xu et~al\mbox{.}(2021b)]%
        {xu2021fedmood}
\bibfield{author}{\bibinfo{person}{Xiaohang Xu}, \bibinfo{person}{Hao Peng}, \bibinfo{person}{Lichao Sun}, \bibinfo{person}{Md~Zakirul~Alam Bhuiyan}, \bibinfo{person}{Lianzhong Liu}, {and} \bibinfo{person}{Lifang He}.} \bibinfo{year}{2021}\natexlab{b}.
\newblock \showarticletitle{Fedmood: Federated learning on mobile health data for mood detection}.
\newblock \bibinfo{journal}{\emph{arXiv preprint arXiv:2102.09342}} (\bibinfo{year}{2021}).
\newblock


\bibitem[Yang et~al\mbox{.}(2021)]%
        {9413453}
\bibfield{author}{\bibinfo{person}{Chao-Han~Huck Yang}, \bibinfo{person}{Jun Qi}, \bibinfo{person}{Samuel Yen-Chi Chen}, \bibinfo{person}{Pin-Yu Chen}, \bibinfo{person}{Sabato~Marco Siniscalchi}, \bibinfo{person}{Xiaoli Ma}, {and} \bibinfo{person}{Chin-Hui Lee}.} \bibinfo{year}{2021}\natexlab{}.
\newblock \showarticletitle{Decentralizing Feature Extraction with Quantum Convolutional Neural Network for Automatic Speech Recognition}. In \bibinfo{booktitle}{\emph{ICASSP 2021 - 2021 IEEE International Conference on Acoustics, Speech and Signal Processing (ICASSP)}}. \bibinfo{pages}{6523--6527}.
\newblock
\urldef\tempurl%
\url{https://doi.org/10.1109/ICASSP39728.2021.9413453}
\showDOI{\tempurl}


\bibitem[Yang et~al\mbox{.}(2022b)]%
        {yang2022federated}
\bibfield{author}{\bibinfo{person}{Fengxiang Yang}, \bibinfo{person}{Zhun Zhong}, \bibinfo{person}{Zhiming Luo}, \bibinfo{person}{Shaozi Li}, {and} \bibinfo{person}{Nicu Sebe}.} \bibinfo{year}{2022}\natexlab{b}.
\newblock \showarticletitle{Federated and generalized person re-identification through domain and feature hallucinating}.
\newblock \bibinfo{journal}{\emph{arXiv preprint arXiv:2203.02689}} (\bibinfo{year}{2022}).
\newblock


\bibitem[Yang et~al\mbox{.}(2019)]%
        {10.1145/3298981}
\bibfield{author}{\bibinfo{person}{Qiang Yang}, \bibinfo{person}{Yang Liu}, \bibinfo{person}{Tianjian Chen}, {and} \bibinfo{person}{Yongxin Tong}.} \bibinfo{year}{2019}\natexlab{}.
\newblock \showarticletitle{Federated Machine Learning: Concept and Applications}.
\newblock \bibinfo{journal}{\emph{ACM Trans. Intell. Syst. Technol.}} \bibinfo{volume}{10}, \bibinfo{number}{2}, Article \bibinfo{articleno}{12} (\bibinfo{date}{jan} \bibinfo{year}{2019}), \bibinfo{numpages}{19}~pages.
\newblock
\showISSN{2157-6904}
\urldef\tempurl%
\url{https://doi.org/10.1145/3298981}
\showDOI{\tempurl}


\bibitem[Yang et~al\mbox{.}(2018)]%
        {yang2018applied}
\bibfield{author}{\bibinfo{person}{Timothy Yang}, \bibinfo{person}{Galen Andrew}, \bibinfo{person}{Hubert Eichner}, \bibinfo{person}{Haicheng Sun}, \bibinfo{person}{Wei Li}, \bibinfo{person}{Nicholas Kong}, \bibinfo{person}{Daniel Ramage}, {and} \bibinfo{person}{Fran{\c{c}}oise Beaufays}.} \bibinfo{year}{2018}\natexlab{}.
\newblock \showarticletitle{Applied federated learning: Improving google keyboard query suggestions}.
\newblock \bibinfo{journal}{\emph{arXiv preprint arXiv:1812.02903}} (\bibinfo{year}{2018}).
\newblock


\bibitem[Yang et~al\mbox{.}(2022a)]%
        {yang2022cross}
\bibfield{author}{\bibinfo{person}{Xiaoshan Yang}, \bibinfo{person}{Baochen Xiong}, \bibinfo{person}{Yi Huang}, {and} \bibinfo{person}{Changsheng Xu}.} \bibinfo{year}{2022}\natexlab{a}.
\newblock \showarticletitle{Cross-Modal Federated Human Activity Recognition via Modality-Agnostic and Modality-Specific Representation Learning}. In \bibinfo{booktitle}{\emph{Proceedings of the AAAI Conference on Artificial Intelligence}}, Vol.~\bibinfo{volume}{36}. \bibinfo{pages}{3063--3071}.
\newblock


\bibitem[Yin et~al\mbox{.}(2020)]%
        {9250516}
\bibfield{author}{\bibinfo{person}{Feng Yin}, \bibinfo{person}{Zhidi Lin}, \bibinfo{person}{Qinglei Kong}, \bibinfo{person}{Yue Xu}, \bibinfo{person}{Deshi Li}, \bibinfo{person}{Sergios Theodoridis}, {and} \bibinfo{person}{Shuguang~Robert Cui}.} \bibinfo{year}{2020}\natexlab{}.
\newblock \showarticletitle{FedLoc: Federated Learning Framework for Data-Driven Cooperative Localization and Location Data Processing}.
\newblock \bibinfo{journal}{\emph{IEEE Open Journal of Signal Processing}}  \bibinfo{volume}{1} (\bibinfo{year}{2020}), \bibinfo{pages}{187--215}.
\newblock
\urldef\tempurl%
\url{https://doi.org/10.1109/OJSP.2020.3036276}
\showDOI{\tempurl}


\bibitem[Young et~al\mbox{.}(1978)]%
        {young1978rating}
\bibfield{author}{\bibinfo{person}{Robert~C Young}, \bibinfo{person}{Jeffery~T Biggs}, \bibinfo{person}{Veronika~E Ziegler}, {and} \bibinfo{person}{Dolores~A Meyer}.} \bibinfo{year}{1978}\natexlab{}.
\newblock \showarticletitle{A rating scale for mania: reliability, validity and sensitivity}.
\newblock \bibinfo{journal}{\emph{The British journal of psychiatry}} \bibinfo{volume}{133}, \bibinfo{number}{5} (\bibinfo{year}{1978}), \bibinfo{pages}{429--435}.
\newblock


\bibitem[Yu et~al\mbox{.}(2023)]%
        {9656620}
\bibfield{author}{\bibinfo{person}{Hongzheng Yu}, \bibinfo{person}{Zekai Chen}, \bibinfo{person}{Xiao Zhang}, \bibinfo{person}{Xu Chen}, \bibinfo{person}{Fuzhen Zhuang}, \bibinfo{person}{Hui Xiong}, {and} \bibinfo{person}{Xiuzhen Cheng}.} \bibinfo{year}{2023}\natexlab{}.
\newblock \showarticletitle{FedHAR: Semi-Supervised Online Learning for Personalized Federated Human Activity Recognition}.
\newblock \bibinfo{journal}{\emph{IEEE Transactions on Mobile Computing}} \bibinfo{volume}{22}, \bibinfo{number}{6} (\bibinfo{year}{2023}), \bibinfo{pages}{3318--3332}.
\newblock
\urldef\tempurl%
\url{https://doi.org/10.1109/TMC.2021.3136853}
\showDOI{\tempurl}


\bibitem[Yu et~al\mbox{.}(2021)]%
        {9657499}
\bibfield{author}{\bibinfo{person}{Wentao Yu}, \bibinfo{person}{Jan Freiwald}, \bibinfo{person}{Soeren Tewes}, \bibinfo{person}{Fabien Huennemeyer}, {and} \bibinfo{person}{Dorothea Kolossa}.} \bibinfo{year}{2021}\natexlab{}.
\newblock \showarticletitle{Federated Learning in ASR: Not as Easy as You Think}. In \bibinfo{booktitle}{\emph{Speech Communication; 14th ITG Conference}}. \bibinfo{pages}{1--5}.
\newblock


\bibitem[Yu et~al\mbox{.}(2022)]%
        {9984667}
\bibfield{author}{\bibinfo{person}{Zhigang Yu}, \bibinfo{person}{Jiahui Liu}, \bibinfo{person}{Mingchuan Yang}, \bibinfo{person}{Yanmin Cheng}, \bibinfo{person}{Jie Hu}, {and} \bibinfo{person}{Xinchi Li}.} \bibinfo{year}{2022}\natexlab{}.
\newblock \showarticletitle{An Elderly Fall Detection Method Based on Federated Learning and Extreme Learning Machine (Fed-ELM)}.
\newblock \bibinfo{journal}{\emph{IEEE Access}}  \bibinfo{volume}{10} (\bibinfo{year}{2022}), \bibinfo{pages}{130816--130824}.
\newblock
\urldef\tempurl%
\url{https://doi.org/10.1109/ACCESS.2022.3229044}
\showDOI{\tempurl}


\bibitem[Yuan et~al\mbox{.}(2023a)]%
        {Yuan_2023_CVPR}
\bibfield{author}{\bibinfo{person}{Liangqi Yuan}, \bibinfo{person}{Yunsheng Ma}, \bibinfo{person}{Lu Su}, {and} \bibinfo{person}{Ziran Wang}.} \bibinfo{year}{2023}\natexlab{a}.
\newblock \showarticletitle{Peer-to-Peer Federated Continual Learning for Naturalistic Driving Action Recognition}. In \bibinfo{booktitle}{\emph{Proceedings of the IEEE/CVF Conference on Computer Vision and Pattern Recognition (CVPR) Workshops}}. \bibinfo{pages}{5250--5259}.
\newblock


\bibitem[Yuan et~al\mbox{.}(2023b)]%
        {10131959}
\bibfield{author}{\bibinfo{person}{Liangqi Yuan}, \bibinfo{person}{Lu Su}, {and} \bibinfo{person}{Ziran Wang}.} \bibinfo{year}{2023}\natexlab{b}.
\newblock \showarticletitle{Federated Transfer–Ordered–Personalized Learning for Driver Monitoring Application}.
\newblock \bibinfo{journal}{\emph{IEEE Internet of Things Journal}} \bibinfo{volume}{10}, \bibinfo{number}{20} (\bibinfo{year}{2023}), \bibinfo{pages}{18292--18301}.
\newblock
\urldef\tempurl%
\url{https://doi.org/10.1109/JIOT.2023.3279273}
\showDOI{\tempurl}


\bibitem[Yuan et~al\mbox{.}(2024)]%
        {10542323}
\bibfield{author}{\bibinfo{person}{Liangqi Yuan}, \bibinfo{person}{Ziran Wang}, \bibinfo{person}{Lichao Sun}, \bibinfo{person}{Philip~S. Yu}, {and} \bibinfo{person}{Christopher~G. Brinton}.} \bibinfo{year}{2024}\natexlab{}.
\newblock \showarticletitle{Decentralized Federated Learning: A Survey and Perspective}.
\newblock \bibinfo{journal}{\emph{IEEE Internet of Things Journal}} (\bibinfo{year}{2024}), \bibinfo{pages}{1--1}.
\newblock
\urldef\tempurl%
\url{https://doi.org/10.1109/JIOT.2024.3407584}
\showDOI{\tempurl}


\bibitem[Zafar et~al\mbox{.}(2021)]%
        {9564936}
\bibfield{author}{\bibinfo{person}{Atiqa Zafar}, \bibinfo{person}{Christian Prehofer}, {and} \bibinfo{person}{Chih-Hong Cheng}.} \bibinfo{year}{2021}\natexlab{}.
\newblock \showarticletitle{Federated Learning for Driver Status Monitoring}. In \bibinfo{booktitle}{\emph{2021 IEEE International Intelligent Transportation Systems Conference (ITSC)}}. \bibinfo{pages}{1463--1469}.
\newblock
\urldef\tempurl%
\url{https://doi.org/10.1109/ITSC48978.2021.9564936}
\showDOI{\tempurl}


\bibitem[Zafari et~al\mbox{.}(2019)]%
        {8692423}
\bibfield{author}{\bibinfo{person}{Faheem Zafari}, \bibinfo{person}{Athanasios Gkelias}, {and} \bibinfo{person}{Kin~K. Leung}.} \bibinfo{year}{2019}\natexlab{}.
\newblock \showarticletitle{A Survey of Indoor Localization Systems and Technologies}.
\newblock \bibinfo{journal}{\emph{IEEE Communications Surveys \& Tutorials}} \bibinfo{volume}{21}, \bibinfo{number}{3} (\bibinfo{year}{2019}), \bibinfo{pages}{2568--2599}.
\newblock
\urldef\tempurl%
\url{https://doi.org/10.1109/COMST.2019.2911558}
\showDOI{\tempurl}


\bibitem[Zehtabian et~al\mbox{.}(2021)]%
        {9488681}
\bibfield{author}{\bibinfo{person}{Sharare Zehtabian}, \bibinfo{person}{Siavash Khodadadeh}, \bibinfo{person}{Ladislau Bölöni}, {and} \bibinfo{person}{Damla Turgut}.} \bibinfo{year}{2021}\natexlab{}.
\newblock \showarticletitle{Privacy-Preserving Learning of Human Activity Predictors in Smart Environments}. In \bibinfo{booktitle}{\emph{IEEE INFOCOM 2021 - IEEE Conference on Computer Communications}}. \bibinfo{pages}{1--10}.
\newblock
\urldef\tempurl%
\url{https://doi.org/10.1109/INFOCOM42981.2021.9488681}
\showDOI{\tempurl}


\bibitem[Zhang et~al\mbox{.}(2022c)]%
        {10.1145/3507971.3507985}
\bibfield{author}{\bibinfo{person}{Bin Zhang}, \bibinfo{person}{Jingya Wang}, \bibinfo{person}{Junyi Fu}, {and} \bibinfo{person}{Jinxiang Xia}.} \bibinfo{year}{2022}\natexlab{c}.
\newblock \showarticletitle{Driver Action Recognition Using Federated Learning}. In \bibinfo{booktitle}{\emph{Proceedings of the 7th International Conference on Communication and Information Processing}} (Beijing, China) \emph{(\bibinfo{series}{ICCIP '21})}. \bibinfo{publisher}{Association for Computing Machinery}, \bibinfo{address}{New York, NY, USA}, \bibinfo{pages}{74–77}.
\newblock
\showISBNx{9781450385190}
\urldef\tempurl%
\url{https://doi.org/10.1145/3507971.3507985}
\showDOI{\tempurl}


\bibitem[Zhang et~al\mbox{.}(2023c)]%
        {9894363}
\bibfield{author}{\bibinfo{person}{Chunjiong Zhang}, \bibinfo{person}{Mingyong Li}, {and} \bibinfo{person}{Di Wu}.} \bibinfo{year}{2023}\natexlab{c}.
\newblock \showarticletitle{Federated Multidomain Learning With Graph Ensemble Autoencoder GMM for Emotion Recognition}.
\newblock \bibinfo{journal}{\emph{IEEE Transactions on Intelligent Transportation Systems}} \bibinfo{volume}{24}, \bibinfo{number}{7} (\bibinfo{year}{2023}), \bibinfo{pages}{7631--7641}.
\newblock
\urldef\tempurl%
\url{https://doi.org/10.1109/TITS.2022.3203800}
\showDOI{\tempurl}


\bibitem[Zhang et~al\mbox{.}(2023e)]%
        {10248272}
\bibfield{author}{\bibinfo{person}{Chong Zhang}, \bibinfo{person}{Xiao Liu}, \bibinfo{person}{Mingrong Xiang}, \bibinfo{person}{Aiting Yao}, \bibinfo{person}{Xiaoliang Fan}, {and} \bibinfo{person}{Gang Li}.} \bibinfo{year}{2023}\natexlab{e}.
\newblock \showarticletitle{Fed4ReID: Federated Learning with Data Augmentation for Person Re-identification Service in Edge Computing}. In \bibinfo{booktitle}{\emph{2023 IEEE International Conference on Web Services (ICWS)}}. \bibinfo{pages}{64--70}.
\newblock
\urldef\tempurl%
\url{https://doi.org/10.1109/ICWS60048.2023.00021}
\showDOI{\tempurl}


\bibitem[Zhang et~al\mbox{.}(2021b)]%
        {9590400}
\bibfield{author}{\bibinfo{person}{Chong Zhang}, \bibinfo{person}{Xiao Liu}, \bibinfo{person}{Jia Xu}, \bibinfo{person}{Tianxiang Chen}, \bibinfo{person}{Gang Li}, \bibinfo{person}{Frank Jiang}, {and} \bibinfo{person}{Xuejun Li}.} \bibinfo{year}{2021}\natexlab{b}.
\newblock \showarticletitle{An Edge based Federated Learning Framework for Person Re-identification in UAV Delivery Service}. In \bibinfo{booktitle}{\emph{2021 IEEE International Conference on Web Services (ICWS)}}. \bibinfo{pages}{500--505}.
\newblock
\urldef\tempurl%
\url{https://doi.org/10.1109/ICWS53863.2021.00070}
\showDOI{\tempurl}


\bibitem[Zhang et~al\mbox{.}(2022d)]%
        {9632695}
\bibfield{author}{\bibinfo{person}{Chenhan Zhang}, \bibinfo{person}{Yuanshao Zhu}, \bibinfo{person}{Christos Markos}, \bibinfo{person}{Shui Yu}, {and} \bibinfo{person}{James J.~Q. Yu}.} \bibinfo{year}{2022}\natexlab{d}.
\newblock \showarticletitle{Toward Crowdsourced Transportation Mode Identification: A Semisupervised Federated Learning Approach}.
\newblock \bibinfo{journal}{\emph{IEEE Internet of Things Journal}} \bibinfo{volume}{9}, \bibinfo{number}{14} (\bibinfo{year}{2022}), \bibinfo{pages}{11868--11882}.
\newblock
\urldef\tempurl%
\url{https://doi.org/10.1109/JIOT.2021.3132056}
\showDOI{\tempurl}


\bibitem[Zhang et~al\mbox{.}(2021a)]%
        {9634881}
\bibfield{author}{\bibinfo{person}{Junpeng Zhang}, \bibinfo{person}{Mengqian Li}, \bibinfo{person}{Shuiguang Zeng}, \bibinfo{person}{Bin Xie}, {and} \bibinfo{person}{Dongmei Zhao}.} \bibinfo{year}{2021}\natexlab{a}.
\newblock \showarticletitle{A survey on security and privacy threats to federated learning}. In \bibinfo{booktitle}{\emph{2021 International Conference on Networking and Network Applications (NaNA)}}. \bibinfo{pages}{319--326}.
\newblock
\urldef\tempurl%
\url{https://doi.org/10.1109/NaNA53684.2021.00062}
\showDOI{\tempurl}


\bibitem[Zhang et~al\mbox{.}(2023b)]%
        {10142016}
\bibfield{author}{\bibinfo{person}{Lei Zhang}, \bibinfo{person}{Guanyu Gao}, {and} \bibinfo{person}{Huaizheng Zhang}.} \bibinfo{year}{2023}\natexlab{b}.
\newblock \showarticletitle{Spatial-Temporal Federated Learning for Lifelong Person Re-identification on Distributed Edges}.
\newblock \bibinfo{journal}{\emph{IEEE Transactions on Circuits and Systems for Video Technology}} (\bibinfo{year}{2023}), \bibinfo{pages}{1--1}.
\newblock
\urldef\tempurl%
\url{https://doi.org/10.1109/TCSVT.2023.3281983}
\showDOI{\tempurl}


\bibitem[Zhang et~al\mbox{.}(2021c)]%
        {zhang2021adaptive}
\bibfield{author}{\bibinfo{person}{Marvin Zhang}, \bibinfo{person}{Henrik Marklund}, \bibinfo{person}{Nikita Dhawan}, \bibinfo{person}{Abhishek Gupta}, \bibinfo{person}{Sergey Levine}, {and} \bibinfo{person}{Chelsea Finn}.} \bibinfo{year}{2021}\natexlab{c}.
\newblock \showarticletitle{Adaptive risk minimization: Learning to adapt to domain shift}.
\newblock \bibinfo{journal}{\emph{Advances in Neural Information Processing Systems}}  \bibinfo{volume}{34} (\bibinfo{year}{2021}), \bibinfo{pages}{23664--23678}.
\newblock


\bibitem[Zhang et~al\mbox{.}(2023g)]%
        {10.1145/3581783.3612350}
\bibfield{author}{\bibinfo{person}{Pengling Zhang}, \bibinfo{person}{Huibin Yan}, \bibinfo{person}{Wenhui Wu}, {and} \bibinfo{person}{Shuoyao Wang}.} \bibinfo{year}{2023}\natexlab{g}.
\newblock \showarticletitle{Improving Federated Person Re-Identification through Feature-Aware Proximity and Aggregation}. In \bibinfo{booktitle}{\emph{Proceedings of the 31st ACM International Conference on Multimedia}} (Ottawa ON, Canada) \emph{(\bibinfo{series}{MM '23})}. \bibinfo{publisher}{Association for Computing Machinery}, \bibinfo{address}{New York, NY, USA}, \bibinfo{pages}{2498–2506}.
\newblock
\showISBNx{9798400701085}
\urldef\tempurl%
\url{https://doi.org/10.1145/3581783.3612350}
\showDOI{\tempurl}


\bibitem[Zhang et~al\mbox{.}(2023a)]%
        {10096500}
\bibfield{author}{\bibinfo{person}{Tuo Zhang}, \bibinfo{person}{Tiantian Feng}, \bibinfo{person}{Samiul Alam}, \bibinfo{person}{Sunwoo Lee}, \bibinfo{person}{Mi Zhang}, \bibinfo{person}{Shrikanth~S. Narayanan}, {and} \bibinfo{person}{Salman Avestimehr}.} \bibinfo{year}{2023}\natexlab{a}.
\newblock \showarticletitle{FedAudio: A Federated Learning Benchmark for Audio Tasks}. In \bibinfo{booktitle}{\emph{ICASSP 2023 - 2023 IEEE International Conference on Acoustics, Speech and Signal Processing (ICASSP)}}. \bibinfo{pages}{1--5}.
\newblock
\urldef\tempurl%
\url{https://doi.org/10.1109/ICASSP49357.2023.10096500}
\showDOI{\tempurl}


\bibitem[Zhang et~al\mbox{.}(2022a)]%
        {9773116}
\bibfield{author}{\bibinfo{person}{Tuo Zhang}, \bibinfo{person}{Lei Gao}, \bibinfo{person}{Chaoyang He}, \bibinfo{person}{Mi Zhang}, \bibinfo{person}{Bhaskar Krishnamachari}, {and} \bibinfo{person}{A.~Salman Avestimehr}.} \bibinfo{year}{2022}\natexlab{a}.
\newblock \showarticletitle{Federated Learning for the Internet of Things: Applications, Challenges, and Opportunities}.
\newblock \bibinfo{journal}{\emph{IEEE Internet of Things Magazine}} \bibinfo{volume}{5}, \bibinfo{number}{1} (\bibinfo{year}{2022}), \bibinfo{pages}{24--29}.
\newblock
\urldef\tempurl%
\url{https://doi.org/10.1109/IOTM.004.2100182}
\showDOI{\tempurl}


\bibitem[Zhang et~al\mbox{.}(2022b)]%
        {zhang2022wifi}
\bibfield{author}{\bibinfo{person}{Weidong Zhang}, \bibinfo{person}{Zexing Wang}, {and} \bibinfo{person}{Xuangou Wu}.} \bibinfo{year}{2022}\natexlab{b}.
\newblock \showarticletitle{WiFi signal-based gesture recognition using federated parameter-matched aggregation}.
\newblock \bibinfo{journal}{\emph{Sensors}} \bibinfo{volume}{22}, \bibinfo{number}{6} (\bibinfo{year}{2022}), \bibinfo{pages}{2349}.
\newblock


\bibitem[Zhang et~al\mbox{.}(2023f)]%
        {10074979}
\bibfield{author}{\bibinfo{person}{Xiao Zhang}, \bibinfo{person}{Qilin Wang}, \bibinfo{person}{Ziming Ye}, \bibinfo{person}{Haochao Ying}, {and} \bibinfo{person}{Dongxiao Yu}.} \bibinfo{year}{2023}\natexlab{f}.
\newblock \showarticletitle{Federated Representation Learning With Data Heterogeneity for Human Mobility Prediction}.
\newblock \bibinfo{journal}{\emph{IEEE Transactions on Intelligent Transportation Systems}} \bibinfo{volume}{24}, \bibinfo{number}{6} (\bibinfo{year}{2023}), \bibinfo{pages}{6111--6122}.
\newblock
\urldef\tempurl%
\url{https://doi.org/10.1109/TITS.2023.3252029}
\showDOI{\tempurl}


\bibitem[Zhang et~al\mbox{.}(2023d)]%
        {10.1145/3581783.3612134}
\bibfield{author}{\bibinfo{person}{Yuxuan Zhang}, \bibinfo{person}{Lei Liu}, {and} \bibinfo{person}{Li Liu}.} \bibinfo{year}{2023}\natexlab{d}.
\newblock \showarticletitle{Cuing Without Sharing: A Federated Cued Speech Recognition Framework via Mutual Knowledge Distillation}. In \bibinfo{booktitle}{\emph{Proceedings of the 31st ACM International Conference on Multimedia}} (<conf-loc>, <city>Ottawa ON</city>, <country>Canada</country>, </conf-loc>) \emph{(\bibinfo{series}{MM '23})}. \bibinfo{publisher}{Association for Computing Machinery}, \bibinfo{address}{New York, NY, USA}, \bibinfo{pages}{8781–8789}.
\newblock
\showISBNx{9798400701085}
\urldef\tempurl%
\url{https://doi.org/10.1145/3581783.3612134}
\showDOI{\tempurl}


\bibitem[Zhao et~al\mbox{.}(2023b)]%
        {zhao2023fedsup}
\bibfield{author}{\bibinfo{person}{Chen Zhao}, \bibinfo{person}{Zhipeng Gao}, \bibinfo{person}{Qian Wang}, \bibinfo{person}{Kaile Xiao}, \bibinfo{person}{Zijia Mo}, {and} \bibinfo{person}{M~Jamal Deen}.} \bibinfo{year}{2023}\natexlab{b}.
\newblock \showarticletitle{FedSup: A communication-efficient federated learning fatigue driving behaviors supervision approach}.
\newblock \bibinfo{journal}{\emph{Future Generation Computer Systems}}  \bibinfo{volume}{138} (\bibinfo{year}{2023}), \bibinfo{pages}{52--60}.
\newblock


\bibitem[Zhao et~al\mbox{.}(2023a)]%
        {10095737}
\bibfield{author}{\bibinfo{person}{Huan Zhao}, \bibinfo{person}{Haijiao Chen}, \bibinfo{person}{Yufeng Xiao}, {and} \bibinfo{person}{Zixing Zhang}.} \bibinfo{year}{2023}\natexlab{a}.
\newblock \showarticletitle{Privacy-Enhanced Federated Learning Against Attribute Inference Attack for Speech Emotion Recognition}. In \bibinfo{booktitle}{\emph{ICASSP 2023 - 2023 IEEE International Conference on Acoustics, Speech and Signal Processing (ICASSP)}}. \bibinfo{pages}{1--5}.
\newblock
\urldef\tempurl%
\url{https://doi.org/10.1109/ICASSP49357.2023.10095737}
\showDOI{\tempurl}


\bibitem[Zhao et~al\mbox{.}(2018)]%
        {zhao2018federated}
\bibfield{author}{\bibinfo{person}{Yue Zhao}, \bibinfo{person}{Meng Li}, \bibinfo{person}{Liangzhen Lai}, \bibinfo{person}{Naveen Suda}, \bibinfo{person}{Damon Civin}, {and} \bibinfo{person}{Vikas Chandra}.} \bibinfo{year}{2018}\natexlab{}.
\newblock \showarticletitle{Federated learning with non-iid data}.
\newblock \bibinfo{journal}{\emph{arXiv preprint arXiv:1806.00582}} (\bibinfo{year}{2018}).
\newblock


\bibitem[Zhao et~al\mbox{.}(2020)]%
        {zhao2020semi}
\bibfield{author}{\bibinfo{person}{Yuchen Zhao}, \bibinfo{person}{Hanyang Liu}, \bibinfo{person}{Honglin Li}, \bibinfo{person}{Payam Barnaghi}, {and} \bibinfo{person}{Hamed Haddadi}.} \bibinfo{year}{2020}\natexlab{}.
\newblock \showarticletitle{Semi-supervised federated learning for activity recognition}.
\newblock \bibinfo{journal}{\emph{arXiv preprint arXiv:2011.00851}} (\bibinfo{year}{2020}).
\newblock


\bibitem[Zheng et~al\mbox{.}(2022)]%
        {9886171}
\bibfield{author}{\bibinfo{person}{Haipeng Zheng}, \bibinfo{person}{Bing Li}, \bibinfo{person}{Guozhu Liu}, \bibinfo{person}{Yuqi Li}, \bibinfo{person}{Yan Zhang}, \bibinfo{person}{Wenhui Gao}, {and} \bibinfo{person}{Xiangtao Zhao}.} \bibinfo{year}{2022}\natexlab{}.
\newblock \showarticletitle{Blockchain-based Federated Learning Framework Applied in Face Recognition}. In \bibinfo{booktitle}{\emph{2022 7th International Conference on Signal and Image Processing (ICSIP)}}. \bibinfo{pages}{265--269}.
\newblock
\urldef\tempurl%
\url{https://doi.org/10.1109/ICSIP55141.2022.9886171}
\showDOI{\tempurl}


\bibitem[Zheng et~al\mbox{.}(2016)]%
        {zheng2016person}
\bibfield{author}{\bibinfo{person}{Liang Zheng}, \bibinfo{person}{Yi Yang}, {and} \bibinfo{person}{Alexander~G Hauptmann}.} \bibinfo{year}{2016}\natexlab{}.
\newblock \showarticletitle{Person re-identification: Past, present and future}.
\newblock \bibinfo{journal}{\emph{arXiv preprint arXiv:1610.02984}} (\bibinfo{year}{2016}).
\newblock


\bibitem[Zhou et~al\mbox{.}(2022)]%
        {9693094}
\bibfield{author}{\bibinfo{person}{Xiaokang Zhou}, \bibinfo{person}{Wei Liang}, \bibinfo{person}{Jianhua Ma}, \bibinfo{person}{Zheng Yan}, {and} \bibinfo{person}{Kevin I-Kai Wang}.} \bibinfo{year}{2022}\natexlab{}.
\newblock \showarticletitle{2D Federated Learning for Personalized Human Activity Recognition in Cyber-Physical-Social Systems}.
\newblock \bibinfo{journal}{\emph{IEEE Transactions on Network Science and Engineering}} \bibinfo{volume}{9}, \bibinfo{number}{6} (\bibinfo{year}{2022}), \bibinfo{pages}{3934--3944}.
\newblock
\urldef\tempurl%
\url{https://doi.org/10.1109/TNSE.2022.3144699}
\showDOI{\tempurl}


\bibitem[Zhu et~al\mbox{.}(2021)]%
        {zhu2021federated}
\bibfield{author}{\bibinfo{person}{Hangyu Zhu}, \bibinfo{person}{Jinjin Xu}, \bibinfo{person}{Shiqing Liu}, {and} \bibinfo{person}{Yaochu Jin}.} \bibinfo{year}{2021}\natexlab{}.
\newblock \showarticletitle{Federated learning on non-IID data: A survey}.
\newblock \bibinfo{journal}{\emph{Neurocomputing}}  \bibinfo{volume}{465} (\bibinfo{year}{2021}), \bibinfo{pages}{371--390}.
\newblock


\bibitem[Zhu et~al\mbox{.}(2019)]%
        {zhu2019deep}
\bibfield{author}{\bibinfo{person}{Ligeng Zhu}, \bibinfo{person}{Zhijian Liu}, {and} \bibinfo{person}{Song Han}.} \bibinfo{year}{2019}\natexlab{}.
\newblock \showarticletitle{Deep leakage from gradients}.
\newblock \bibinfo{journal}{\emph{Advances in neural information processing systems}}  \bibinfo{volume}{32} (\bibinfo{year}{2019}).
\newblock


\bibitem[Zhu(2005)]%
        {zhu2005semi}
\bibfield{author}{\bibinfo{person}{Xiaojin~Jerry Zhu}.} \bibinfo{year}{2005}\natexlab{}.
\newblock \showarticletitle{Semi-supervised learning literature survey}.
\newblock  (\bibinfo{year}{2005}).
\newblock


\bibitem[Zhu et~al\mbox{.}(2022)]%
        {9514368}
\bibfield{author}{\bibinfo{person}{Yuanshao Zhu}, \bibinfo{person}{Yi Liu}, \bibinfo{person}{James J.~Q. Yu}, {and} \bibinfo{person}{Xingliang Yuan}.} \bibinfo{year}{2022}\natexlab{}.
\newblock \showarticletitle{Semi-Supervised Federated Learning for Travel Mode Identification From GPS Trajectories}.
\newblock \bibinfo{journal}{\emph{IEEE Transactions on Intelligent Transportation Systems}} \bibinfo{volume}{23}, \bibinfo{number}{3} (\bibinfo{year}{2022}), \bibinfo{pages}{2380--2391}.
\newblock
\urldef\tempurl%
\url{https://doi.org/10.1109/TITS.2021.3092015}
\showDOI{\tempurl}


\bibitem[Zhu et~al\mbox{.}(2023)]%
        {10283763}
\bibfield{author}{\bibinfo{person}{Yonghui Zhu}, \bibinfo{person}{Ronghui Zhang}, \bibinfo{person}{Yuanhao Cui}, \bibinfo{person}{Sheng Wu}, \bibinfo{person}{Chunxiao Jiang}, {and} \bibinfo{person}{XiaoJun Jing}.} \bibinfo{year}{2023}\natexlab{}.
\newblock \showarticletitle{Communication-Efficient Personalized Federated Edge Learning for Decentralized Sensing in ISAC}. In \bibinfo{booktitle}{\emph{2023 IEEE International Conference on Communications Workshops (ICC Workshops)}}. \bibinfo{pages}{207--212}.
\newblock
\urldef\tempurl%
\url{https://doi.org/10.1109/ICCWorkshops57953.2023.10283763}
\showDOI{\tempurl}


\bibitem[Zhu et~al\mbox{.}(2020)]%
        {9359187}
\bibfield{author}{\bibinfo{person}{Yuanshao Zhu}, \bibinfo{person}{Shuyu Zhang}, \bibinfo{person}{Yi Liu}, \bibinfo{person}{Dusit Niyato}, {and} \bibinfo{person}{James~J.Q. Yu}.} \bibinfo{year}{2020}\natexlab{}.
\newblock \showarticletitle{Robust Federated Learning Approach for Travel Mode Identification from Non-IID GPS Trajectories}. In \bibinfo{booktitle}{\emph{2020 IEEE 26th International Conference on Parallel and Distributed Systems (ICPADS)}}. \bibinfo{pages}{585--592}.
\newblock
\urldef\tempurl%
\url{https://doi.org/10.1109/ICPADS51040.2020.00081}
\showDOI{\tempurl}


\bibitem[Zhuang et~al\mbox{.}(2023)]%
        {10.1145/3531013}
\bibfield{author}{\bibinfo{person}{Weiming Zhuang}, \bibinfo{person}{Xin Gan}, \bibinfo{person}{Yonggang Wen}, {and} \bibinfo{person}{Shuai Zhang}.} \bibinfo{year}{2023}\natexlab{}.
\newblock \showarticletitle{Optimizing Performance of Federated Person Re-Identification: Benchmarking and Analysis}.
\newblock \bibinfo{journal}{\emph{ACM Trans. Multimedia Comput. Commun. Appl.}} \bibinfo{volume}{19}, \bibinfo{number}{1s}, Article \bibinfo{articleno}{38} (\bibinfo{date}{jan} \bibinfo{year}{2023}), \bibinfo{numpages}{18}~pages.
\newblock
\showISSN{1551-6857}
\urldef\tempurl%
\url{https://doi.org/10.1145/3531013}
\showDOI{\tempurl}


\bibitem[ZHuang et~al\mbox{.}(2021)]%
        {zhuang2021towards}
\bibfield{author}{\bibinfo{person}{Weiming ZHuang}, \bibinfo{person}{Xin Gan}, \bibinfo{person}{Yonggang Wen}, \bibinfo{person}{Xuesen Zhang}, \bibinfo{person}{Shuai Zhang}, {and} \bibinfo{person}{Shuai Yi}.} \bibinfo{year}{2021}\natexlab{}.
\newblock \showarticletitle{Towards unsupervised domain adaptation for deep face recognition under privacy constraints via federated learning}.
\newblock \bibinfo{journal}{\emph{arXiv preprint arXiv:2105.07606}} (\bibinfo{year}{2021}).
\newblock


\bibitem[Zhuang et~al\mbox{.}(2022)]%
        {9859587}
\bibfield{author}{\bibinfo{person}{Weiming Zhuang}, \bibinfo{person}{Xin Gan}, \bibinfo{person}{Xuesen Zhang}, \bibinfo{person}{Yonggang Wen}, \bibinfo{person}{Shuai Zhang}, {and} \bibinfo{person}{Shuai Yi}.} \bibinfo{year}{2022}\natexlab{}.
\newblock \showarticletitle{Federated Unsupervised Domain Adaptation for Face Recognition}. In \bibinfo{booktitle}{\emph{2022 IEEE International Conference on Multimedia and Expo (ICME)}}. \bibinfo{pages}{1--6}.
\newblock
\urldef\tempurl%
\url{https://doi.org/10.1109/ICME52920.2022.9859587}
\showDOI{\tempurl}


\bibitem[Zhuang et~al\mbox{.}(2021)]%
        {10.1145/3474085.3475182}
\bibfield{author}{\bibinfo{person}{Weiming Zhuang}, \bibinfo{person}{Yonggang Wen}, {and} \bibinfo{person}{Shuai Zhang}.} \bibinfo{year}{2021}\natexlab{}.
\newblock \showarticletitle{Joint Optimization in Edge-Cloud Continuum for Federated Unsupervised Person Re-Identification}. In \bibinfo{booktitle}{\emph{Proceedings of the 29th ACM International Conference on Multimedia}} (Virtual Event, China) \emph{(\bibinfo{series}{MM '21})}. \bibinfo{publisher}{Association for Computing Machinery}, \bibinfo{address}{New York, NY, USA}, \bibinfo{pages}{433–441}.
\newblock
\showISBNx{9781450386517}
\urldef\tempurl%
\url{https://doi.org/10.1145/3474085.3475182}
\showDOI{\tempurl}


\bibitem[Zhuang et~al\mbox{.}(2020)]%
        {10.1145/3394171.3413814}
\bibfield{author}{\bibinfo{person}{Weiming Zhuang}, \bibinfo{person}{Yonggang Wen}, \bibinfo{person}{Xuesen Zhang}, \bibinfo{person}{Xin Gan}, \bibinfo{person}{Daiying Yin}, \bibinfo{person}{Dongzhan Zhou}, \bibinfo{person}{Shuai Zhang}, {and} \bibinfo{person}{Shuai Yi}.} \bibinfo{year}{2020}\natexlab{}.
\newblock \showarticletitle{Performance Optimization of Federated Person Re-Identification via Benchmark Analysis}. In \bibinfo{booktitle}{\emph{Proceedings of the 28th ACM International Conference on Multimedia}} (Seattle, WA, USA) \emph{(\bibinfo{series}{MM '20})}. \bibinfo{publisher}{Association for Computing Machinery}, \bibinfo{address}{New York, NY, USA}, \bibinfo{pages}{955–963}.
\newblock
\showISBNx{9781450379885}
\urldef\tempurl%
\url{https://doi.org/10.1145/3394171.3413814}
\showDOI{\tempurl}


\end{thebibliography}

\end{document}